\def\year{2019}\relax
\documentclass[letterpaper]{article} 
\usepackage{aaai19}  
\usepackage{times}  
\usepackage{helvet}  
\usepackage{courier}  
\usepackage{url}  
\usepackage{graphicx}  
\usepackage{booktabs}       
\usepackage{amsfonts}       
\usepackage{nicefrac}       
\usepackage{microtype}      
\usepackage{bm}
\usepackage{subfigure}
\usepackage{algorithm}
\usepackage{algorithmic}
\usepackage{amsmath}
\usepackage{amssymb}
\usepackage{enumitem}
\usepackage{multicol}
\newenvironment{proof}[1][Proof]{\begin{trivlist}
		\item[\hskip \labelsep {\bfseries #1}]}{\end{trivlist}}
\newcommand{\invisible}[1]{{}}
\newtheorem{theorem}{Theorem}

\newtheorem{prop}[theorem]{Proposition}
\newtheorem{defin}[theorem]{Definition}

\newtheorem{criterion}{Criterion}

\newcommand{\x}{\mathbf{x}}
\newcommand{\y}{\mathbf{y}}
\newcommand{\z}{\mathbf{z}}
\newcommand{\e}{\mathbf{e}}
\newcommand{\w}{\mathbf{w}}

\newcommand{\ssp}{\mathbf{p}}
\newcommand{\ssq}{\mathbf{q}}

\newcommand{\D}{\mathbf{D}}
\newcommand{\Y}{\mathbf{Y}}
\newcommand{\W}{\mathbf{W}}

\newcommand{\bd}{\mathbf{d}}
\begin{document}

\title{Task Embedded Coordinate Update: A Realizable Framework \\for Multivariate Non-convex Optimization}
\author{Yiyang Wang$^{1,2}$ \quad Risheng Liu$^{3,4,}$\thanks{Corresponding author.} \quad Long Ma$^{3,4}$ \quad Xiaoliang Song$^5$\\
$^1$ Institute of Atmospheric Sciences, Fudan University\\
$^2$ Shanghai Ecological Forecasting and Remote Sensing Center\\
$^3$ DUT-RU International School of Information Science \& Engineering, Dalian University of Technology\\
$^4$ Key Laboratory for Ubiquitous Network and Service Software of Liaoning Province, Dalian University of Technology\\
$^5$ Department of Applied Mathematics, The Hong Kong Polytechnic University\\
\emph{yywerica@fudan.edu.cn \quad rsliu@dlut.edu.cn \quad malone94319@gmail.com \quad xiaoliang.song@polyu.edu.hk}
}
\maketitle
\begin{abstract}
We in this paper propose a realizable framework TECU, which embeds task-specific strategies into update schemes of coordinate descent, for optimizing multivariate non-convex problems with coupled objective functions.
On one hand, TECU is capable of improving algorithm efficiencies through embedding productive numerical algorithms, for optimizing univariate sub-problems with nice properties.
From the other side, it also augments probabilities to receive desired results, by embedding advanced techniques in optimizations of realistic tasks.
Integrating both numerical algorithms and advanced techniques together, TECU is proposed in a unified framework for solving a class of non-convex problems.
Although the task embedded strategies bring inaccuracies in sub-problem optimizations, we provide a realizable criterion to control the errors, meanwhile, to ensure robust performances with rigid theoretical analyses.
By respectively embedding ADMM and a residual-type CNN in our algorithm framework, the experimental results verify both efficiency and effectiveness of embedding task-oriented strategies in coordinate descent for solving practical problems.
\end{abstract}

\section{Introduction}
Over the past few decades, multivariate non-convex optimization has been widely concerned in the realms of pattern recognition and machine learning.
Achieving well performances in the tasks such as matrix factorization \cite{Vu2016Fast,bao2016dictionary} and image enhancement task \cite{fu2016weighted,hasinoff2017Deep}, multivariate non-convex problems have
motivated a revived interest in designing and analyzing numerical algorithms.

Compared to univariate optimization, it is much more complicated to optimize multivariate problems with coupled objective functions.
Taking two variables as instance, this kind of coupled problem can be formulated as:
\begin{equation}\label{eq:2variable}
\min\limits_{\z:=(\x,\y)}\Psi(\z):= f(\x) + g(\y) + H(\x, \y),
\end{equation}
with vectors/matrices $\x$ and $\y$.
Being employed to varieties of tasks, the coordinate descent (CD) \cite{Luo1989On} is widely used for solving problem \eqref{eq:2variable}, which optimizes the objective over each direction, while fixing the remaining one with its latest value, i.e, solving univariate optimization problems in a loop.
Doing in this way, calculating the coordinate updates are much simpler than computing a full update, requiring less memory and computational cost.
However, in addition to these benefits, few CD algorithms consider useful traits of univariate sub-problems for improving either convergence speeds or optimized results for solving the generic problem \eqref{eq:2variable} with non-convex, non-smooth objective function.

In most cases, though $\Psi(\z)$ is non-convex and even non-smooth, it is quite likely to have univariate sub-problems with nice properties: e.g., the sub-problems can be optimized via convex optimization, or may have unique solutions.
Moreover, the univariate problems usually have entirely distinct formations referred to different variables.
For example, many literatures have posted superiorities on restricting dictionaries with normalized bases, meanwhile, constraining sparsity for the codes with various non-convex penalties for dictionary learning tasks \cite{Gregor2010Learning,Wang2016A,bao2016dictionary}.
Though these models are non-convex and non-smooth, their univariate sub-problems of dictionary can be efficiently solved, compared with the other sub-problem.
Thus, in view of the nice traits and specificities of univariate problems, it is significant to specifically integrate effective algorithms for optimizing task-specific sub-problems, to improve the efficiency and effectiveness of CD schemes.

More critically, though there are not a few CD algorithms for solving multivariate optimization problems,
converging to a critical point is still a nice result for generic non-convex and non-smooth problems  \cite{bolte2014proximal,xu2013block,Pock2017Inertial}.
While, we have noticed that, many univariate sub-problems of real-world image processing tasks are referred to specific application problems.
E.g., in tasks like image deblurring and super-resolution, one univariate sub-problem can be regarded as an image denoising task.
Rather than numerical algorithms, techniques such as BM3D and CNN are effective for solving image denoising problem.
Although such advanced techniques mostly lack theoretical support, they have the ability to efficiently project the variables on small neighborhoods of the desired solutions.
Considering the effectiveness of these advanced techniques, it is significant to integrate them into CD schemes, expecting to get desired results with high probability \cite{Zhang2017Learning,Chan2017Plug}.

The above mentioned strategies, integrating either numerical algorithms or advanced techniques, have already appeared in applications for specific tasks, which will be briefly stated later.
However, the success of those CD schemes, designed for specific problems, can not be straightforwardly replicated to other tasks.
Moreover, there has not yet been proposed a unified CD framework, integrating both numerical algorithms and advanced techniques, for optimizing the generic multivariate non-convex problem \eqref{eq:2variable}.
More importantly, few of them are able to provide rigid theoretical analyses on illuminating the properties of the final optimized results.
Considering all the mentioned aspects, we in this paper propose a realizable algorithm framework, which embeds various task-oriented strategies in the update of CD scheme, for effectively solving the generic problem \eqref{eq:2variable}.
We name our proposed algorithm as TECU (task embedded coordinate update), and the main contributions are sketched out as follows:
\begin{enumerate}
	\item For optimizing the generic multivariate problem \eqref{eq:2variable} with coupled non-convex objective, we propose an algorithm TECU, which embeds task-oriented techniques for optimizing specific univariate sub-problems of CD update.
	Moreover, we further provide a realizable condition to ensure robust performances of TECU with theoretical analyses.
	
	\item Considering the nice properties of univariate sub-problems, TECU is able to improve the algorithm efficiency by embedding high-efficient numerical algorithms into its framework.
	We utilize the $\ell_0$-regularized dictionary learning task and design to embed ADMM to accelerate the convergence speed of the whole algorithm.
	Experiments conducted on synthetic data give verifications on the efficiency of TECU,  in comparison with other existing numerical algorithms.
	
	\item Through embedding advanced techniques, TECU is likely to obtain desired solutions with high probability, which is superior to most numerical algorithms for non-convex optimization.
	Taking low-light image enhancement as an example, we embed a residual-type CNN to optimize the univariate problem of illumination layer.
	Then, comparing to state-of-the-art methods, the experimental results show the superiority of embedding networks for real-world tasks, meanwhile, verify the effectiveness of integrating networks and CD schemes in a unified algorithm framework.
	
\end{enumerate}

\subsection{Related Work}\label{relate}

For solving general multivariate non-convex problems, the most classical case that adopts CD scheme is the proximal alternating method (PAM) \cite{Attouch2010Proximal}.
However, it is limited to most coupled problems for requiring explicit solutions for every univariate sub-problems.
To get around this limitation, the PALM linearizes the coupled function, in pursuit of explicit solutions \cite{bolte2014proximal}.
However, it requires computing exact Lipschitz constants during iterations, which sometimes is time-consuming even for estimating their tight upper bounds \cite{bao2016dictionary,xu2013block}.
Moreover, improper upper bounds definitely slow down the convergence speeds of PALM.
These troubles on estimating Lipschitz constants also exist in CD variants like BCU \cite{Xu2017A} and iPALM \cite{Pock2017Inertial}.
Besides the mentioned defects, the updates of existing CD algorithms utterly lose sight of task specificities, i.e., optimizing every univariate sub-problems in the same scheme, which is less efficient in practice.

Unlike algorithms for general problems, it is common to embed numerical algorithms for optimizing sub-problems in real-world applications \cite{guo2014robust,Li2014Single,Li2016Underwater,Yue2017Contrast}.
Such algorithms often make good uses on the nice traits of univariate sub-problems, thus they always possess high efficiencies and superior performances.
However, their specificities give rise to less generalization: the well-designed algorithms usually cannot be borrowed to other models.
Not only this, those specified algorithms mostly have relatively weak convergence in theory, thus their efficiencies are mostly lack of robustness.

Quite recently, fusing advanced techniques into optimization framework has been a hot research interest for real-world applications \cite{Schmidt2014Shrinkage,Liu2018learning,Zhang2017Learning}.
For example, the authors in \cite{Schmidt2014Shrinkage} learn a cascade of shrinkage fields to replace artificially designed priors in a half-quadratic optimization for image restoration.
Instead of designing complex regularizers, \cite{Zhang2017Learning} learns a CNN-based denoiser to replace corresponding sub-problem in their optimization framework.
These novel methods usually have remarkable performances with the power of advanced techniques, however,
their successes rely on completely replacing univariate sub-problems with advanced techniques, thus few of them are able to illuminate the properties on final results with rigid theoretical analyses.

\subsection{Preliminaries}
In general, the objective function of problem \eqref{eq:2variable} is assumed to have:
(1) $f$ and $g$ are proper, lower semi-continuous (l.s.c); (2) $H$ is a $C^1$ function; its gradient and partial gradients are Lipschitz continuous on bounded sets; (3) $\Psi$ is coercive, that is, it is bounded from below and $\Psi(\z)\rightarrow \infty$ when $\|\z\| \rightarrow \infty$, where $\|\cdot\|$ denotes the Frobenius norm.
Meanwhile, $\Psi$ is a Kurdyka-{\L}ojasiewicz (K{\L}) function.

Notice that, all semialgebraic functions and subanalytic functions satisfy the K{\L} property.
Typical semialgebraic functions include real polynomial functions, $\|\cdot\|_p$ with $p\geq 0$, indicator functions of semialgebraic sets, Stiefel manifolds and constant rank matrices \cite{Attouch2010Proximal}.


\section{Task Embedded Coordinate Update}
Corresponding to specific tasks, the univariate sub-problems of the generic model \eqref{eq:2variable} usually either have desirable characteristics or can be corresponding to certain tasks with single variable.
Considering these available traits, we embed powerful strategies in CD scheme, for optimizing task-oriented univariate sub-problems, and then improving the convergence speeds and optimized results of the whole algorithm.

The most basic CD scheme optimizes the objective $\Psi(\z)$ cyclically over each variable, that is,  successively solving the following sub-problems to update $\x^t$ and $\y^t$ at iteration $t$.
\begin{equation}\label{ex_sequence}
\begin{aligned}
&\min\nolimits_{\x} f(\x) + H(\x, \y^{t-1}) + \frac{\eta_1}{2}\|\x-\x^{t-1}\|^2, \\
&\min\nolimits_{\y} g(\y) + H(\x^{t}, \y) + \frac{\eta_2}{2}\|\y-\y^{t-1}\|^2.
\end{aligned}
\end{equation}

Targeting to these univariate sub-problems, we respectively introduce numerical algorithms and advanced techniques to optimize them with a practical error control condition, which provides a criteria on optimization precisions, meanwhile, helps illuminating the properties of final optimized results.

\begin{algorithm}[!t]
	\caption{Task Embedded Coordinate Update}
	\begin{algorithmic}[1]
		\STATE Set constants $C_x$, $C_y$, $\eta_1$, $\eta_2$.
		\STATE Initialize variables $\x^0$, $\y^0$.
		\WHILE {not converged}
		\STATE $\x^{t} \leftarrow$ either $\psi(\x; \x^{t-1}, \y^{t-1}, f, H, \eta_1, C_x, \epsilon_x^{t})$ or the updates in Table \ref{updates}.
		\STATE $\y^{t} \leftarrow$ either $\psi(\y; \y^{t-1}, \x^{t}, g, H, \eta_2, C_y, \epsilon_y^{t})$ or the updates in Table \ref{updates}.
		\IF {$t \geq 2$}
		\STATE $\epsilon_x^{t} = \|\x^{t-1}-\x^{t-2}\|$ and $\epsilon_y^{t} = \|\y^{t-1}-\y^{t-2}\|$.
		\ENDIF
		\ENDWHILE
	\end{algorithmic}\label{Alg:iPAM1}
\end{algorithm}

\begin{algorithm}[!t]
	\caption{$\psi(\mathbf{u}; \mathbf{u}^{t-1}, \mathbf{v}, h, H, \eta, C_u, \epsilon_u^{t})$}
	\begin{algorithmic}[1]
		\STATE Initialize $\mathbf{u}^{t,0}=\mathbf{u}^{t-1}$.
		\WHILE {$\|\e_u^{t,i}\| > C_u\epsilon_u^{t}$}
		\STATE Update $\mathbf{u}^{t,i} = \mathcal{A}_{u}^i(\mathbf{u}^{t, i-1})$.
		\STATE Calculate $\widetilde{\mathbf{u}}^{t,i}$ and inexact error $\e_u^{t,i}$ by Eq. \eqref{Alg_Imp:implem_err}.
		\ENDWHILE
		\STATE $\mathbf{u}^{t}=\widetilde{\mathbf{u}}^{t,i}$. Output $\mathbf{u}^t$.
	\end{algorithmic}\label{Alg:iPAM2}
\end{algorithm}

\subsection{Task Embedded Strategies}

Owing to the diversities of realistic tasks, the univariate sub-problems in Eq. \eqref{ex_sequence} usually have distinct objective functions.
Taking their specificities into consideration, we introduce two targeted strategies, i.e., numerical algorithms and advanced techniques, for optimizing these univariate sub-problems.

\subsubsection{Numerical Algorithms Embedding}
Mostly, it makes no sense to employ extra numerical algorithms for optimizing sub-problems in Eq. \eqref{ex_sequence} if their explicit solutions are easily obtained.
However, for most cases with coupled $H$, it is common to adopt linearization for easy-to-solve sub-problems.
But as mentioned in the section of related work, the linearization skill requires estimating Lipschitz constants during every iteration, which brings a series of time-consuming troubles.

We have noticed that, though the objective function $\Psi(\z)$ is non-convex and non-smooth, sub-problems corresponding to specific tasks may possess nice traits, e.g, they are convex sometimes even differentiable problems, or they have unique solutions.
Thus there are plenty of high-efficient numerical algorithms designed for solving such univariate problems, e.g., greedy algorithms \cite{Elad2010Sparse}, PCG \cite{Spillane2016An}, FISTA \cite{Kim2018ANOTHER} and ADMM \cite{Wang2018Global}.
Thus it is advantaged to embed efficient algorithms in the CD scheme, for improving the efficiency and effectiveness of the whole algorithm, which is one of the motivations for proposing TECU.

\subsubsection{Advanced Techniques Embedding}
For real-world image processing tasks, most univariate sub-problems can be corresponding to specific applications.
For example, in tasks of image deblurring and super-resolution, one univariate sub-problem can be seen as image denoising task \cite{Chan2017Plug,Zhang2017Learning}.
While for low-light image enhancement, one of two sub-problems is for estimating the illumination layer, while the other one is for restoring the reflection image.

Except for optimization algorithms, there have been plenty advanced techniques such like BM3D \cite{Chan2017Plug} and variants of neural networks \cite{Zhang2017Learning,hasinoff2017Deep} for solving single-variable tasks.
Different from numerical algorithms, advanced techniques like neural networks, obtain the final results by a pre-trained propagations, rather than optimizing mathematical models.
Such advanced techniques are mostly quite efficient, meanwhile, are able to propagate variables very close to the desired solutions.
Taking these advantages into consideration, we propose to embed advanced techniques into CD scheme, to improve the convergence speeds of the whole algorithm, meanwhile, to get desired solutions with high probability.

Targeting to embed the above mentioned strategies into a unified framework, we adopt $K_x^t$ and $K_y^t$ steps of updates, starting at $\x^{t-1}$ and $\y^{t-1}$, for optimizing problems of Eq. \eqref{ex_sequence} to some extents, namely:
\begin{equation}\label{task_emb}
\begin{aligned}
&\x^{t,K^t_x} = \mathcal{A}_{x}^{K^t_x}\circ\mathcal{A}_{x}^{K^t_x-1}\cdots \mathcal{A}_{x}^1\circ\mathcal{A}_{x}^0(\x^{t-1}), \\
&\y^{t, K^t_y} = \mathcal{A}_{y}^{K^t_y}\circ\mathcal{A}_{y}^{K^t_y-1}\cdots \mathcal{A}_{y}^1\circ\mathcal{A}_{y}^0(\y^{t-1}),
\end{aligned}
\end{equation}
where $\circ$ denotes the composition operator.
Each $\mathcal{A}_u^{i}$ can be set as either one-step iteration of numerical algorithms or a propagation of advanced techniques.

This general framework covers various existing methods \cite{Yue2017Contrast,Schmidt2014Shrinkage,Zhang2017Learning}: e.g., for solvers like \cite{Yue2017Contrast}, each $\mathcal{A}_u^i$ can be seen as one-step iteration of numerical algorithms;  for others like \cite{Zhang2017Learning}, there exists only one step propagation of advanced techniques.
Besides, Eq. \eqref{task_emb} also includes cases that employ numerical algorithms and advanced techniques in hybrid manners, which is far more flexible than existing solvers.
Furthermore, considering two scenarios: 1) one problem of Eq. \eqref{ex_sequence} has closed-form solution; 2) not all the sub-problems have efficient numerical algorithms, thus we propose TECU in a hybrid updating scheme (as shown in Alg. \ref{Alg:iPAM1}) with other two classical CD updates (see Table \ref{updates}).

\begin{table*}[!t]
	\caption{Two classical CD updates for optimizing problem \eqref{eq:2variable}. Moreover, parameters should satisfy $\zeta_1^t>0$, $\zeta_2^t>0$, $\gamma_1^t>L_1^t$ and $\gamma_2^t>L_2^t$, where $L_1^t$ and $L_2^t$ are Lipschitz constants of $\nabla_{x}H(\mathbf{x},\y^t)$ and $\nabla_{y}H(\mathbf{x}^{t+1},\y)$.}\label{updates}
	\begin{center}
		\renewcommand\arraystretch{0.8}{
			\begin{tabular}{c|cc}
				\toprule
				&  Proximal update  & Prox-linear update \\
				\midrule
				$\x^{t+1}\in \arg\min_{\x}$    & $f(\x) + H(\x,\y^t) + \frac{\zeta_1^t}{2}\|\x-\x^t\|^2$
				& $f(\x) +\frac{\gamma_1^t}{2}\|\x-(\x^t-\nabla_{\x}H(\x^t, \y^t)/\gamma_1^t)\|^2$    \\
				$\y^{t+1}\in \arg\min_{\y}$  & $g(\y) + H(\x^{t+1},\y) + \frac{\zeta_2^t}{2}\|\y-\y^t\|^2$
				& $g(\y) + \frac{\gamma_2^t}{2}\|\y-(\y^t - \nabla_{\y}H(\x^{t+1},\y^t)/\gamma_2^t)\|^2$   \\
				\bottomrule
		\end{tabular}}
	\end{center}
\end{table*}


\subsection{Error Control and Estimation}

Notice that our task embedded strategies do not require exactly optimizing problems in Eq. \eqref{ex_sequence}: updating $\x^t$ and $\y^t$ by task embedded strategies bring errors $\e_x^{t}$ and $\e_y^{t}$ to the first-order optimality conditions of univariate sub-problems:
\begin{equation}\label{iPAM:iex_KKT}
\begin{aligned}
&\e_x^{t} = \mathbf{g}_x^{t} + \nabla_{\x} H(\x^{t},\y^{t-1}) + \eta_1(\x^{t}-\x^{t-1}),\\
&\e_y^{t} = \mathbf{g}_y^{t} + \nabla_{\y} H(\x^{t},\y^{t}) + \eta_2(\y^{t}-\y^{t-1}),
\end{aligned}
\end{equation}
where $\mathbf{g}_x^{t} \in \partial f(\x^{t})$ and $\mathbf{g}_y^{t} \in \partial g(\y^{t})$, with $\partial$ represents the Fr\'{e}chet limiting-subdifferential \cite{Attouch2010Proximal}.

Apparently, imprecise task embedded calculations certainly slow down the convergence speed of the whole algorithm, while over-precise optimizations are time-consuming and unnecessary for practical use.
Hence, we provide the following criterion to control the accuracies for optimizing univariate sub-problems.

\begin{criterion}\label{ecrit}
	The errors $\e_x^t$ and $\e_y^t$ brought by task embedded strategies should be controlled by certain constants, i.e.,
	\begin{equation}\label{Alg_Imp:err_crit}
	\|\e_x^{t}\| \leq C_x\epsilon_x^{t},  \quad  \quad \quad \|\e_y^{t}\| \leq C_y\epsilon_y^{t},
	\end{equation}
	with $\epsilon_x^{t} = \|\x^{t-1}-\x^{t-2}\|$ and $\epsilon_y^{t} = \|\y^{t-1}-\y^{t-2}\|$. Moreover, $0< 2C_x < \eta_1$ and $0< 2C_y <\eta_2$ should be satisfied.
\end{criterion}

Notice that the conditions in Eq. \eqref{Alg_Imp:err_crit} are certainly attainable for converged algorithms since the inaccurate errors $\e_x^{t}$ and $\e_y^{t}$ are approaching to zero when the algorithms are identified as converged.
However, since $\mathbf{g}_x^{t}$ and $\mathbf{g}_y^{t}$ should not be optionally selected from the sets of $\partial f$ and $\partial g$, we in Prop. \ref{prop} provide an implementation for estimating the errors, to make our proposed TECU more practical\footnote{All the proofs in this paper are presented in \cite{Wang2018Supp}}.
\begin{prop}\label{prop}
	Two intermediate variables $\widetilde{\x}^{t,K_x^t}$ and $\widetilde{\y}^{t,K_y^t}$ are calculated with respect to $\x^{t,K_x^t}$ and $\y^{t,K_y^t}$ as follows\footnote{$\mathrm{prox}^{\tau}_{\sigma}(\x)\in\arg\min_{\z}\sigma(\z)+\frac{\tau}{2}\|\z-\x\|^2$ denotes proximal mapping to proper, l.s.c function $\sigma$.}:
	\begin{equation}\label{Alg_Imp:extra_variables}
	\begin{aligned}
	&\widetilde{\x}^{t,K_x^t}=\mathrm{prox}^{1}_{f}(\eta_1\x^{t-1} + \mathcal{P}_x^{t-1}(\x^{t,K_x^t})), \\ &\widetilde{\y}^{t,K_y^t}=\mathrm{prox}^{1}_{g}(\eta_2\y^{t-1} + \mathcal{P}_y^{t-1}(\y^{t,K_y^t})),
	\end{aligned}
	\end{equation}
	with functions $\mathcal{P}_x^{t-1}(\mathbf{u}) = (1-\eta_1)\mathbf{u} - \nabla_{\x}H(\mathbf{u},\y^{t-1})$ and $\mathcal{P}_y^{t-1}(\mathbf{u}) = (1-\eta_2)\mathbf{u} - \nabla_{\y}H(\x^t, \mathbf{u})$.
	Then, the following
	\begin{equation}\label{Alg_Imp:implem_err}
	\begin{aligned}
	\e_x^{t,K_x^t}=& \mathcal{P}_x^{t-1}(\x^{t,K_x^t}) - \mathcal{P}_x^{t-1}(\widetilde{\x}^{t,K_x^t}),\\
	\e_y^{t,K_y^t}=& \mathcal{P}_y^{t-1}(\y^{t,K_y^t}) - \mathcal{P}_y^{t-1}(\widetilde{\y}^{t,K_y^t}),\\
	\end{aligned}
	\end{equation}
	are implementations of $\e_x^{t}$ and $\e_y^{t}$ in Eq. \eqref{iPAM:iex_KKT}, by assigning $\widetilde{\x}^{t,K_x^t}$ to $\x^{t}$ and $\widetilde{\y}^{t,K_y^t}$ to $\y^{t}$.
\end{prop}

So far, we have introduced the whole process of TECU and further provide its detailed procedures in Alg. \ref{Alg:iPAM2}.
As follows, we will demonstrate that our error control conditions are more persuasive than previously used criteria \cite{li2014splitting,Yue2017Contrast}, since it is well converged in theory.

\section{Theoretical Analyses}\label{sec:converge}
With properties of the objective function and the error control criterion, we in this section present some nice properties of TECU.
Firstly, we demonstrate that along with the iteration progresses of TECU, there exists a bounded function that satisfies sufficient-descent property.

\begin{prop}\label{Converg_Ana:main_theorem}
	Suppose that the sequence $\{\z^t\}_{t\in \mathbb{N}}$ is generated by TECU, then with Criterion \ref{ecrit}, there exist a bounded function $\Phi(\z, \w)$, such that for $\forall t \geq 0$:
	\begin{subequations}\label{suff_subound}
		\begin{align}
		&\Phi(\z^t, \z^{t-1}) - \Phi(\z^{t+1}, \z^t)\geq a \|\z^{t+1}-\z^t\|^2,\\
		&\mbox{dist}(0, \partial \Phi(\z^t, \z^{t-1})) \leq b(\|\z^t - \z^{t-1}\| + \|\z^{t-1}-\z^{t-2}\|),
		\end{align}
	\end{subequations}
	with definite positive constants $a$ and $b$.
	In addition, the sequence $\{\z^t\}_{t\in \mathbb{N}}$ generated by TECU is bounded.
\end{prop}

Since our proposed TECU adopts a hybrid update scheme (presented in Alg. \ref{Alg:iPAM1}), thus the $\Phi(\z, \w)$, $a$ and $b$ have different concrete formations, corresponding to distinct combinations of updates \cite{Wang2018Supp}.
Then with this proposition, we are ready to illuminate the property of the final solution optimized by TECU.

\begin{theorem}\label{Converg_Ana:key_theorem}
	The $\{\z^t\}_{t\in \mathbb{N}}$ generated by TECU is a Cauchy sequence, which converges to a critical point $\z^{\ast}:=(\x^{\ast},\y^{\ast})$ of the original objective function $\Psi(\z)$.
\end{theorem}

The Theorem \ref{Converg_Ana:key_theorem} presents the property of the final optimized solution, meanwhile, demonstrates the robust performances of our proposed algorithm framework.
Moreover, TECU has at least sub-linear convergence rate when the desingularising function $\phi (s)=\frac{C}{\theta}s^{\theta}:[0,\mu)\to \mathbb{R}_+$ of the objective $\Psi(\z)$ is satisfied with positive constant $C$ and $\theta \in [0, 1).$
Better yet, it will further have linear convergence rate if $\theta \in [\frac{1}{2}, 1)$.
Though this convergence rate is accordance with previous CD schemes like \cite{Attouch2010Proximal}, our TECU is far more efficient and effective for realistic tasks.

\begin{table*}[!t]
	\caption{The comparison results on iteration number and whole computation time (s). The data dimensions $(n, m, p)$ are set as $(64, 600, 4000)$, $(144, 900, 10000)$ and $(256, 1600, 16000)$ respectively. }\label{Tab:Synthetic}
	\begin{center}\renewcommand\arraystretch{0.7}{
			\begin{tabular}{c|ccccc|ccccc}\toprule
				&\multicolumn{5}{c|}{Iteration number / Total propagation steps} &  \multicolumn{5}{c}{Computation time (s)}\\ \midrule
				Algorithms &PALM  &   INV &   BCU  &   iPALM   & TECU & PALM  &  INV &   BCU &  iPALM &   TECU    \\\midrule
				$n=64$&21   &  109  &  19    &  21   &  \textbf{12} / 54   &  4.45  &  16.20   &  3.87  &  4.68    &  \textbf{2.22}    \\
				$n=144$&21   &  35  &  18    &  21   &  \textbf{14} / 152  &  17.16  &  21.40   &  14.45  &  18.01    &  \textbf{11.96}    \\
				$n=256$&22   &  39  &  18    &  22   &  \textbf{17} / 153   &  121.47  &  137.27   &  100.19  &  124.67    &  \textbf{75.19}    \\\bottomrule
		\end{tabular}}
	\end{center}
\end{table*}

\begin{table*}[!t]
	\vspace{-0.4cm}
	\caption{Computation time (s) of one-step iteration of CD algorithms, and one-step propagation of TECU on different data scales. }\label{Tab:time}
	\begin{center}\renewcommand\arraystretch{0.5}{
			\begin{tabular}{c@{\extracolsep{0.5em}}c@{\extracolsep{0.5em}}c@{\extracolsep{0.5em}}c@{\extracolsep{0.5em}}c|@{\extracolsep{0.5em}}c@{\extracolsep{0.5em}}c@{\extracolsep{0.5em}}c@{\extracolsep{0.5em}}c@{\extracolsep{0.5em}}c|@{\extracolsep{0.5em}}c@{\extracolsep{0.5em}}c@{\extracolsep{0.5em}}c@{\extracolsep{0.5em}}c@{\extracolsep{0.5em}}c@{\extracolsep{0.5em}}}\toprule
				\multicolumn{5}{c|}{$n=64$} &  \multicolumn{5}{c|}{$n=144$} &\multicolumn{5}{c}{$n=256$}\\ \midrule
				PALM  &   INV &   BCU  &   iPALM   & TECU & PALM  &  INV &   BCU &  iPALM &   TECU &PALM  &   INV &   BCU  &   iPALM   & TECU   \\\midrule
				0.21   &  0.15  &  0.20    &  0.22   &  \textbf{0.04}   &  0.82  &  0.61   &  0.80  &  0.86    &  \textbf{0.08}  & 5.52  &  3.52   &  5.57  &  5.67    &  \textbf{0.49} \\\bottomrule
		\end{tabular}}
	\end{center}
\end{table*}

\begin{figure*}[!t]
	\setlength{\abovecaptionskip}{0cm} 
	\setlength{\belowcaptionskip}{0cm} 
	\hspace{0.5cm}
	\begin{minipage}[b]{0.2\linewidth}
		\centering
		\centerline{\includegraphics[width=1.1\textwidth]{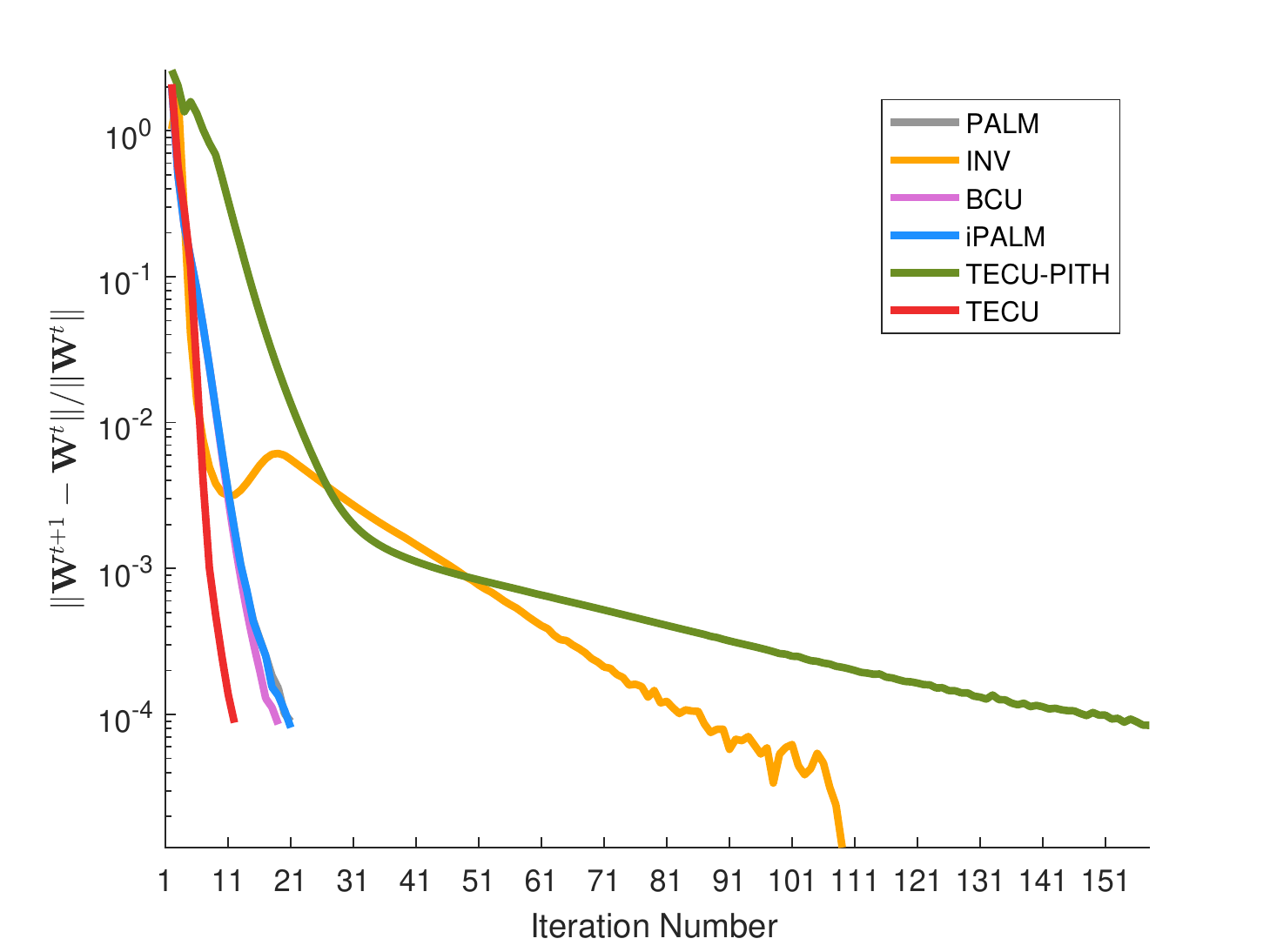}}
		\vspace{-0.6cm}
	\end{minipage}
	\hspace{-0.2cm}
	\hfill
	\begin{minipage}[b]{0.2\linewidth}
		\centering
		\centerline{\includegraphics[width=1.1\textwidth]{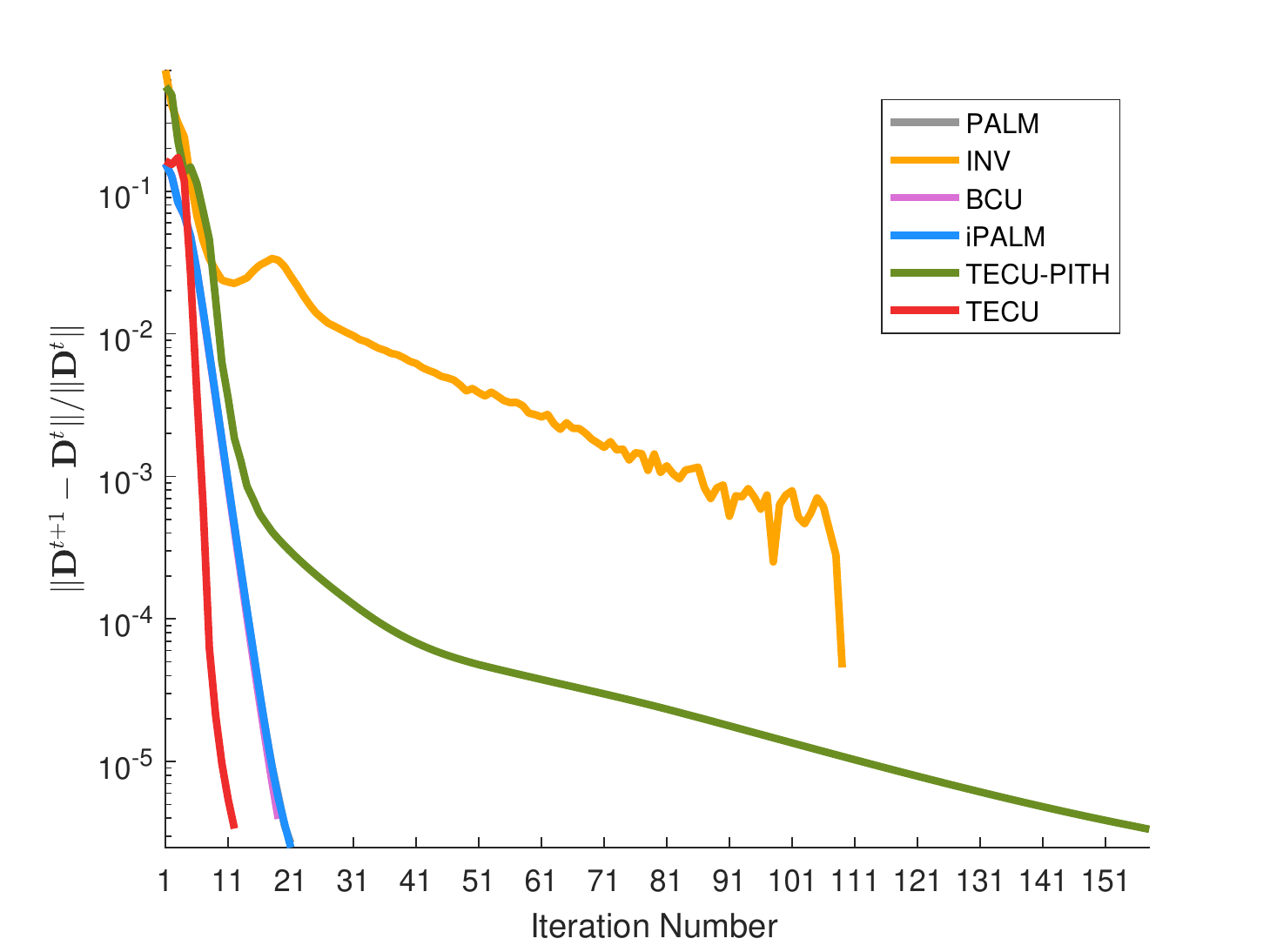}}
		\vspace{-0.6cm}
	\end{minipage}
	\hspace{-0.2cm}
	\hfill
	\begin{minipage}[b]{0.2\linewidth}
		\centering
		\centerline{\includegraphics[width=1.1\textwidth]{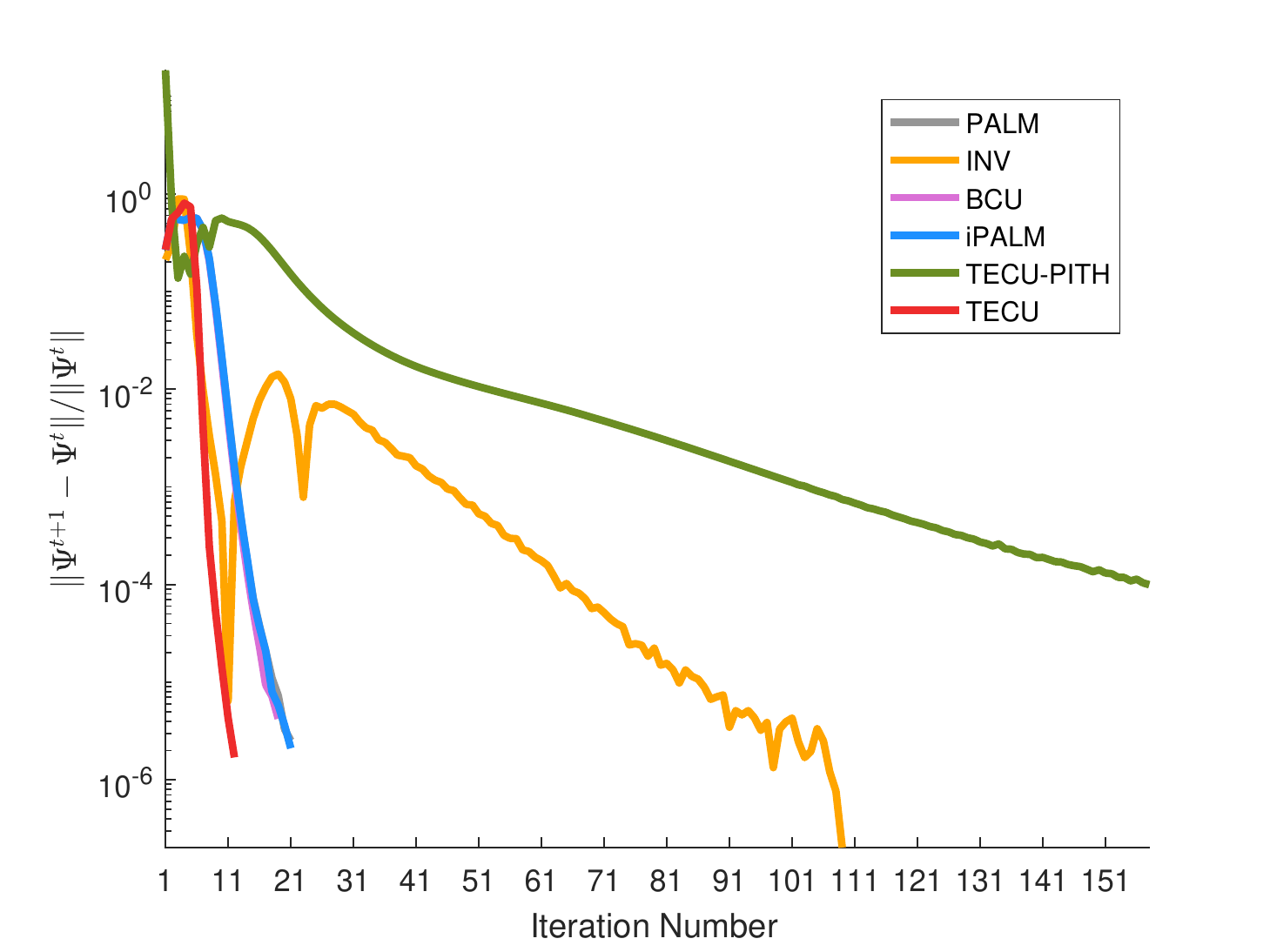}}
		\vspace{-0.6cm}
	\end{minipage}
	\hspace{-0.2cm}
	\hfill
	\begin{minipage}[b]{0.2\linewidth}
		\centering
		\centerline{\includegraphics[width=1.1\textwidth]{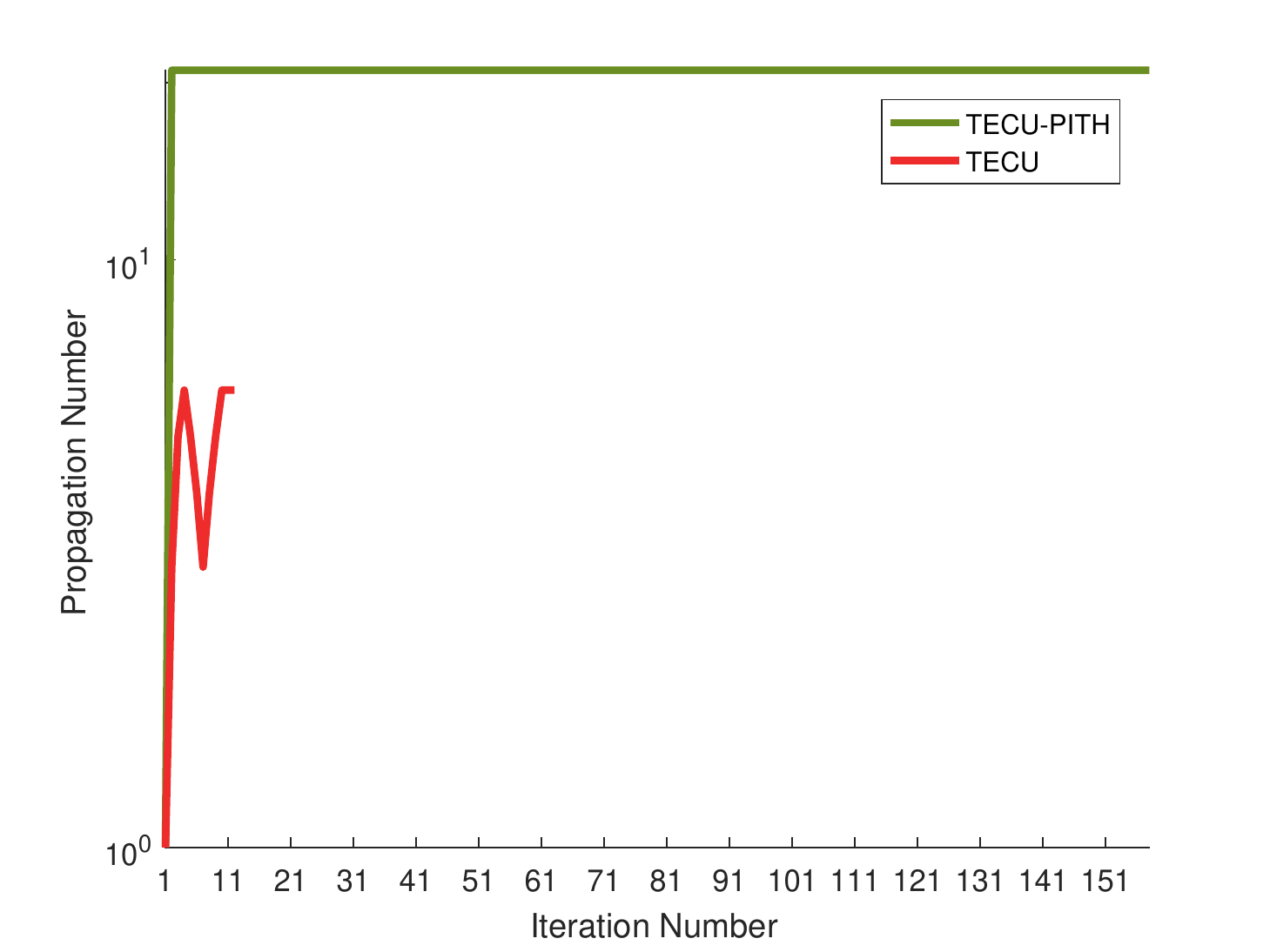}}
		\vspace{-0.6cm}
	\end{minipage}
\hspace{0.5cm}
\end{figure*}
\begin{figure*}[!t]
	\setlength{\abovecaptionskip}{0cm} 
	\setlength{\belowcaptionskip}{0cm} 
	\hspace{0.5cm}
	\begin{minipage}[b]{0.2\linewidth}
		\centering
		\centerline{\includegraphics[width=1.1\textwidth]{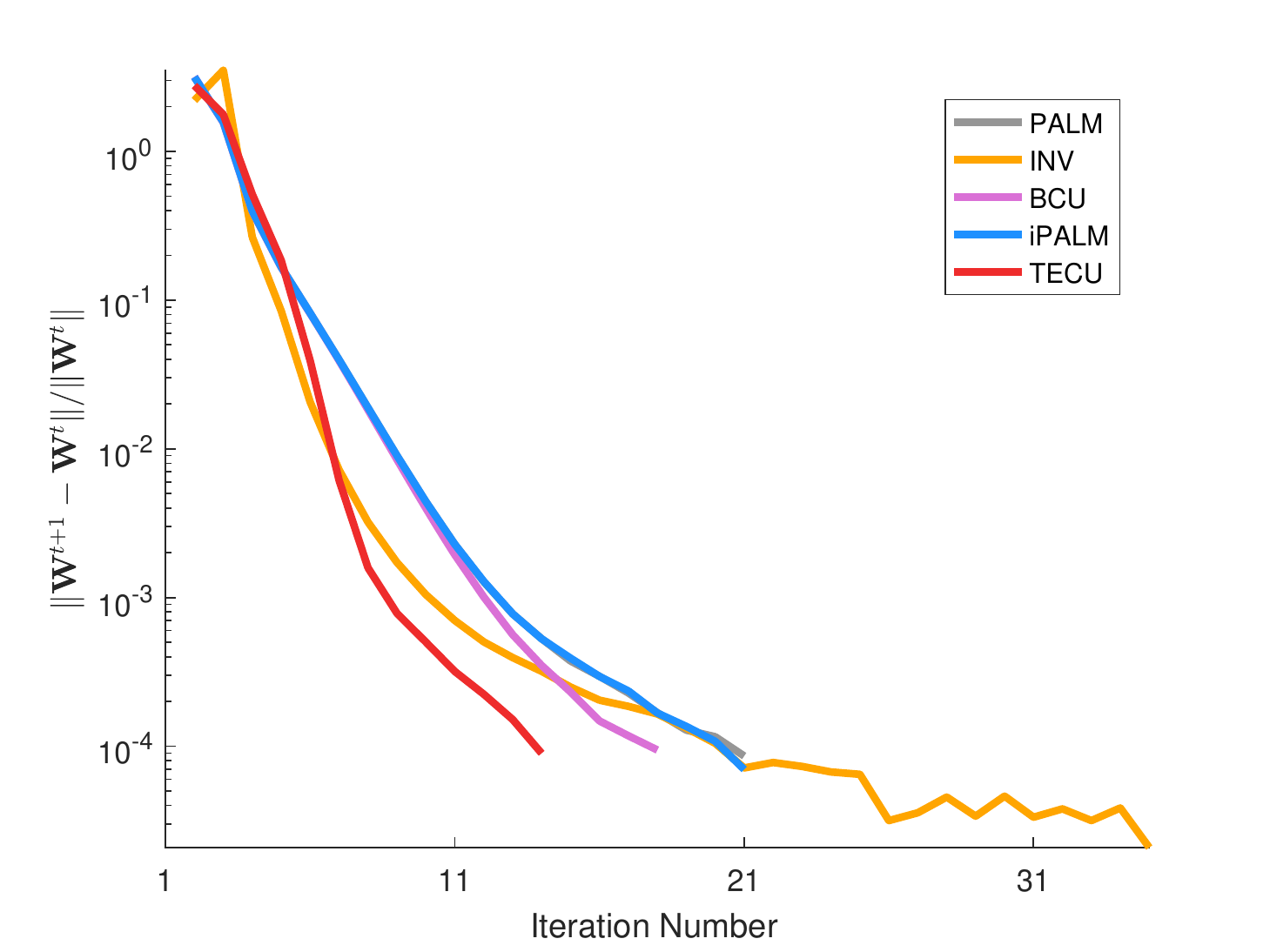}}
		\centerline{\quad \ (a) $\|\W^{t+1}-\W^t\|/\|\W^t\|$}\medskip
	\end{minipage}
	\hspace{-0.5cm}
	\hfill
	\begin{minipage}[b]{0.2\linewidth}
		\centering
		\centerline{\includegraphics[width=1.1\textwidth]{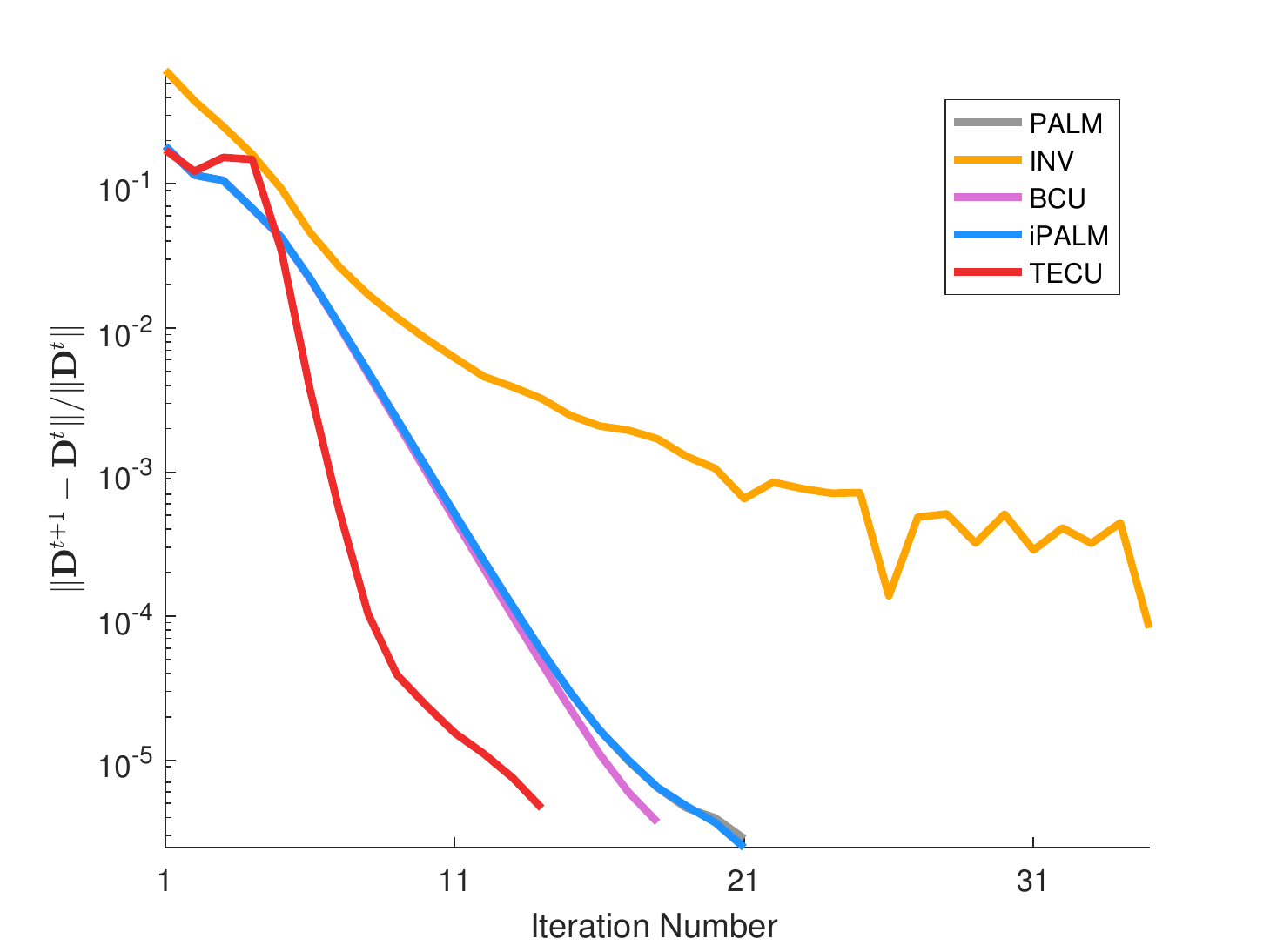}}
		\centerline{\quad \  (b) $\|\D^{t+1}-\D^t\|/\|\D^t\|$}\medskip
	\end{minipage}
	\hspace{-0.5cm}
	\hfill
	\begin{minipage}[b]{0.2\linewidth}
		\centering
		\centerline{\includegraphics[width=1.1\textwidth]{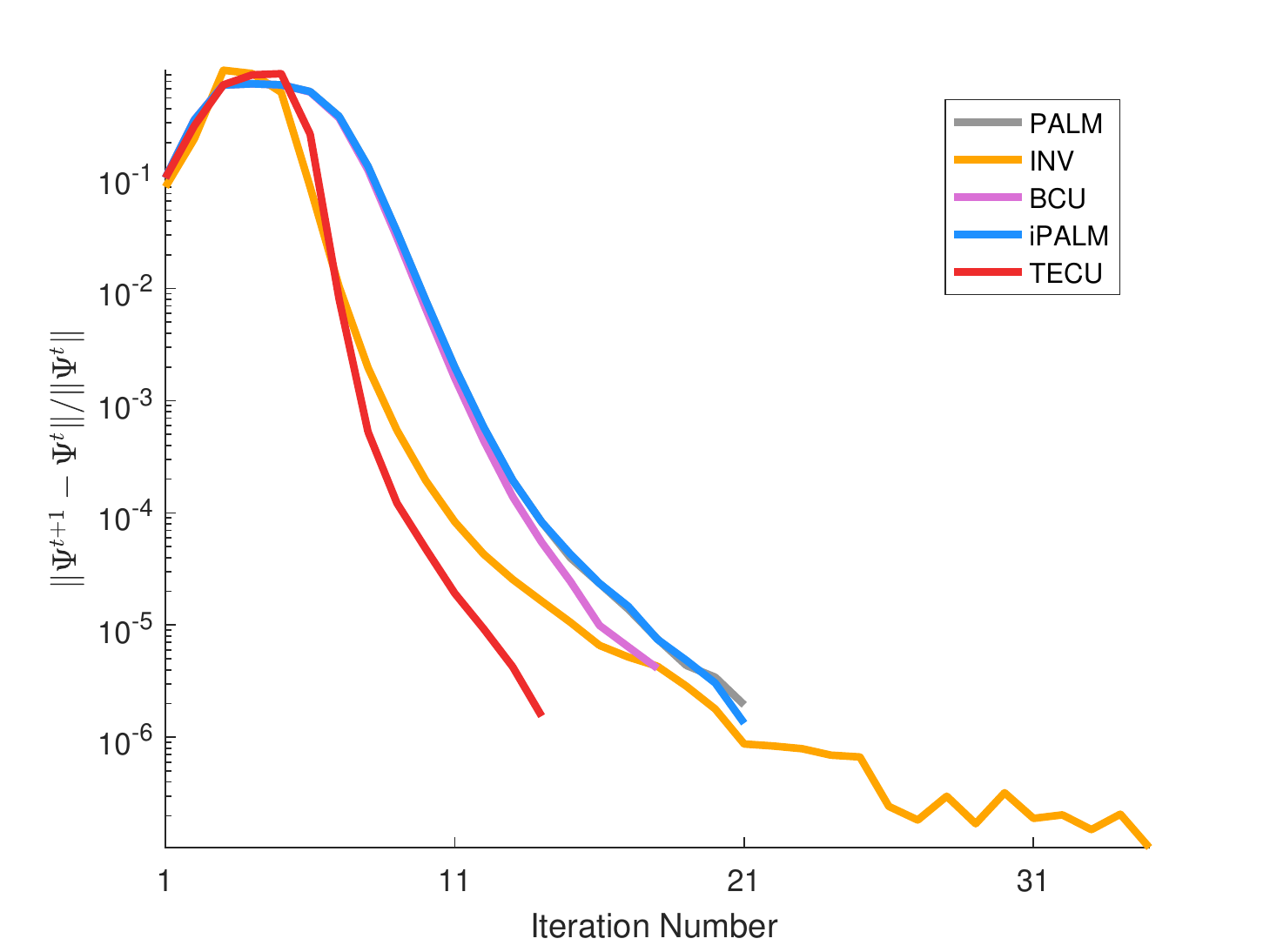}}
		\centerline{\quad \  (c) $\|\Psi^{t+1}-\Psi^t\|/\|\Psi^t\|$}\medskip
	\end{minipage}
	\hspace{-0.5cm}
	\hfill
	\begin{minipage}[b]{0.2\linewidth}
		\centering
		\centerline{\includegraphics[width=1.1\textwidth]{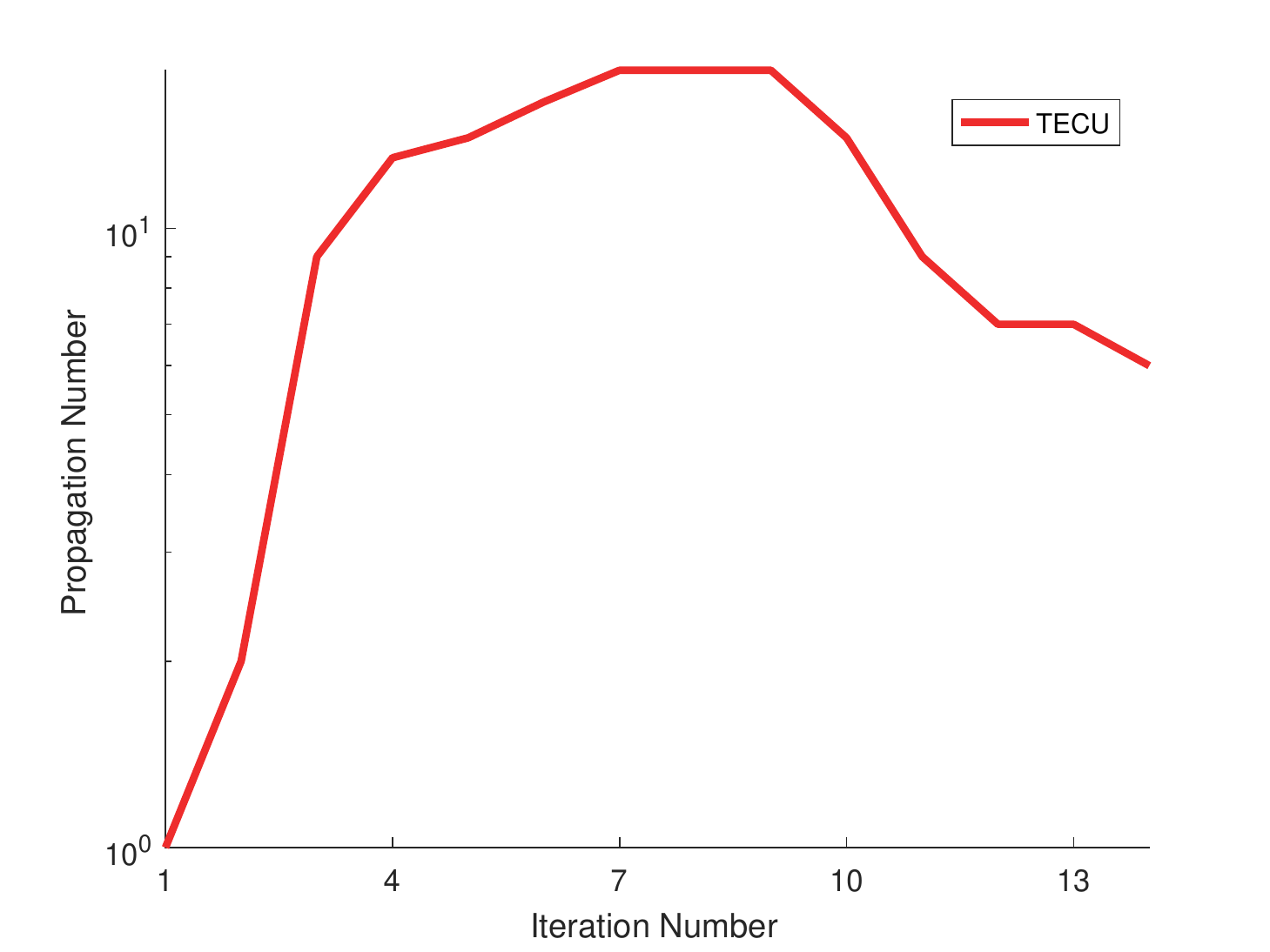}}
		\centerline{\quad  (d) Propagation Number}\medskip
	\end{minipage}
    \hspace{0.5cm}
	\caption{The compared convergence property of PALM, INV, BCU, iPALM and our TECU. The results in the first row belongs to the data with $n=64$, while the comparison results in the second row belongs to $n=144$.}\label{fig:144}
\end{figure*}

\section{TECU for Realistic Tasks}
As previously stated, TECU allows embedding both numerical algorithms and advance techniques in its algorithm framework.
Hence, we consider two realistic tasks, $\ell_0$-regularized dictionary learning ($\ell_0$DL) and low-light image enhancement (LIE), to verify both efficiency and effectivenss of embedding task-oriented strategies.

\subsection{$\ell_0$-Regularized Dictionary Learning}
Dictionary learning is a powerful tool for learning features from data.
Its basic idea is to factorize the data $\Y$ as $\D\W^{\top}$, where $\D$ is the dictionary and $\W$ is the corresponding coefficients.
We consider the previously proposed $\ell_0$DL problem \cite{bao2016dictionary}, which can be modeled as
\begin{equation}\label{Exp:DL_model}
\min_{\W,\D}  \lambda\|\W\|_0  + \mathcal{X}_{\mathcal{D}}(\D) + \frac{1}{2}\|\Y - \D\W^{\top}\|^2,
\end{equation}
where $\|\cdot\|_0$ denotes the $\ell_0$ penalty that counts the number of non-zero elements, and indicator function $\mathcal{X}$ acts on the set
$\mathcal{D}=\{\D=\{\bd_i\}_{i=1}^m: \|\bd_i\|=1, \forall i\}$ for normalized bases.

Notice that, the two sub-problems of $\ell_0$DL have entirely different characteristics.
The sub-problem of $\W$ is a sparse coding task with $\ell_0$ penalty, which can be optimized by a proximal iterative hard-thresholding (PITH) method \cite{Bach2011Optimization}.
While the sub-problem of $\D$ minimizes a strongly convex quadratic function with unit ball constraint, thus it can be efficiently solved.
Considering this nice property, we embed ADMM for optimizing $\D$ sub-problem, and the experimental results verify that embedding ADMM improve the efficiency of optimization.
However, due to the non-convexity of the sub-problem of $\W$, additional experiments are conducted to show that embedding PITH is not a good choice for TECU, which also indicates the necessity of hybrid scheme.

Rather than TECU, the $\ell_0$DL problem can also be optimized by other numerical algorithms.
However, as repeatedly stated, though the update of PALM can be simply computed, its request on estimating Lipschitz constants decelerate the overall convergence speeds, especially for large-scale data (see Table \ref{Tab:time}).
Except for PALM, its two variants, i.e., BCU \cite{Xu2017A} and iPALM \cite{Pock2017Inertial} are also employed for solving the $\ell_0$DL problem.
But these variants do not avoid estimating Lipschitz constants, even worse, their efficiencies are restricted by additional parameter conditions.
To avoid these troubles, \cite{bao2016dictionary} utilize a strategy (named INV), which solves the sub-problem of $\D$ by
leaving out term $\mathcal{X}_{\mathcal{D}}(\mathbf{D})$ first, and then directly projecting the results on $\mathcal{X}_{\mathcal{D}}$.
However, we have noticed that the update of INV is inaccurate without theoretical supports,
thus its performances are always lack of robustness (see Fig. \ref{fig:144}).

\begin{figure*}[t]
	\setlength{\abovecaptionskip}{0cm} 
	\setlength{\belowcaptionskip}{0cm} 
	\hspace{0.5cm}
	\begin{minipage}[b]{0.16\linewidth}
		\centering
		\centerline{\includegraphics[width=1\textwidth]{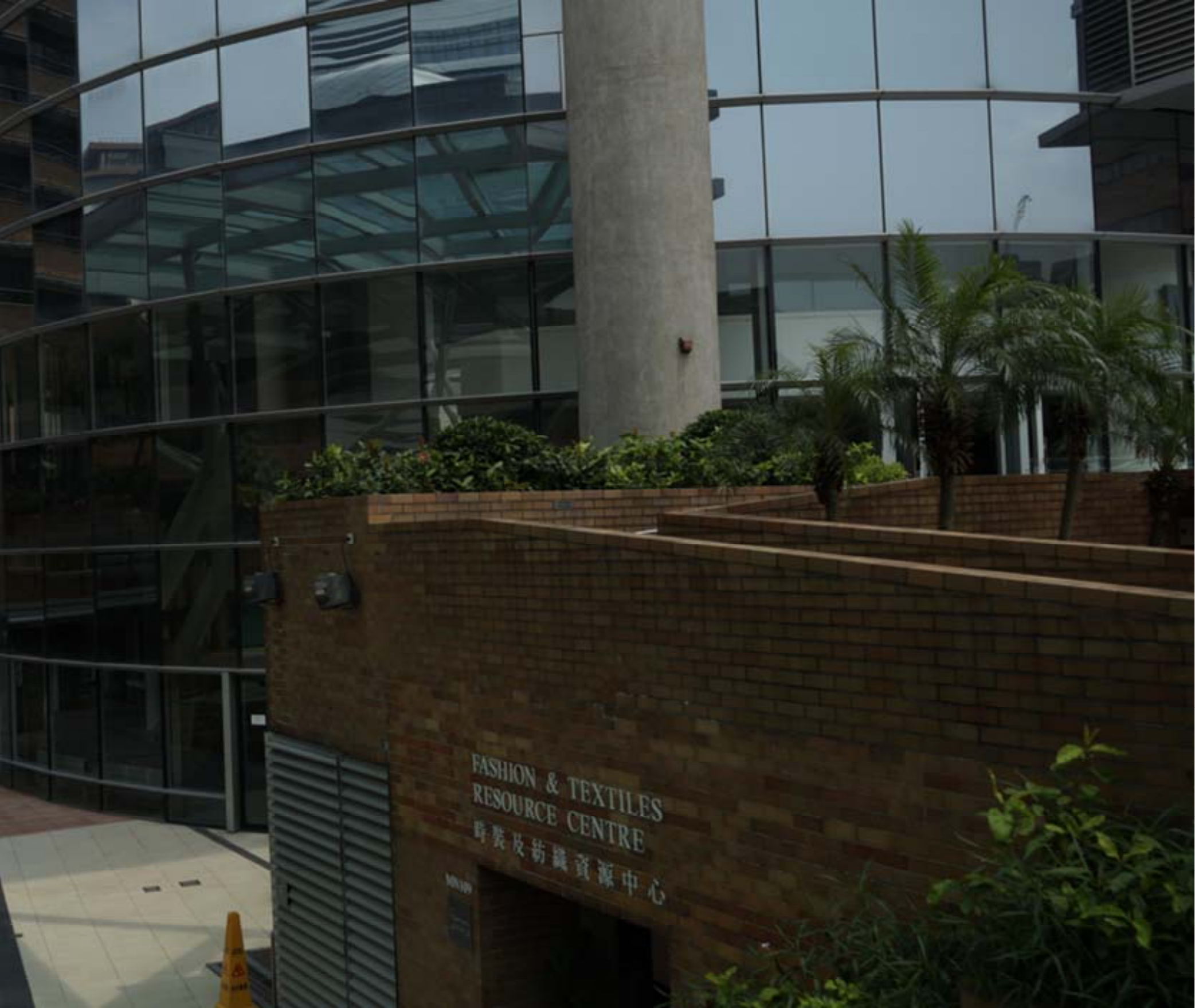}}
		\vspace{0.3cm}
		\centerline{(a) Input}\medskip
	\end{minipage}
	\hspace{-0.2cm}
	\hfill
	\begin{minipage}[b]{0.16\linewidth}
		\centering
		\centerline{\includegraphics[width=1\textwidth]{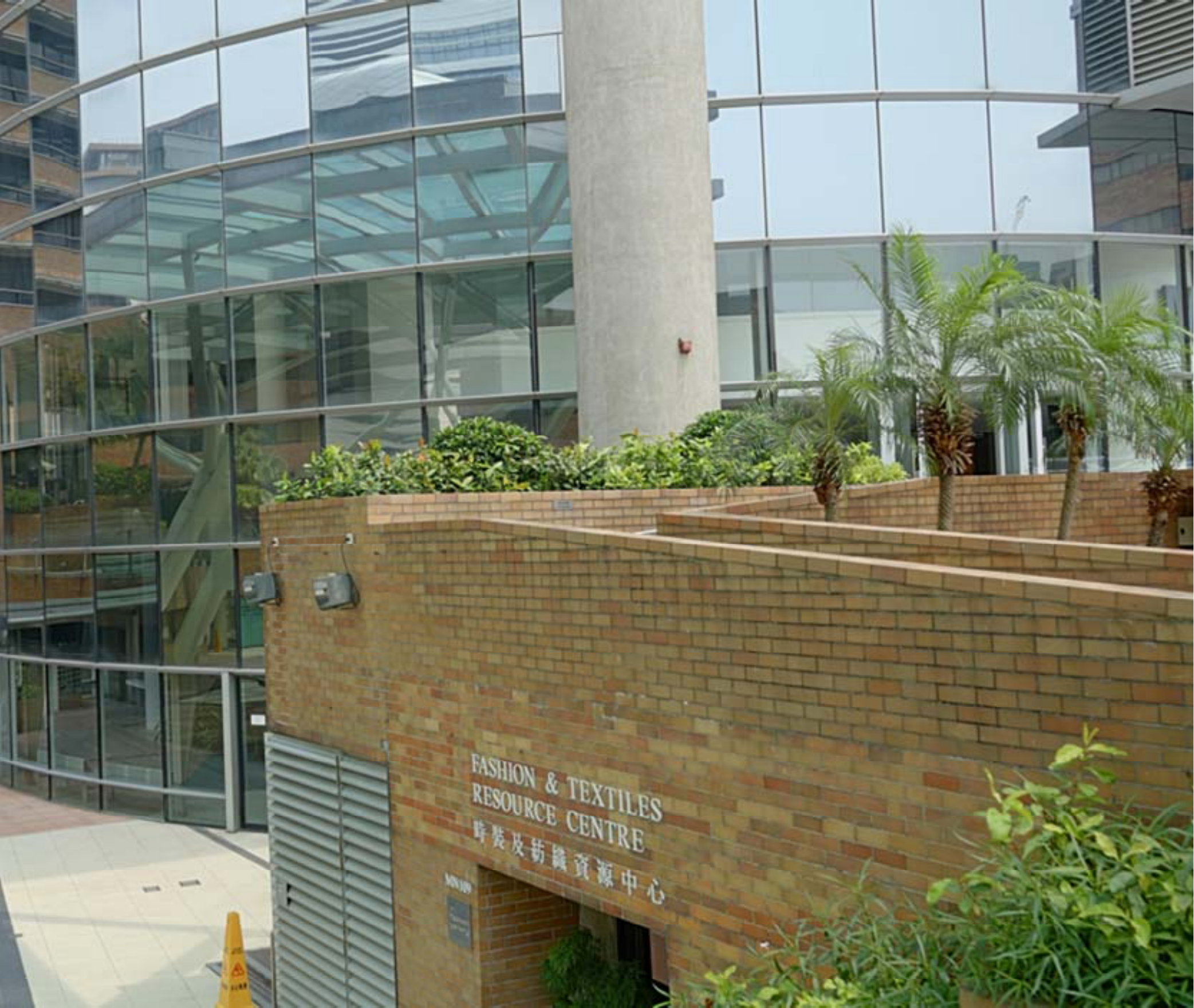}}
		\vspace{0.3cm}
		\centerline{ (b) Reference}\medskip
	\end{minipage}
	\hspace{-0.2cm}
	\hfill
	\begin{minipage}[b]{0.16\linewidth}
		\centering
		\centerline{\includegraphics[width=1\textwidth]{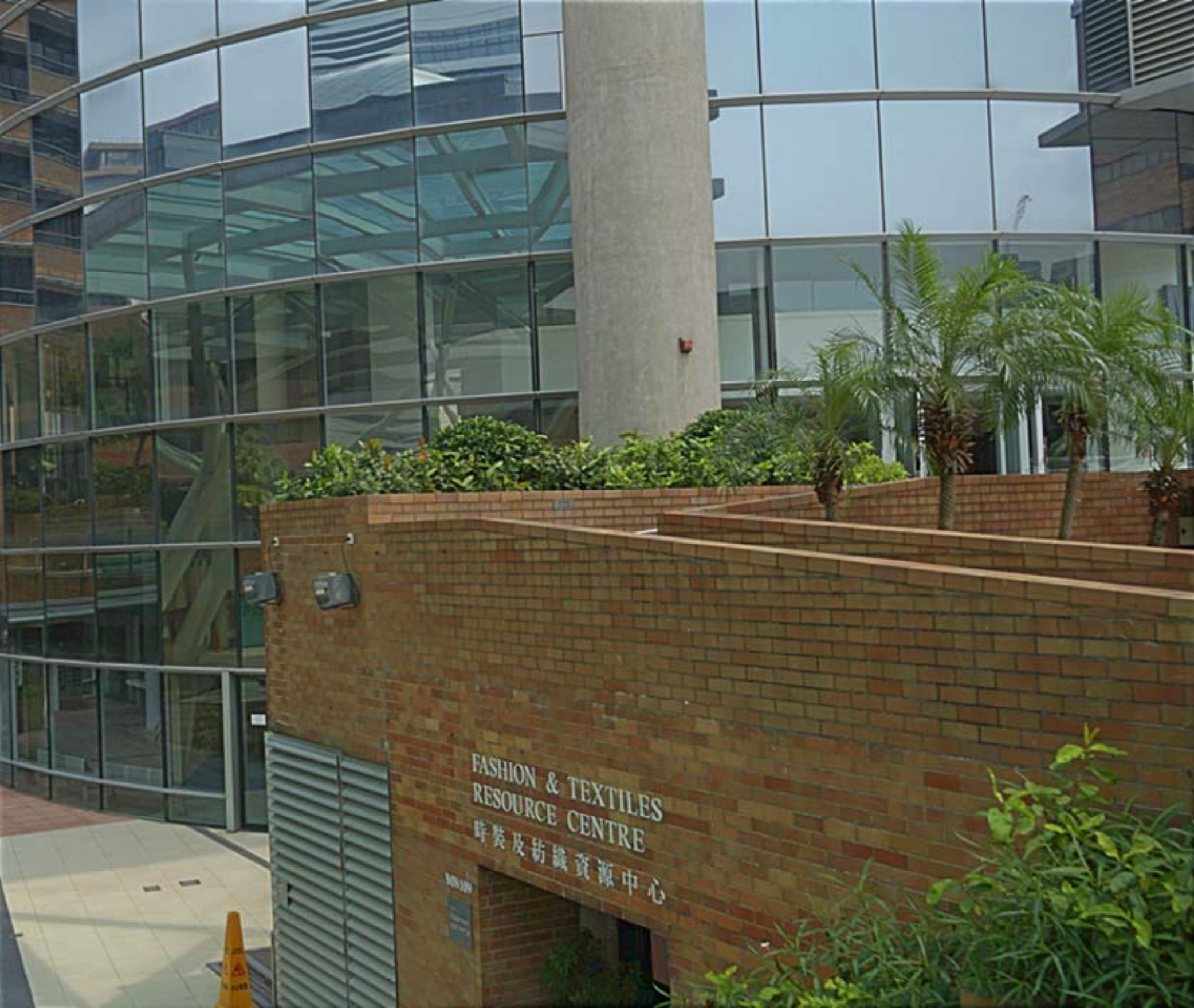}}
		\vspace{0.3cm}
		\centerline{ (c) Classical CD / 22.31 }\medskip
	\end{minipage}
	\hspace{-0.2cm}
	\hfill
	\begin{minipage}[b]{0.16\linewidth}
		\centering
		\centerline{\includegraphics[width=1\textwidth]{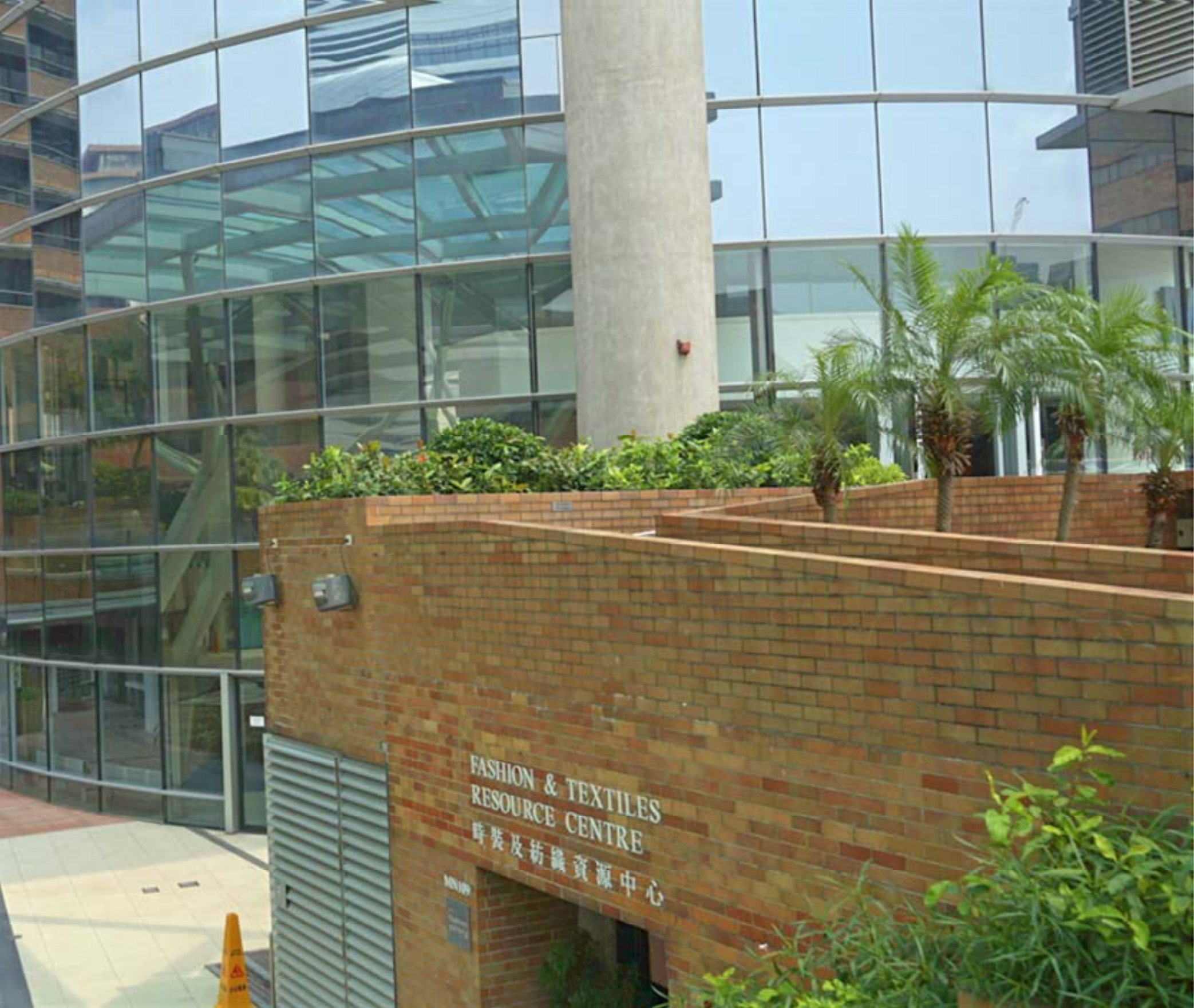}}
		\vspace{0.3cm}
		\centerline{ (d) TECU / \textbf{28.00}}\medskip
	\end{minipage}
	\hspace{-0.3cm}
	\hfill
	\begin{minipage}[b]{0.195\linewidth}
		\centering
		\centerline{\includegraphics[width=1\textwidth]{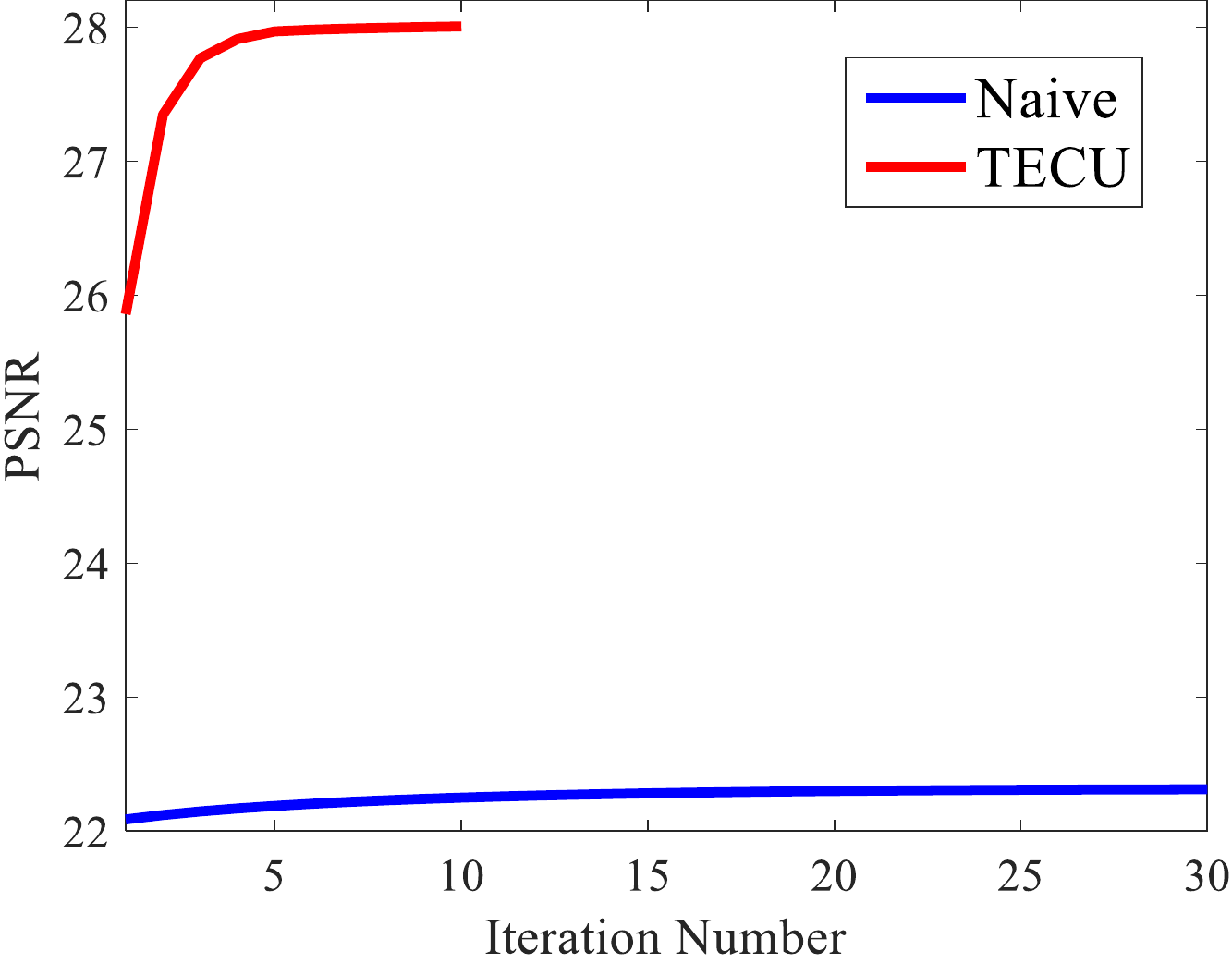}}
		\centerline{ (e) PSNR curve}\medskip
	\end{minipage}
\hspace{0.5cm}
	\caption{Comparisons between the classical CD algorithm, i.e., PAM \cite{Attouch2010Proximal} and TECU. The PSNR scores are reported below (c) and (d) sub-figures. The PSNR curves with respect to iteration number is plotted in sub-figure (e).}\label{fig:analysis}
\end{figure*}

\begin{table}[t]
	\vspace{-0.2cm}
	\centering
	\caption{Quantitative evaluations of NIQE (the lower, the better) on the NASA and Non-uniform dataset. }
	\vspace{0.1cm}
	\renewcommand\arraystretch{0.8}{
		\begin{tabular}{c|@{\extracolsep{0.4em}}c@{\extracolsep{0.4em}}c@{\extracolsep{0.4em}}c@{\extracolsep{0.4em}}c@{\extracolsep{0.4em}}c@{\extracolsep{0.4em}}c@{\extracolsep{0.4em}}c@{\extracolsep{0.4em}}c@{\extracolsep{0.4em}}c@{\extracolsep{0.4em}}c}
			\toprule
			Dataset&HE&BPDHE&MSRCR&GOLW&NPEA\\
			\midrule
			NASA&3.68&3.77&3.66&3.74&{{3.43}}\\
			Non-uniform&4.28&3.09&3.04&2.99&2.99\\
			\midrule
			Dataset&SRIE&WVM&JIEP&HDRNet&TECU\\
			\midrule
			NASA&3.97&3.86&3.72&3.72&\textbf{{3.41}}\\
			Non-uniform&3.02&{{2.94}}&2.97&3.11&\textbf{{2.91}}\\
			\bottomrule
	\end{tabular}}
	\label{tab:rescomp}
\end{table}

\subsection{Low-light Image Enhancement}
The purpose of LIE is to enhance the captured low-visibility images so that high-quality images can be obtained.
For LIE, Retinex-based decomposition~\cite{cai2017joint,fu2016weighted} is widely concerned:
$
\mathbf{O} = \mathbf{I}\odot \mathbf{R}$,
with element-wise multiplication operator $\odot$.
Thus LIE is to factorize the observed low-light image $\mathbf{O}$ into an illumination layer $\mathbf{I}$ that represents the light intensity, and a reflection layer $\mathbf{R}$ which describes the physical characteristic of objects.

Considering the characteristics of illumination layer, many literatures \cite{fu2015probabilistic,cai2017joint} enforce a smooth constraint, i.e., $\|\nabla \mathbf{I}\|^2$ to represent the smooth changes of the illumination layer.
Then together with the range constraints of both layers, we establish the following optimization model for LIE task:
\begin{equation}\label{LIE}
\min_{\mathbf{I}, \mathbf{R}} \frac{\alpha}{2}\|\nabla \mathbf{I}\|^2  + \mathcal{X}_{\mathcal{I}}(\mathbf{I}) + \mathcal{X}_{\mathcal{R}}(\mathbf{R}) + \frac{1}{2}\|\mathbf{O} - \mathbf{I}\odot \mathbf{R}\|^2,
\end{equation}
with $\mathcal{I}=\{\mathbf{I}: \mathbf{I}_{i} \in [0, \mathbf{O}_{i}], \forall i\}$, $\mathcal{R}=\{\mathbf{R}: \mathbf{R}_{i}\in [0, 1], \forall i\}$.

Employing TECU for optimizing it, we embed a residual-type CNN to propagate the illumination layer very close to the desired solution.
Specifically, we first randomly choose 800 image-pairs from ImageNet database \cite{Krizhevsky2012ImageNet} and crop them into $35 \times 35$ small patches, to train a neural network with only one residual block including 7 convolutional layers and ReLU activations.
Then at each iteration, we first use the pre-trained network to propagate the latest value of $\mathbf{I}$.
Then by considering that the pre-trained network may not always satisfy the error control conditions, thus we further employ prox-linear updates as the remaining propagations, until the Criterion \ref{ecrit} is satisfied.
For the other sub-problem of reflection layer, we adopt proximal update to get the closed-form solution.

\section{Experimental Results} \label{Sec:Experiments}

With the task embedded strategies respectively designed for $\ell_0$DL and LIE problems, we apply TECU to
$\ell_0$DL with synthetic data to verify its nice convergence properties.
While for LIE task, TECU is applied to real-world images.
The experimental results of both two realistic tasks demonstrate the efficiency and effectiveness of embedding strategies into CD algorithm, by comparing with other state-of-the-art methods.

\begin{figure*}[t]
	\centering
	\hspace{-0.3cm}
	\begin{tabular}{c@{\extracolsep{0.4em}}c@{\extracolsep{0.4em}}c@{\extracolsep{0.4em}}c@{\extracolsep{0.4em}}c@{\extracolsep{0.4em}}c}
		\includegraphics[width=.15\linewidth]{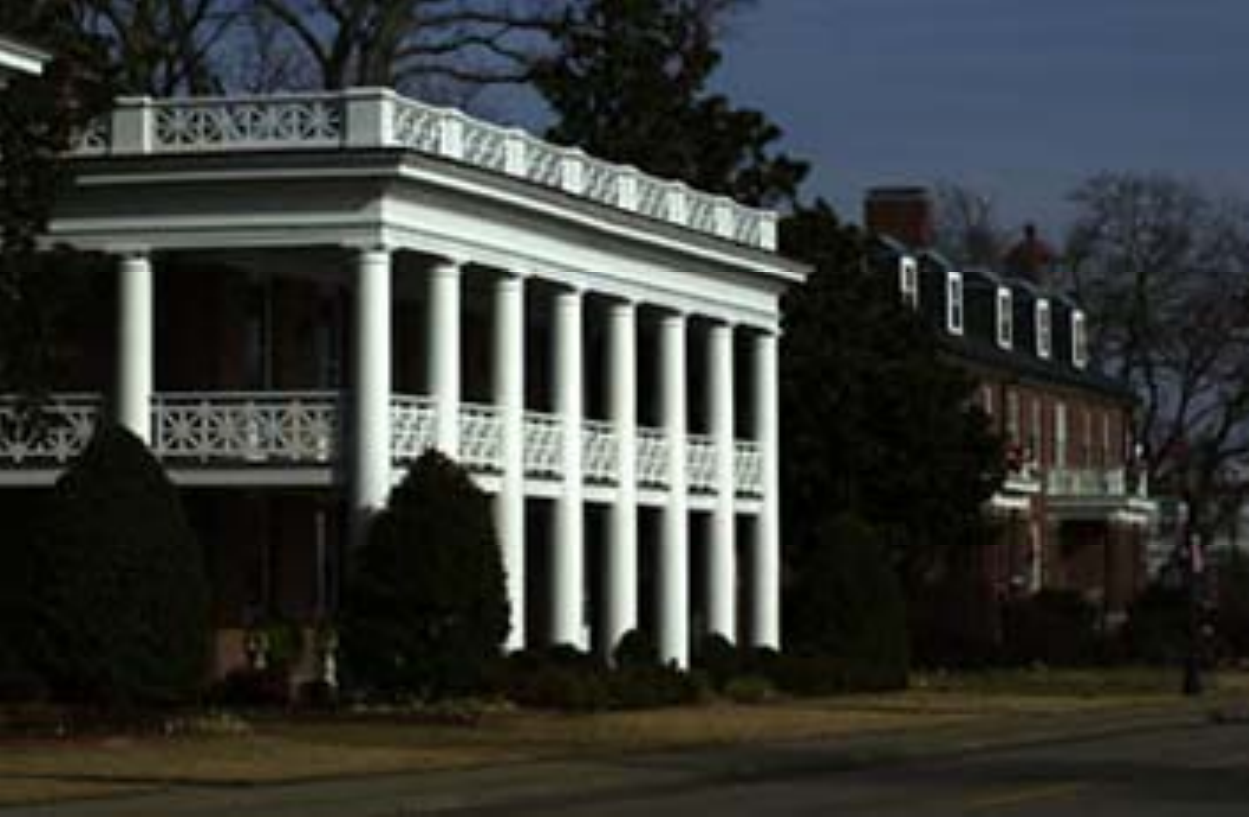}&
		\includegraphics[width=.15\linewidth]{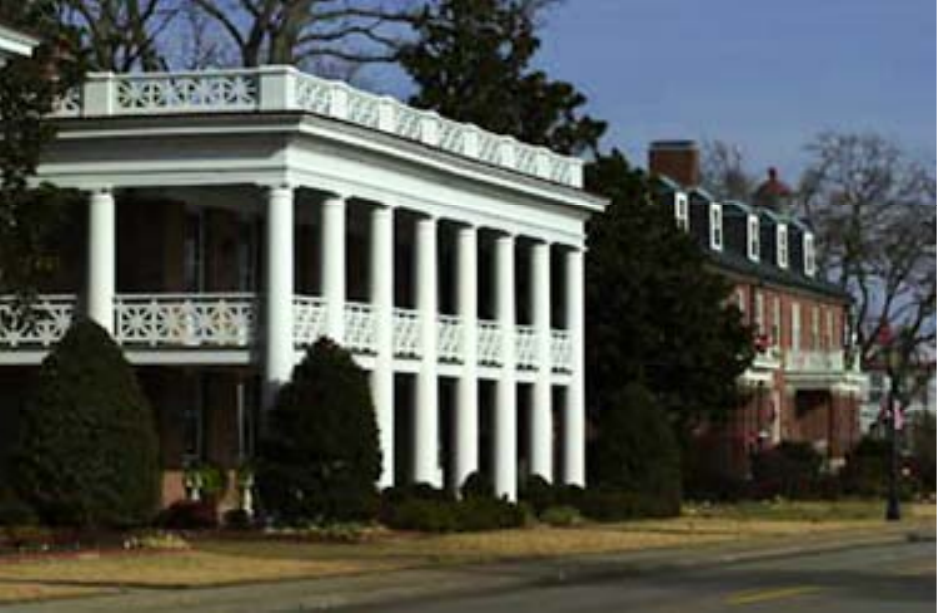}&
		\includegraphics[width=.15\linewidth]{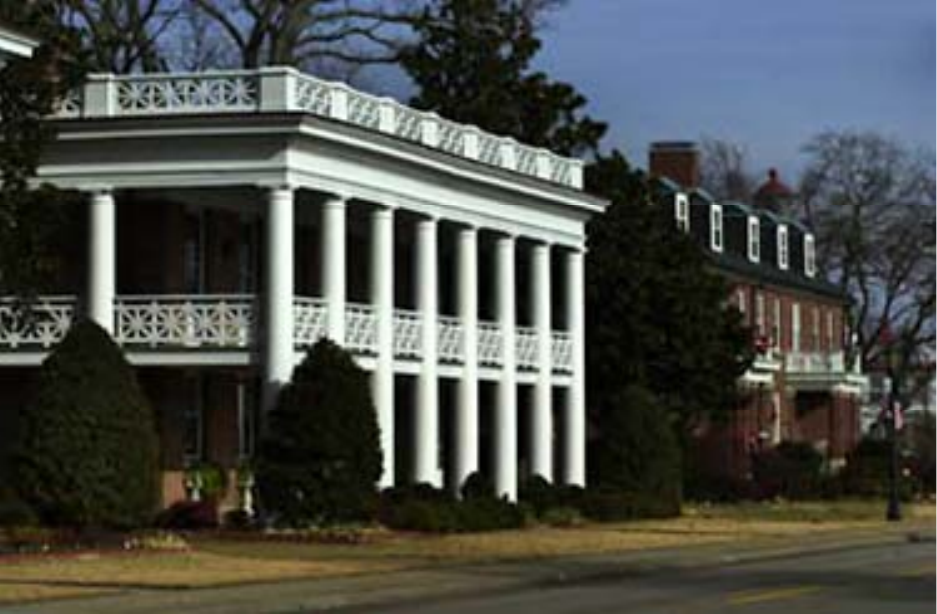}&
		\includegraphics[width=.15\linewidth]{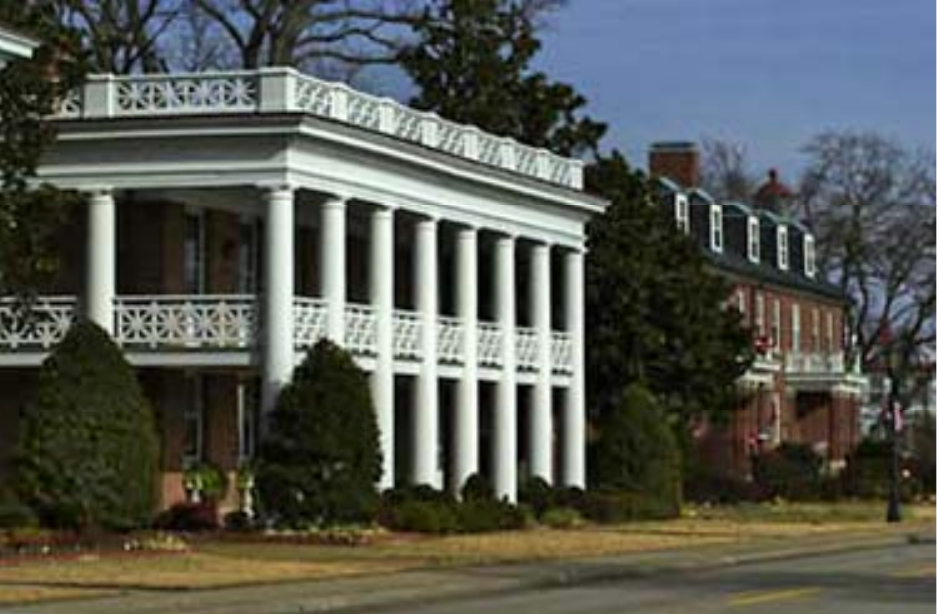}&
		\includegraphics[width=.15\linewidth]{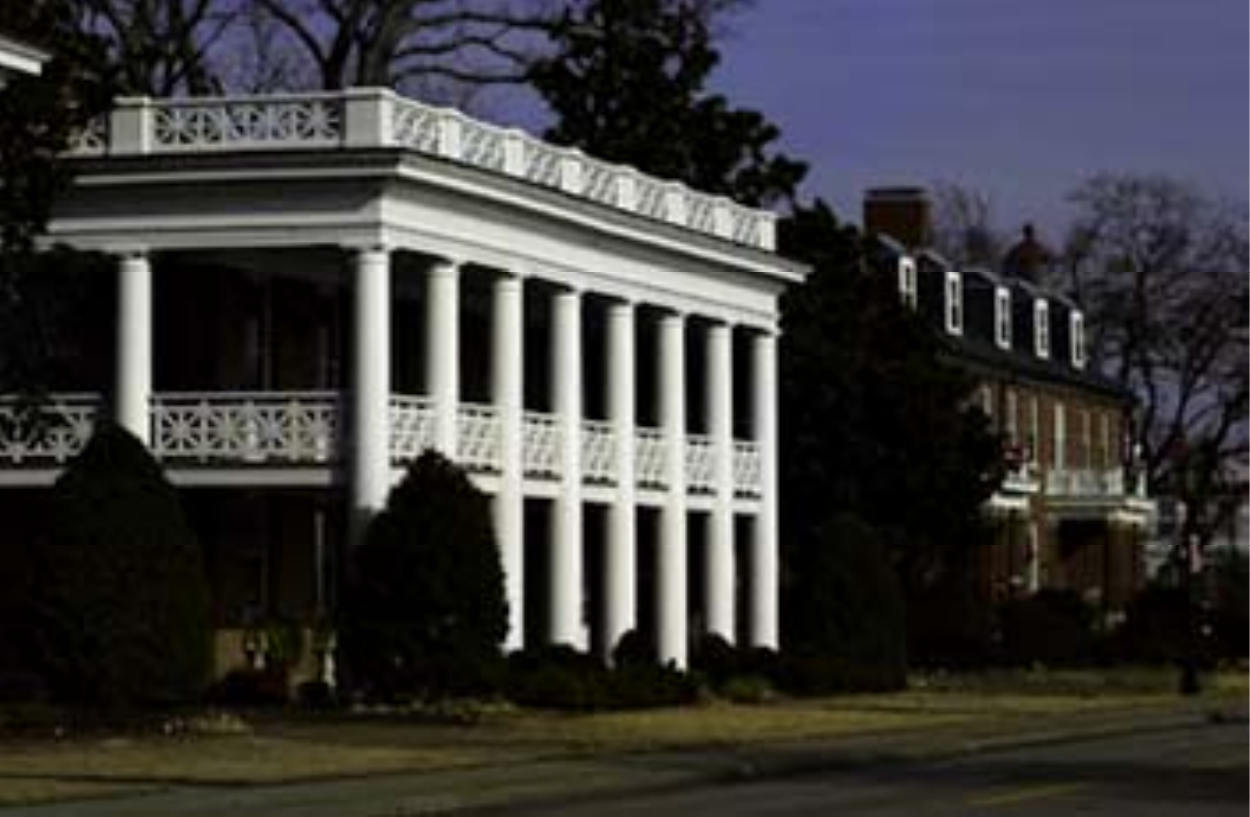}&
		\includegraphics[width=.15\linewidth]{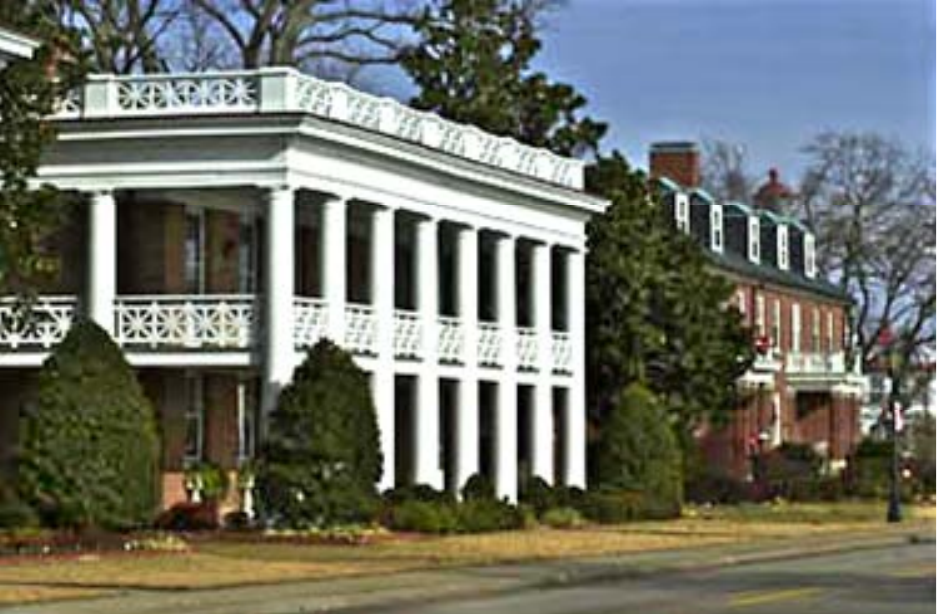}\\
		{4.95 / -}&{4.95 / 0.13}&{4.60 / 2.08}&{4.67 / 1.11}&{4.79 / 4.94}&\textbf{4.56 / 0.10}\\
		\includegraphics[width=.15\linewidth]{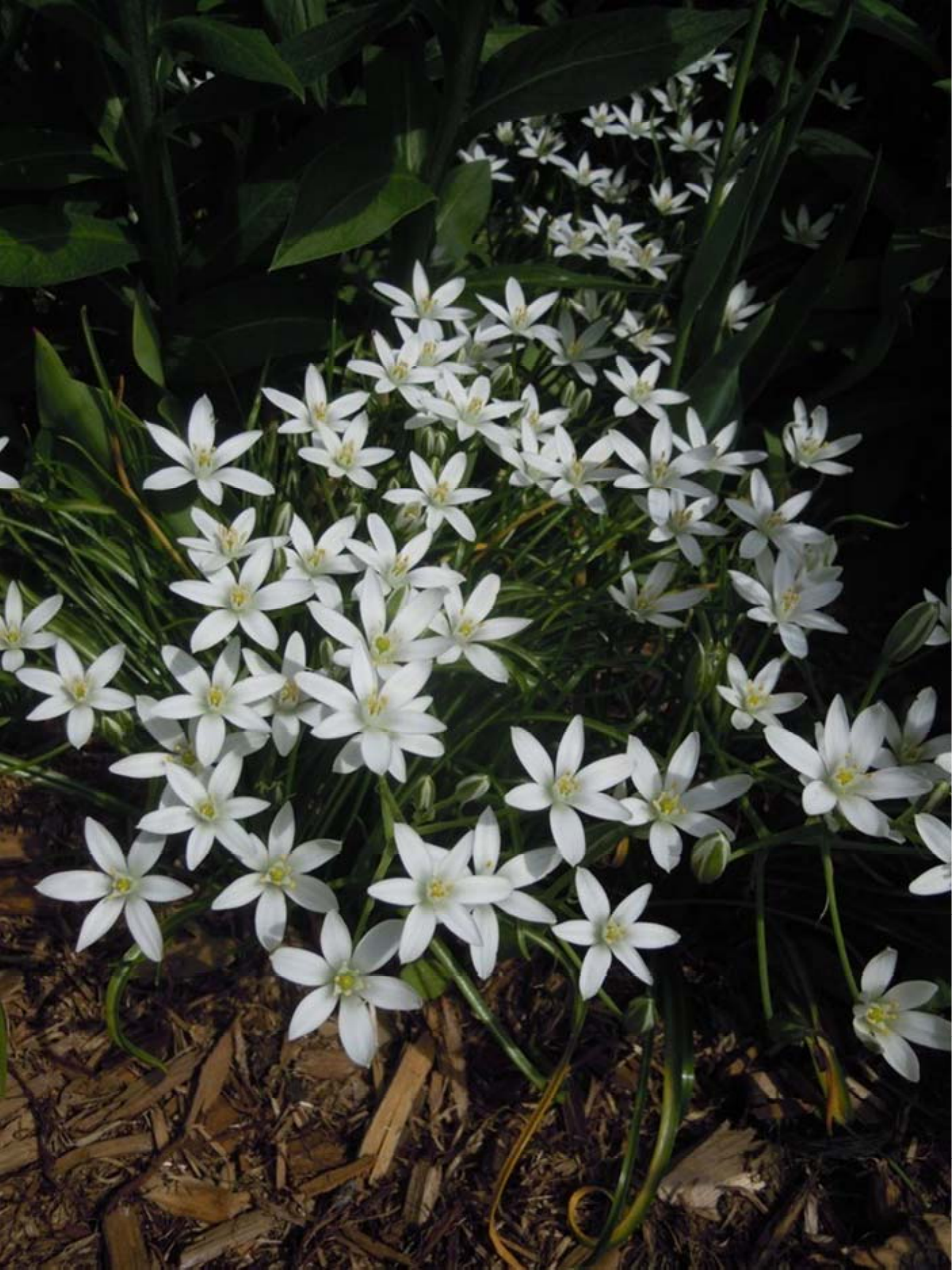}&
		\includegraphics[width=.15\linewidth]{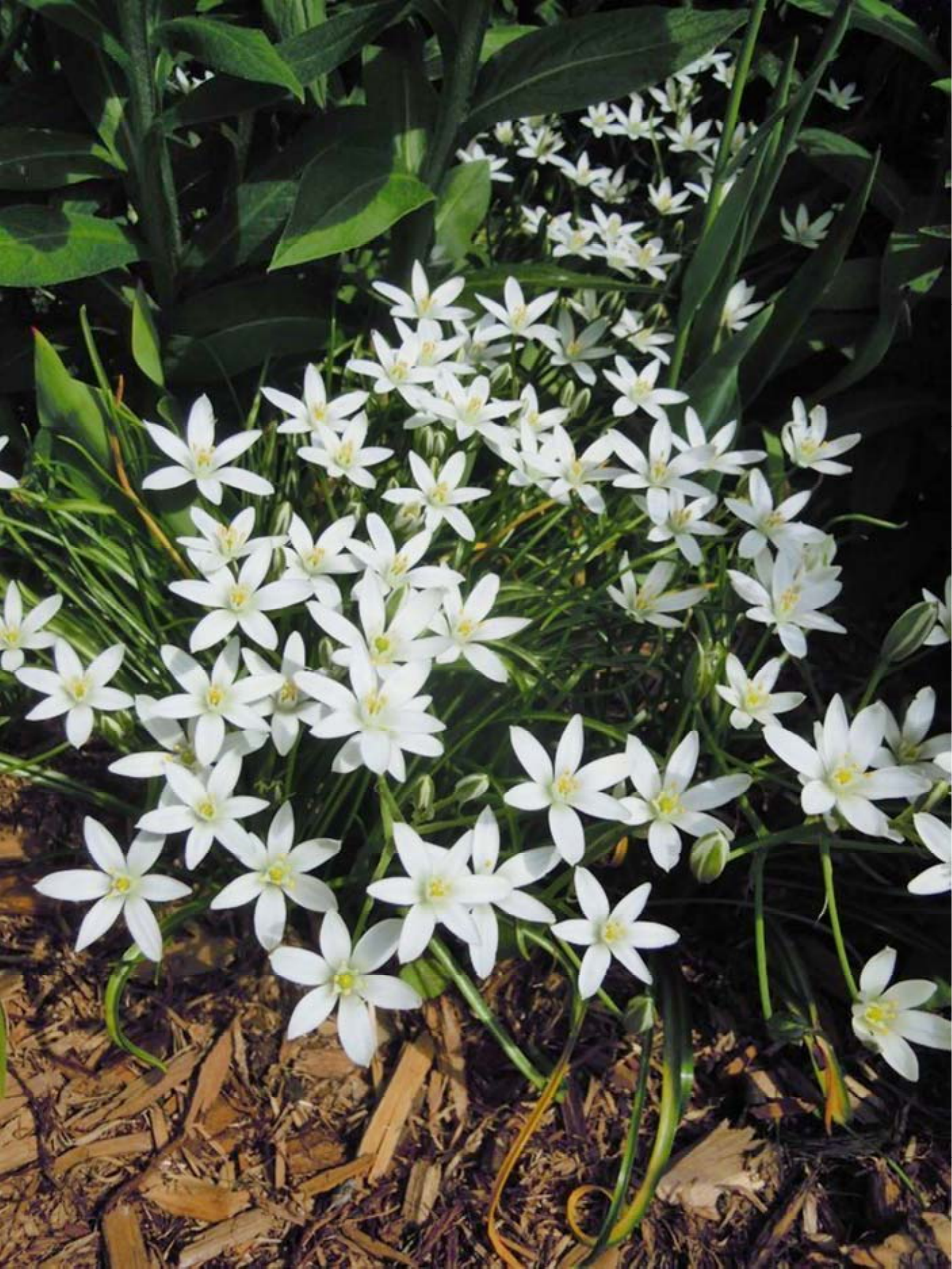}&
		\includegraphics[width=.15\linewidth]{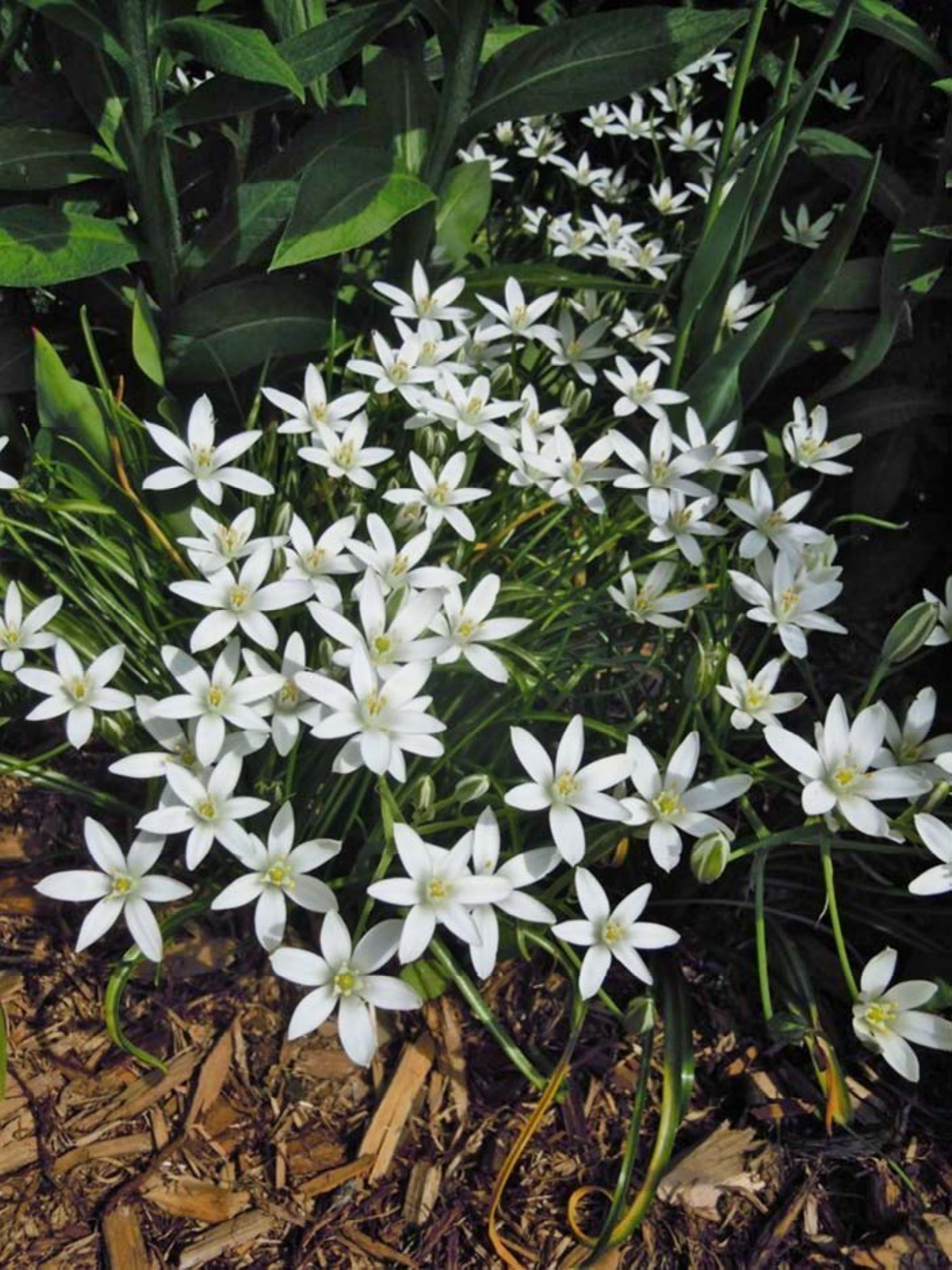}&
		\includegraphics[width=.15\linewidth]{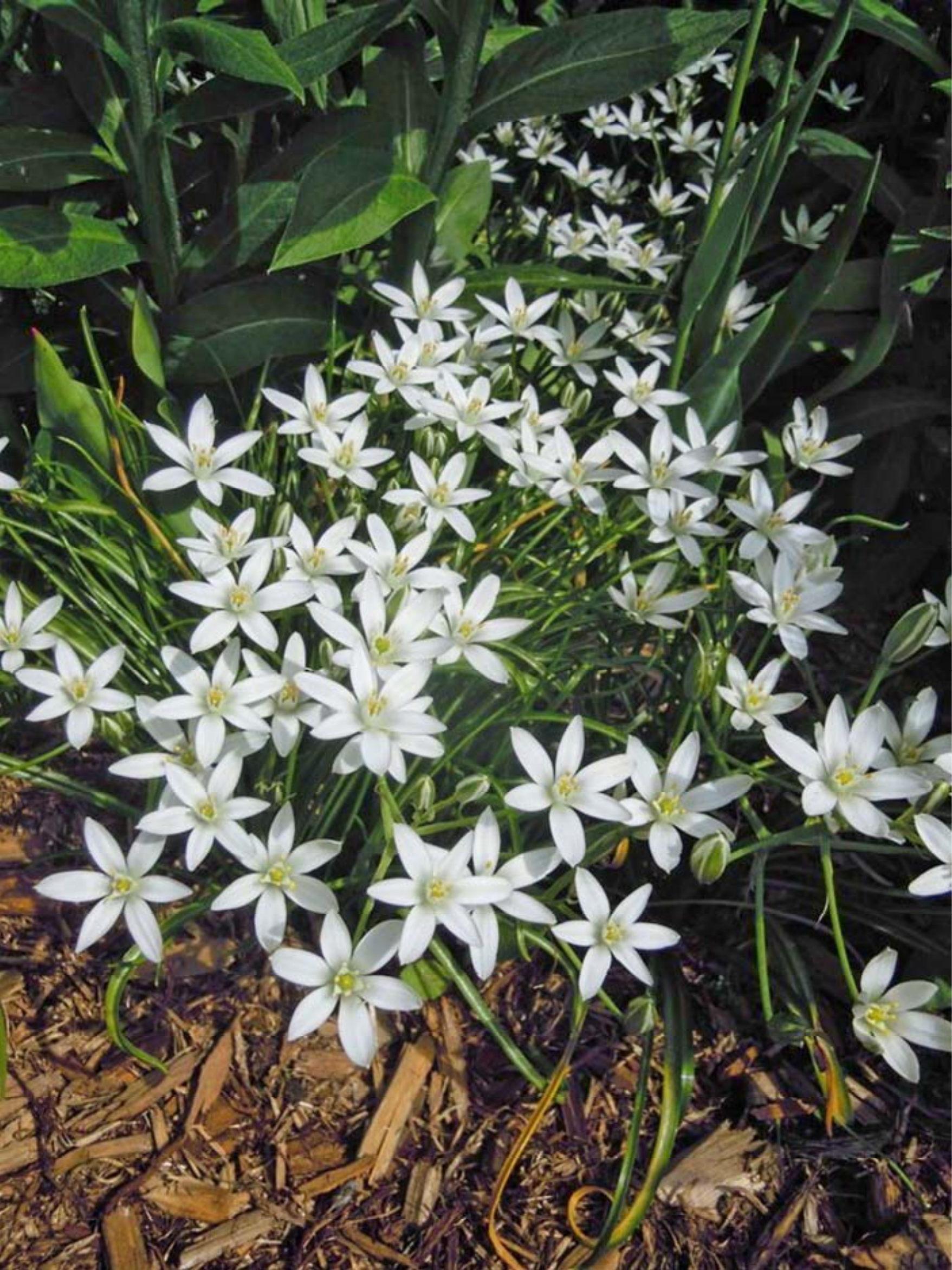}&
		\includegraphics[width=.15\linewidth]{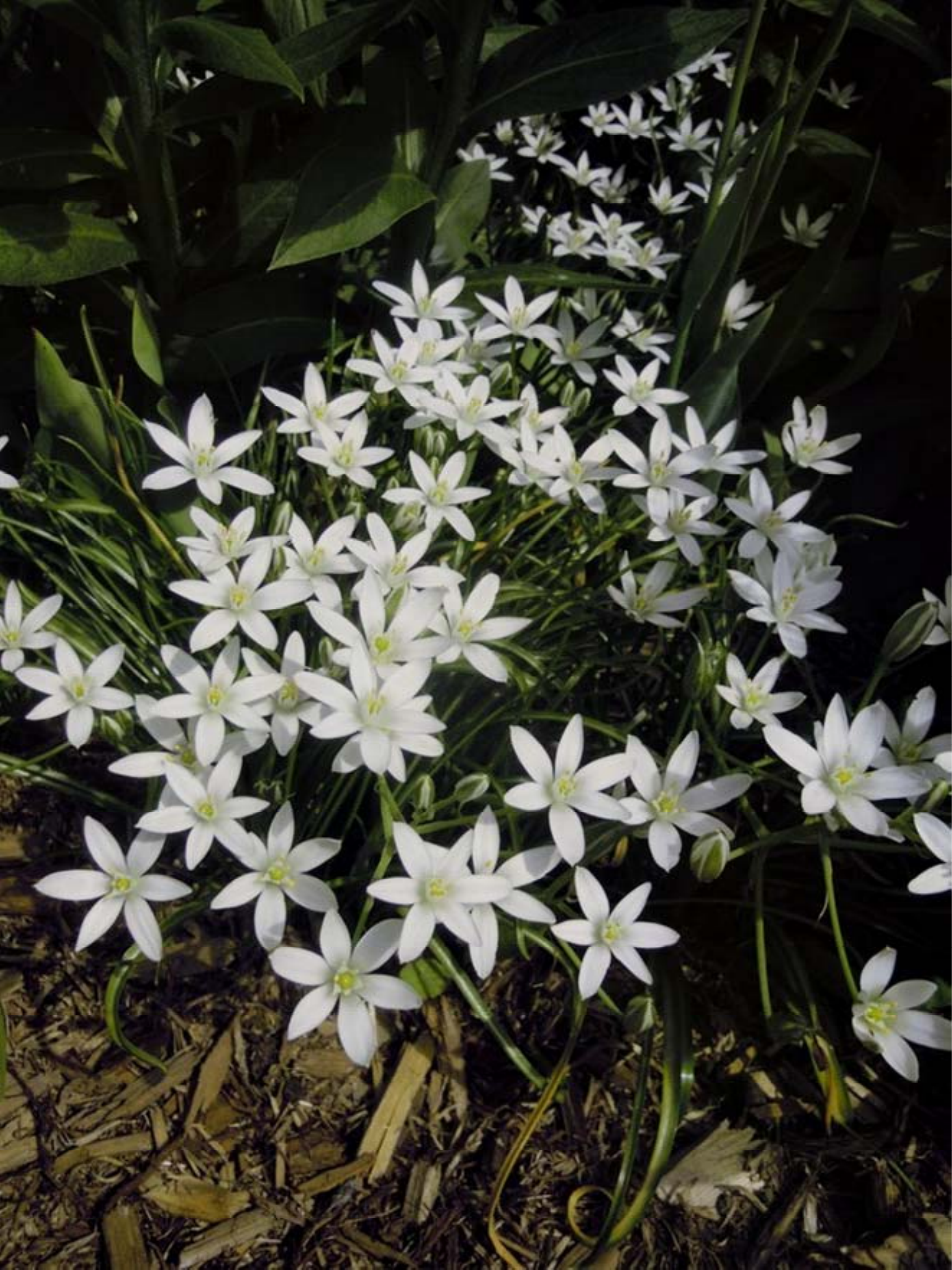}&
		\includegraphics[width=.15\linewidth]{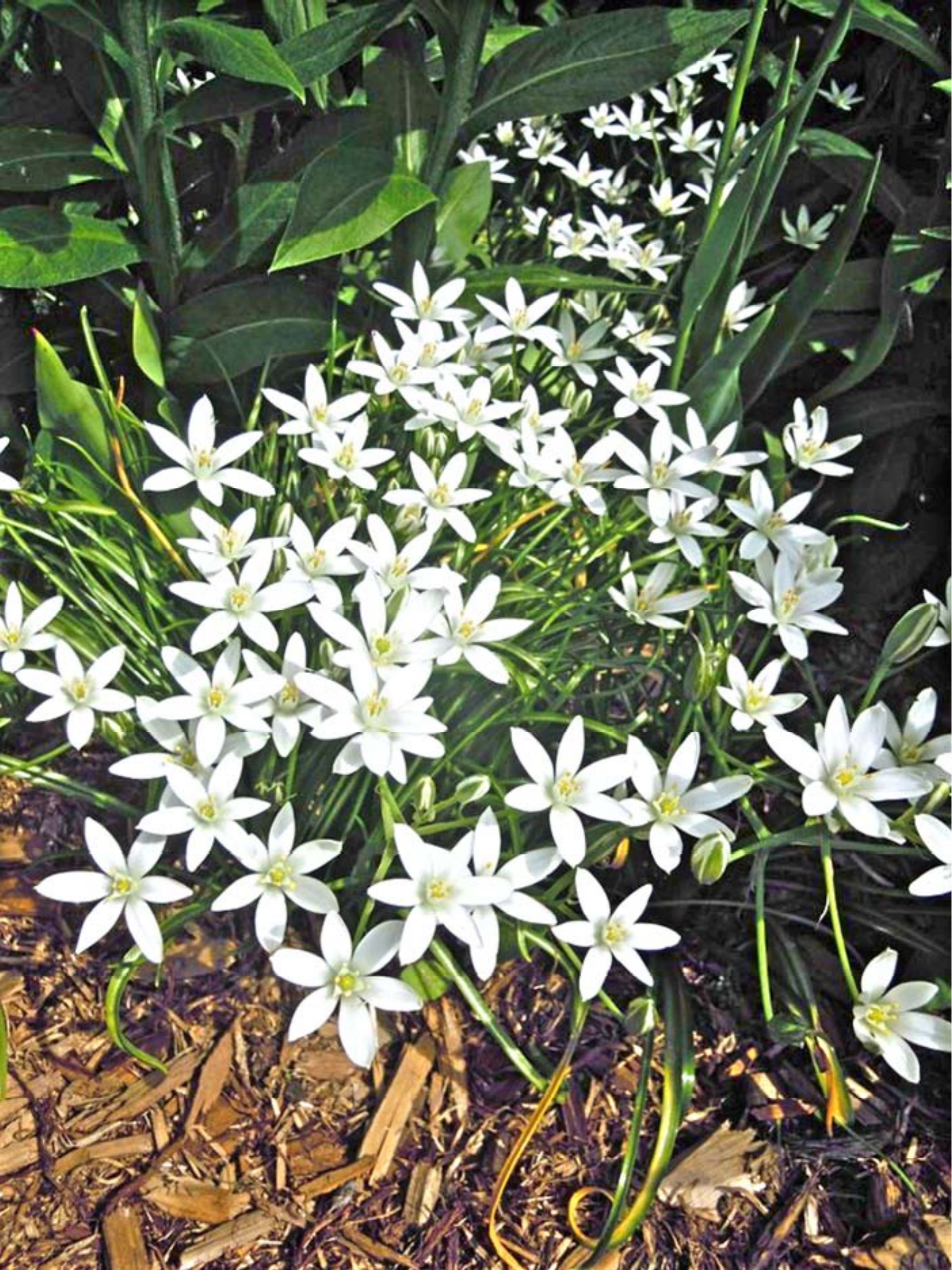}\\
		{2.67 / -}&{2.77 / 0.51}&{2.76 / 5.90}&{2.71 / 3.12}&{2.63 / 13.79}&\textbf{2.55 / 0.31}\\
		{Input}&{SRIE}&{WVM}&{JIEP}&{HDRNet}&{TECU}\\
	\end{tabular}
	\caption{Visual comparisons selected from the NASA dataset (top row) and the Non-uniform dataset (bottom row). Moreover, we post the NIQE  / computation time (s) below each image.}
	\label{fig:dataset}
\end{figure*}

\subsection{$\ell_0$-Regularized Dictionary Learning}

We generate synthetic data with different sizes (see Table \ref{Tab:Synthetic}) to help analyze the convergence properties of TECU.
Specifically, all the algorithms are terminated when satisfying the following condition:
\begin{equation}\small
\max\left\{\frac{\|\D^{t+1}-\D^{t}\|}{\|\D^t\|}, \ \frac{\|\W^{t+1}-\W^{t}\|}{\|\W^t\|}, \  \frac{\|\Psi^{t+1}-\Psi^{t}\|}{\|\Psi^t\|}\right\}<1e^{-4}.
\end{equation}

\subsubsection{Comparisons of Different Embedded Strategies}
Firstly, we conduct experiments to compare two particular cases of our proposed framework.
Precisely, the sign ``TECU'' in the Fig. \ref{fig:144}, Table \ref{Tab:Synthetic} and Table \ref{Tab:time} represents the case of embedding ADMM for updating $\D$, while using prox-linear update for $\W$ sub-problem;
the ``TECU-PITH'' in Fig. \ref{fig:144} refers to embedding PITH for updating $\W$, while applying prox-linear update for $\D$ sub-problem.

From the first row of Fig. \ref{fig:144}, it is distinct that these two cases of TECU have quite different convergence  performances.
From the top row of Fig. \ref{fig:144}(d), we can see that TECU requires a few propagation steps, however, TECU-PITH reaches the maximum inner steps (set as $20$) at almost every iteration.
The excessive inner propagations definitely decelerate the overall convergence speed, i.e., TECU uses 2.22s but TECU-PITH takes nearly 600s to converge.

This comparison result from one side shows that ADMM is more productive for optimizing $D$ sub-problem, but PITH is less effective for solving the sub-problem of $\W$.
On the other side, the different performances are also influenced by the characteristics of sub-problems.
The sub-problem of $\D$ minimizes strongly convex quadratic function with unit ball constraint, thus it can be efficiently solved.
However, due to the $\ell_0$ penalty, the sub-problem of $\W$ is NP hard, which is more difficult to optimize.
Therefore, this comparisons of different embedded strategies indicate the necessity of employing hybrid scheme in the framework of TECU, meanwhile, suggest embedding high-efficient numerical algorithms for sub-problem optimization.

\subsubsection{Comparisons with Other Algorithms}
Comparing with other existing algorithms, we can see from Table \ref{Tab:Synthetic} and Fig. \ref{fig:144} that TECU converges with less iteration steps, especially has better convergence performances on the optimization of $\D$ sub-problem.
Moreover, we give further comparisons in Table \ref{Tab:time} to show that the computation time of one propagation in TECU is much less than the ones in other algorithms.
Thus it is the reason why TECU totally adopts more propagations but has less computation time.
Moreover, since all the PALM, BCU and iPALM require estimating Lipschitz constants at every iteration, thus they have similar one-step computation time on different data scales.
It is obvious that estimating Lipschitz constants during iterations is extremely time-consuming especially for large scale data.
Thus though embedding numerical algorithms brings more propagations, it avoids estimating Lipschitz constants thus is far more efficient than existing algorithms.

\subsection{Low-light Image Enhancement}

Firstly, we conduct an experiment in Fig. \ref{fig:analysis} to compare TECU with the classical CD algorithm, i.e., PAM \cite{Attouch2010Proximal} on an example image from~\cite{Cai2018deep}.
Since \cite{Cai2018deep} provides image pairs of low-light images and the references obtained by other techniques, we provide PSNR values with respect to the given reference to give quantitative evaluations.
As shown in Fig. \ref{fig:analysis}, TECU achieves superior performances in terms of visual effect and PSNR score.
Moreover, we further plot the PSNR curves in Fig.~\ref{fig:analysis}(e).
From which we can tell that embedding network into the classical CD scheme certainly produces an excellent growth trend than employing itself.

We compare TECU with state-of-the-art approaches including: HE~\cite{cheng2004simple},  BPDHE~\cite{sheet2010brightness}, MSRCR~\cite{rahman2004retinex}, GOLW~\cite{shan2010globally}, NPEA~\cite{wang2013naturalness}, SRIE~\cite{fu2015probabilistic}, WVM~\cite{fu2016weighted}, JIEP~\cite{cai2017joint} and HDRNet~\cite{hasinoff2017Deep}, on the NASA dataset~\cite{nasa} and the Non-uniform dataset~\cite{wang2013naturalness}.
There are 23 images of different indoor and outdoor scenes in NASA dataset, while the Non-uniform dataset consists 130 low-quality images in different natural scenes including sunshine, overcast sky and nightfall scenarios.

For the lack of ground truth, it is impossible to give standard metrics (i.e., PSNR) to evaluate the quantitative performances for LIE task.
In previous literatures~\cite{wang2013naturalness,fu2016weighted,cai2017joint}, a blind image quality assessment called Natural Image Quality Evaluator (NIQE) is widely used to give quantitative evaluation for LIE.
Following this, we also present the NIQE scores in Table \ref{tab:rescomp}, comparing with all these state-of-the-art methods on the two different benchmarks.
The comparison results in Table \ref{tab:rescomp} indicate that TECU with embedded network has the lowest NIQE score and thus achieves the highest image quality.
We also provide a visual comparison on examples selected from both two datasets.
It is obvious that TECU is able to enhance image quality with high contrast, but other results are still contain details in dark, which are hard to recognize.
Thus from both quantitative and quality analyses, we can conclude that embedding networks in the framework of TECU is effective and competitive for the challenging LIE task.

\section{Conclusion}
We propose a realizable algorithm framework TECU, which embeds both numerical algorithms and advance techniques for optimizing a generic multivariate non-convex problem.
Through embedding task-oriented strategies, TECU is able to improve the convergence speed of the whole algorithm and obtain desired solutions with high probability.
Moreover, we further provide a realizable error control condition, to ensure robust performances with rigid theoretical supports.
The experimental results on two practical problems verify the superiorities of our proposed algorithm.

\section{Acknowledgments}
This work was supported by National Natural Science Foundation of China (Grant Nos. 61672125, 61733002, 61572096, 61632019 and 61806057), China Postdoctoral Science Foundation (Grant No. 2018M632018) and the Fundamental Research Funds for the Central Universities.


\begin{onecolumn}

\section{\LARGE Supplementary Material of\\ Task Embedded Coordinate Update: A Realizable Framework\\ for Multivariate Non-convex Optimization}
In this supplementary material, the contents are presented according to the following order:
\begin{enumerate}
	\item Revisit the definition of Kurdyka-{\L}ojasiewicz (K{\L}) property/function.
	\item Give detailed proofs of Proposition \ref{prop}.
	\item Provide detailed proofs of convergence: proofs of Proposition \ref{Converg_Ana:main_theorem} and Theorem \ref{Converg_Ana:key_theorem}.	
	\item Give more experimental results of low-light image enhancement task.
\end{enumerate}

\section{Kurdyka-{\L}ojasiewicz Property/Function}
\begin{defin}\label{Def:1}(Kurdyka-{\L}ojasiewicz function)
	Proper, lower semi-continuous function $\sigma: \mathbb{R}^d \to (-\infty, + \infty)$ is said to have the Kurdyka-{\L}ojasiewicz property at $\tilde{x}\in\mbox{dom}\partial \sigma:=\{x\in\mathbb{R}^d: \partial \sigma(x) \neq \emptyset\}$ if there exist $\mu\in(0,+\infty]$, a neighborhood $\mathcal{U}_{\tilde{x}}$ of $\tilde{x}$ and a desingularizing function $\phi:[0,\mu)\to \mathbb{R}_+$ which satisfies (1) $\phi(0)=0$; (2) $\phi$ is $C^1$ on $(0,\mu)$ and continuous at $0$; (3) for all $s\in(0,\mu): \phi'(s)>0$, such that for all
	\begin{equation}
	x\in\mathcal{U}_{\tilde{x}}\cap[\sigma(\tilde{x})<\sigma(x)<\sigma(\tilde{x})+\mu],
	\end{equation}
	the following inequality holds
	\begin{equation}
	\phi'(\sigma(x)-\sigma(\tilde{x}))\mbox{dist}(0,\partial \sigma(x)) \geq 1.
	\end{equation}
	Moreover, if $\sigma$ satisfies the K{\L} property at each point of $\mbox{dom}\partial \sigma$ then $\sigma$ is called a K{\L} function.
\end{defin}

\section{Detailed Proofs of Proposition \ref{prop}}
\begin{proof}
	From the calculations in Eq. \eqref{Alg_Imp:extra_variables}, we can deduce the following equalities.
	\begin{equation}\label{eq:xxx}
	\begin{aligned}
	\widetilde{\x}^{t,K_x^t}=&\mathrm{prox}^{1}_{f}(\x^{t,K_x^t} - \nabla_{\x}H(\x^{t,K_x^t}, \y^{t-1}) - \eta_1(\x^{t,K_x^t}-\x^{t-1}))\\
	=&\mathrm{prox}^{1}_{f}(\widetilde{\x}^{t,K_x^t} - \nabla_{\x}H(\widetilde{\x}^{t,K_x^t}, \y^{t-1}) - \eta_1(\widetilde{\x}^{t,K_x^t}-\x^{t-1})+\e_x^{t,K_x^t}).
	\end{aligned}
	\end{equation}
	Once the error $\e_x^{t,K_x^t}$ satisfies the Criterion \ref{ecrit}, $\widetilde{\x}^{t,K_x^t}$ will be assigned as $\x^t$ in the Eq. (\ref{eq:xxx}), thus we get
	\begin{equation}\label{eq:xo}
	\x^{t} =\mathrm{prox}^{1}_{f}(\x^{t} - \nabla_{\x}H(\x^{t}, \y^{t-1}) - \eta_1(\x^{t}-\x^{t-1})+\e_x^{t,K_x^t}).
\end{equation}
	The above deductions can be similarly extended to the case of $\y^t$:
	\begin{equation}\label{eq:xoo}
	\y^{t} =\mathrm{prox}^{1}_{g}(\y^{t} - \nabla_{\y}H(\x^{t}, \y^{t}) - \eta_2(\y^{t}-\y^{t-1})+\e_y^{t,K_y^t}).
	\end{equation}
	From the definition of proximal mapping operator, Eq. \eqref{eq:xo} and \eqref{eq:xoo} are equal to
	\begin{equation}
	\begin{aligned}
	\e_x^{t,K_x^t} &= \mathbf{g}_x^{t} + \nabla_{\x} H(\x^{t}, \y^{t-1}) + \eta_1(\x^{t}-\x^{t-1}),\\
	\e_y^{t,K_y^t} &= \mathbf{g}_y^{t} + \nabla_{\y} H(\x^{t}, \y^{t}) + \eta_2(\y^{t}-\y^{t-1}).
	\end{aligned}
	\end{equation}
	where $\mathbf{g}_x^{t+1} \in \partial f(\x^{t+1})$ and $\mathbf{g}_y^{t+1} \in \partial g(\y^{t+1})$.
	The above equalities show that $\e_x^{t,K_x^t}$ and $\e_y^{t,K_y^t}$ are implementations of $\e_x^t$ and $\e_y^t$ in Eq. \eqref{iPAM:iex_KKT}.
	$\blacksquare$
\end{proof}

\section{Detailed Proofs of Convergence}
Firstly, we can conclude from Eq. \eqref{iPAM:iex_KKT} that, the $\x^{t}$ and $\y^{t}$ updated by task embedding strategy can be regarded as solutions to the following subproblems:
\begin{equation}\label{Converg_Ana:inexact_exact}
\begin{aligned}
&\min\limits_{\x} f(\x) + H(\x,\y^{t-1}) + \frac{\eta_1}{2}\|\x-\x^{t-1}\|^2 - (\e_x^{t})^{\top}\x,\\
&\min\limits_{\y} g(\y) + H(\x^{t},\y) + \frac{\eta_2}{2}\|\y-\y^{t-1}\|^2 - (\e_y^{t})^{\top}\y.
\end{aligned}
\end{equation}
This equivalent conversion is strict since the first-order optimality conditions of Eq. (\ref{Converg_Ana:inexact_exact}) are exactly the same with Eq. \eqref{iPAM:iex_KKT}.
However, we have to emphasize that it is only used for theoretic analyses: we do not directly optimize Eq. \eqref{Converg_Ana:inexact_exact} in practice, instead, $\x^{t}$ and $\y^{t}$ are updated by task embedding strategy, as claimed in Alg. 1.

Since our proposed TECU is a hybrid framework which contains three different updates at each iteration, we would like to revisit the proximal update, prox-linear update and the theoretically-equivalent form of our novel task-embedding update:

For solving $\x$ sub-problem:
\begin{enumerate}	
	\item \emph{Proximal:} $\x^{t+1} \in \arg\min\limits_{\x} f(\x) + H(\x, \y^t) + \frac{\zeta_1^t}{2}\|\x-\x^t\|^2$,  $\zeta_1^t>0$.
	
	\item \emph{Prox-linear: } $\x^{t+1} \in \arg\min\limits_{\x} f(\x) +\frac{\gamma_1^t}{2}\|\x-(\x^t-\nabla_{\x}H(\x^t, \y^t)/\gamma_1^t)\|^2$,  $\gamma_1^t>L_1^t$.
	
	\item \emph{Task embedding:} $\x^{t+1} \in \arg\min\limits_{\x} f(\x) + H(\x, \y^t) + \frac{\eta_1}{2}\|\x-\x^t\|^2 - (\e_x^{t+1})^{\top}\x$,  $\eta_1>2C_x$.
\end{enumerate}

For solving $\y$ sub-problem:
\begin{enumerate}[resume]	
	\item \emph{Proximal:} $\y^{t+1} \in \arg\min\limits_{\y} g(\y) + H(\x^{t+1}, \y) + \frac{\zeta_2^t}{2}\|\y-\y^t\|^2$,  $\zeta_2^t>0$.
	
	\item \emph{Prox-linear:} $\y^{t+1} \in \arg\min\limits_{\y} g(\y) + \frac{\gamma_2^t}{2}\|\y-(\y^t - \nabla_{\y}H(\x^{t+1}, \y^t)/\gamma_2^t)\|^2$,  $\gamma_2^t>L_2^t$.
	
	\item \emph{Task embedding:} $\y^{t+1} \in \arg\min\limits_{\x} g(\y) + H(\x^{t+1}, \y) + \frac{\eta_2}{2}\|\y-\y^t\|^2 - (\e_y^{t+1})^{\top}\y$,  $\eta_2>2C_y$.
\end{enumerate}

There are totally 9 combinations under TECU framework, that is:
$
1-4, \ 1-5, \ 1-6, \ 2-4, \ 2-5, \ 2-6, \ 3-4, \ 3-5, \ 3-6.
$
However, we are only interested the ones that consist at least one task embedding update, that is, we consider the cases:
$$
1-6, \quad 2-6, \quad 3-4, \quad 3-5, \quad 3-6.
$$
In the subsequence, we will prove that this hybrid algorithm TECU has nice convergence property: it generates a Cauchy sequence that converges to a critical point of the original objective function.

\subsection{Proof for Proposition \ref{Converg_Ana:main_theorem}}
\begin{proof}
	Notice that ``$1-6$'', ``$2-6$''are the same with ``$3-4$'', ``$3-5$'' since $\x$ and $\y$ can be switched to each other.
	Thus we only give detailed proofs on cases of ``$1-6$'', ``$2-6$'' and ``$3-6$''.
	
	(\emph{Sufficient descent property: Eq. (\ref{suff_subound}a)})
	
	For ``$3-6$'' with task embedding updates on both subproblems, we have the following inequalities:
	\begin{equation}\label{eq_1}
	\begin{aligned}
	&f(\x^{t+1}) + H(\x^{t+1}, \y^t) + \frac{\eta_1}{2}\|\x^{t+1}-\x^t\|^2 - (\e_x^{t+1})^{\top}\x^{t+1} \leq  f(\x^{t}) + H(\x^{t}, \y^t) - (\e_x^{t+1})^{\top}\x^{t},\\
	&g(\y^{t+1}) + H(\x^{t+1}, \y^{t+1}) + \frac{\eta_2}{2}\|\y^{t+1}-\y^t\|^2 - (\e_y^{t+1})^{\top}\y^{t+1} \leq g(\y^{t}) + H(\x^{t+1}, \y^{t}) - (\e_y^{t+1})^{\top}\y^{t}.
	\end{aligned}
	\end{equation}
	Adding the above two inequalities, then we can get the following inequalities with positive real numbers $\rho_1$ and $\rho_2$:
	\begin{equation}
	\begin{aligned}
	\Psi(\z^t) - \Psi(\z^{t+1}) \geq& \frac{\eta_1}{2}\|\x^{t+1}-\x^t\|^2  + \frac{\eta_2}{2}\|\y^{t+1}-\y^t\|^2 + (\e_x^{t+1})^{\top}(\x^{t}-\x^{t+1}) + (\e_y^{t+1})^{\top}(\y^{t}-\y^{t+1})\\
	\geq& \frac{\eta_1}{2}\|\x^{t+1}-\x^t\|^2  + \frac{\eta_2}{2}\|\y^{t+1}-\y^t\|^2 \\
	&- (\frac{\rho_1}{2}\|\e_x^{t+1}\|^2 + \frac{1}{2\rho_1}\|\x^{t+1}-\x^t\|^2)- (\frac{\rho_2}{2}\|\e_y^{t+1}\|^2 + \frac{1}{2\rho_2}\|\y^{t+1}-\y^t\|^2).
	\end{aligned}
	\end{equation}
	The last inequality comes from applying Young's inequality.
	Then by combining the Criterion \ref{ecrit} for TECU, we have:
	\begin{equation}
	\begin{aligned}
	&\Psi(\z^t) - \Psi(\z^{t+1}) \\
	\geq& (\frac{\eta_1}{2}-\frac{1}{2\rho_1})\|\x^{t+1}-\x^t\|^2  + (\frac{\eta_2}{2}-\frac{1}{2\rho_2})\|\y^{t+1}-\y^t\|^2 - \frac{\rho_1(C_x)^2}{2}\|\x^{t}-\x^{t-1}\|^2 - \frac{\rho_2(C_y)^2}{2}\|\y^{t}-\y^{t-1}\|^2\\
	\geq& \frac{\eta_1}{4}\|\x^{t+1}-\x^t\|^2 + \frac{\eta_2}{4}\|\y^{t+1}-\y^t\|^2- \frac{(C_x)^2}{\eta_1}\|\x^{t}-\x^{t-1}\|^2 - \frac{(C_y)^2}{\eta_2}\|\y^{t}-\y^{t-1}\|^2,
	\end{aligned}
	\end{equation}
	where the last equality holds by setting $\rho_1=\frac{2}{\eta_1}$ and $\rho_2 = \frac{2}{\eta_2}$.
	
	Denoting $\Phi^1(\z, \w) := \Psi(\z) + \frac{(C_x)^2}{\eta_1}\|\x-\ssp\|^2 + \frac{(C_y)^2}{\eta_2}\|\y-\ssq\|^2$ with $\w:=(\ssp, \ssq)$. Then by denoting $\z=\z^t$, $\ssp=\x^{t-1}$ and $\ssq=\y^{t-1}$, we denote $\Phi^1(\z^t, \z^{t-1}) := \Psi(\z^t) + \frac{(C_x)^2}{\eta_1}\|\x^{t}-\x^{t-1}\|^2 + \frac{(C_y)^2}{\eta_2}\|\y^{t}-\y^{t-1}\|^2$, then the above inequality is equal to:
	\begin{equation}
	\Phi^1(\z^t, \z^{t-1}) \geq \Phi^1(\z^{t+1}, \z^{t}) + (\frac{\eta_1}{4} - \frac{(C_x)^2}{\eta_1}) \|\x^{t+1}-\x^t\|^2 + (\frac{\eta_2}{4} - \frac{(C_y)^2}{\eta_2})\|\y^{t+1}-\y^{t}\|^2.
	\end{equation}
	
	For ``$1-6$'', we have the following inequality from the iterative scheme of proximal update:
	\begin{equation}
	f(\x^{t+1}) + H(\x^{t+1}, \y^t)+\frac{\zeta_1^t}{2}\|\x^{t+1}-\x^t\|^2 \leq f(\x^{t}) + H(\x^{t}, \y^t).
	\end{equation}
	Then, together with the second inequality of Eq. \eqref{eq_1}, we have that
	\begin{equation}
	\begin{aligned}
	\Psi(\z^t) \geq& \Psi(\z^{t+1}) + \frac{\zeta_1^t}{2}\|\x^{t+1}-\x^t\|^2 + \frac{\eta_2}{2}\|\y^{t+1}-\y^t\|^2 - (\frac{\rho}{2}\|\e_y^{t+1}\|^2 + \frac{1}{2\rho}\|\y^{t+1}-\y^t\|^2)\\
	\geq& \Psi(\z^{t+1}) + \frac{\zeta_1^t}{2}\|\x^{t+1}-\x^t\|^2  + \frac{\eta_2}{4}\|\y^{t+1}-\y^t\|^2 - \frac{(C_y)^2}{\eta_2}\|\y^t-\y^{t-1}\|^2,\\
	\end{aligned}
	\end{equation}
	where the last inequality holds by setting $\rho = \frac{2}{\eta_2}$.
	
	Then by denoting  $\Phi^2(\z, \w) = \Psi(\z) + \frac{(C_y)^2}{\eta_2}\|\y-\ssq\|^2$ and assign $\z = \z^t$, $\ssp = \x^{t-1}$ and $\ssq=\y^{t-1}$, we have $\Phi^2(\z^{t}, \z^{t-1}) = \Psi(\z^t) + \frac{(C_y)^2}{\eta_2}\|\y^t-\y^{t-1}\|^2$.
	Then, the following inequality holds:
	\begin{equation}
	\Phi^2(\z^{t+1}, \z^t) \geq \Phi^2(\z^t, \z^{t-1}) + \frac{\zeta_1^t}{2}\|\x^{t+1}-\x^t\|^2 + (\frac{\eta_2}{4} - \frac{(C_y)^2}{\eta_2})\|\y^{t+1}-\y^{t}\|^2.
	\end{equation}
	
	While, for the case ``$2-6$'', its prox-linear update indicates that
	\begin{equation}
	f(\x^{t+1}) + (\x^{t+1}-\x^t)^{\top}\nabla_{\x}H(\x^t, \y^t)+\frac{\gamma_1^t}{2}\|\x^{t+1}-\x^t\|^2 \leq f(\x^{t}).
	\end{equation}
	Together with the descent lemma for gradient Lipschitz functions described in \cite{Ortega1970Iterative}:
	\begin{equation}
	H(\x^{t+1}, \y^t) \leq H(\x^t, \y^t) + (\x^{t+1}-\x^t)^{\top}\nabla_{\x}H(\x^t, \y^t) + \frac{L_1^t}{2}\|\x^{t+1}-\x^t\|^2,
	\end{equation}
	we have
	\begin{equation}
	f(\x^{t+1}) + H(\x^{t+1}, \y^t) +\frac{\gamma_1^t-L_1^t}{2}\|\x^{t+1}-\x^t\|^2 \leq  f(\x^t) + H(\x^t, \y^t).
	\end{equation}
	Then in a similar way as ``$1-6$'', we have that
	\begin{equation}
	\Phi^2(\z^{t+1}, \z^t) \geq \Phi^2(\z^t, \z^{t-1}) + \frac{\gamma_1^t-L_1^t}{2}\|\x^{t+1}-\x^t\|^2 + (\frac{\eta_2}{4} - \frac{(C_y)^2}{\eta_2})\|\y^{t+1}-\y^{t}\|^2
	\end{equation}
	
	Thus we conclude that there exists a function $\Phi$ that for all the $5$ inexact cases, there holds:
	\begin{equation}\label{eq_2}
	\Phi(\z^{t+1}, \z^{t}) -\Phi(\z^{t}, \z^{t-1}) \geq a\|\z^{t+1}-\z^t\|^2,
	\end{equation}
	where $\Phi(\z, \w)$ varies from different combination forms:
	\begin{itemize}
		\item For the cases ``1-6'' and ``2-6'': $\Phi(\z^{t}, \z^{t-1}) := \Psi(\z^t) + \frac{(C_y)^2}{\eta_2}\|\y^{t}-\y^{t-1}\|^2$,
		\item For the cases ``3-4'' and ``3-5'': $\Phi(\z^{t}, \z^{t-1}) := \Psi(\z^t) + \frac{(C_x)^2}{\eta_1}\|\x^{t}-\x^{t-1}\|^2$,
		\item For the case ``3-6'': $\Phi(\z^{t}, \z^{t-1}) := \Psi(\z^t) + \frac{(C_x)^2}{\eta_1}\|\x^t-\x^{t-1}\|^2 + \frac{(C_y)^2}{\eta_2}\|\y^t-\y^{t-1}\|^2$.
	\end{itemize}
	Under the parameter conditions related to each update, it can be concluded that there exists $a>0$, which is adhering to specific combination forms:
	
	For ``$1-6$'': $a=\min_{t\in \mathbb{N}}\{\frac{\zeta_1^t}{2}, \frac{\eta_2}{4}-\frac{(C_y)^2}{\eta_2}\}$; \quad \quad \quad \
	For ``$3-4$'': $a=\min_{t\in \mathbb{N}}\{\frac{\eta_1}{4}-\frac{(C_x)^2}{\eta_1}, \frac{\zeta_2^t}{2}\}$;
	
	For ``$2-6$'': $a=\min_{t\in \mathbb{N}}\{\frac{\gamma_1^t-L_1^t}{2}, \frac{\eta_2}{4}-\frac{(C_y)^2}{\eta_2}\}$;
	\quad \quad
	For ``$3-5$'': $a=\min_{t\in \mathbb{N}}\{\frac{\eta_1}{4}-\frac{(C_x)^2}{\eta_1}, \frac{\gamma_2^t-L_2^t}{2}\}$;
	
	For ``$3-6$'': $a=\min\{\frac{\eta_1}{4}-\frac{(C_x)^2}{\eta_1}, \frac{\eta_2}{4}-\frac{(C_y)^2}{\eta_2}\}$.
	
	Since $\Psi(\z)$ is a bounded function, thus $\Phi(\z, \w)$ is bounded from the definition of $\Phi(\z, \w)$.
	Thus we have proved the first assertion in Proposition \ref{Converg_Ana:main_theorem}.
	
	(\emph{Bounded subgradient property: Eq. (\ref{suff_subound}b)})
	
	From the definition of $\Psi(\x, \y)$ and $\Phi(\z, \w)=\Phi(\x, \y, \ssp, \ssq)$, we have \begin{equation}
	\partial \Psi(\z^t) = (\partial_{\x}\Psi(\z^t), \partial_{\y}\Psi(\z^t)) = (\mathbf{g}_x^{t} + \nabla_{\x} H(\x^t, \y^t), \mathbf{g}_y^{t} + \nabla_{\y} H(\x^t, \y^t)),
	\end{equation}
	and $\partial\Phi(\z^t, \z^{t-1}) = (\partial_{\x}, \partial_{\y}, \partial_{\ssp}, \partial_{\ssq})\Phi(\z^t, \z^{t-1})$.
	
	For the cases ``1-6'' and ``2-6'', we have the following formula with the formation of $\Phi(\z^t, \z^{t-1})$.
	\begin{equation}\label{eq_3}
	(\mathbf{g}_x^{t} + \nabla_{\x} H(\x^t, \y^t), \mathbf{g}_y^{t} + \nabla_{\y} H(\x^t, \y^t) + \frac{2(C_y)^2}{\eta_2}(\y^t-\y^{t-1}), \mathbf{0}, \frac{2(C_y)^2}{\eta_2}(\y^{t-1}-\y^{t})) \in \partial\Phi(\z^t, \z^{t-1}).
	\end{equation}
	From the update of ``1-6'', we have
	\begin{equation}
	\begin{aligned}
	\mathbf{g}_x^{t} + \nabla_{\x} H(\x^t, \y^{t-1}) + \zeta_1^{t-1}(\x^t-\x^{t-1}) &= 0,\\
	\mathbf{g}_y^{t} + \nabla_{\y} H(\x^t, \y^{t})  + \eta_2(\y^t-\y^{t-1}) &= \e_y^{t}.
	\end{aligned}
	\end{equation}
	Together with Eq. \eqref{eq_3}, we have
	$(\nabla_{\x} H(\x^t, \y^t)- \nabla_{\x} H(\x^t, \y^{t-1}) + \zeta_1^{t-1}(\x^{t-1}-\x^{t}),  \e_y^{t} + \eta_2(\y^{t-1}-\y^{t})+ \frac{2(C_y)^2}{\eta_2}(\y^t-\y^{t-1}),\mathbf{0}, \frac{2(C_y)^2}{\eta_2}(\y^{t-1}-\y^{t})) \in \partial\Phi(\z^t, \z^{t-1})$.
	Thus we have
	\begin{equation}
	\begin{aligned}
	\|\partial\Phi(\z^t, \z^{t-1})\| &\leq \zeta_1^{t-1} \|\x^t-\x^{t-1}\| + (M + \eta_2 + \frac{4(C_y)^2}{\eta_2})\|\y^t-\y^{t-1}\| + C_y\|\y^{t-1}-\y^{t-2}\|\\
	&\leq (\zeta_1^{t-1} + M + \eta_2 + \frac{4(C_y)^2}{\eta_2}) \|\z^t-\z^{t-1}\| +  C_y\|\z^{t-1}-\z^{t-2}\|,
	\end{aligned}
	\end{equation}
	where $M$ is the Lipschitz moduli of $\nabla H$.
	Similarly, we have $\|\partial\Phi(\z^t, \z^{t-1})\| \leq (\gamma_1^{t-1} + M + \eta_2 + \frac{4(C_y)^2}{\eta_2}) \|\z^t-\z^{t-1}\| +  C_y\|\z^{t-1}-\z^{t-2}\|$ for the case ``2-6''.
	
	While for the case ``3-4'' and ``3-5'', we have
	\begin{equation}\label{eq_4}
	(\mathbf{g}_x^{t} + \nabla_{\x} H(\x^t, \y^t) + \frac{2(C_x)^2}{\eta_1}(\x^t-\x^{t-1}), \mathbf{g}_y^{t} + \nabla_{\y} H(\x^t, \y^t), \frac{2(C_x)^2}{\eta_1}(\x^{t-1}-\x^{t}), \mathbf{0}) \in \partial\Phi(\z^t, \z^{t-1}).
	\end{equation}
	From the update of ``3-4'', we have
	\begin{equation}
	\begin{aligned}
	\mathbf{g}_x^{t} + \nabla_{\x} H(\x^t, \y^{t-1}) + \eta_1(\x^t-\x^{t-1}) &= \e_x^{t},\\
	\mathbf{g}_y^{t} + \nabla_{\y} H(\x^t, \y^{t})  + \zeta_2^{t-1}(\y^t-\y^{t-1}) &= 0.
	\end{aligned}
	\end{equation}
	Together with Eq. \eqref{eq_4}, we have
	$
	(\nabla_{\x} H(\x^t, \y^t)- \nabla_{\x} H(\x^t, \y^{t-1}) + \eta_1(\x^{t-1}-\x^{t}) + \e_x^{t} + \frac{2(C_x)^2}{\eta_1}(\x^t-\x^{t-1}),  \zeta_2^{t-1}(\y^{t-1}-\y^{t}), \frac{2(C_x)^2}{\eta_1}(\x^{t-1}-\x^{t}),\mathbf{0}) \in \partial\Phi(\z^t, \z^{t-1})$.
	Thus we have
	\begin{equation}
	\begin{aligned}
	\|\partial\Phi(\z^t, \z^{t-1})\| &\leq (\eta_1+\frac{4(C_x)^2}{\eta_1}) \|\x^t-\x^{t-1}\| + (M + \zeta_2^{t-1})\|\y^t-\y^{t-1}\| + C_x\|\x^{t-1}-\x^{t-2}\|\\
	&\leq (\eta_1+\frac{4(C_x)^2}{\eta_1} + M + \zeta_2^{t-1}) \|\z^t-\z^{t-1}\| +  C_x\|\z^{t-1}-\z^{t-2}\|.
	\end{aligned}
	\end{equation}
	Similarly, we have $\|\partial\Phi(\z^t, \z^{t-1})\| \leq (\eta_1+\frac{4(C_x)^2}{\eta_1} + M + \gamma_2^{t-1}) \|\z^t-\z^{t-1}\| +  C_x\|\z^{t-1}-\z^{t-2}\|$ for the case ``3-5''.
	
	Lastly, for the case ``3-6'', we have
	\begin{equation}\label{eq_5}
	\begin{aligned}
	(\mathbf{g}_x^{t} + \nabla_{\x} H(\x^t, \y^t) + \frac{2(C_x)^2}{\eta_1}(\x^t-\x^{t-1}), \mathbf{g}_y^{t} + \nabla_{\y} H(\x^t, \y^t) + \frac{2(C_y)^2}{\eta_2}(\y^t-\y^{t-1}),&\\ \frac{2(C_x)^2}{\eta_1}(\x^{t-1}-\x^{t}), \frac{2(C_y)^2}{\eta_2}(\y^{t-1}-\y^{t})) \in \partial\Phi(\z^t, \z^{t-1})&.
	\end{aligned}
	\end{equation}
	On the other hand, from the update of ``3-6'', we have
	\begin{equation}
	\begin{aligned}
	\mathbf{g}_x^{t} + \nabla_{\x} H(\x^t, \y^{t-1}) + \eta_1(\x^t-\x^{t-1}) &= \e_x^{t},\\
	\mathbf{g}_y^{t} + \nabla_{\y} H(\x^t, \y^{t})  + \eta_2(\y^t-\y^{t-1}) &= \e_y^{t}.
	\end{aligned}
	\end{equation}
	From Eq. \eqref{eq_5}:
	$
	(\nabla_{\x} H(\x^t, \y^t)- \nabla_{\x} H(\x^t, \y^{t-1}) + \eta_1(\x^{t-1}-\x^{t}) + \e_x^{t} + \frac{2(C_x)^2}{\eta_1}(\x^t-\x^{t-1}),  \e_y^{t} + \eta_2(\y^{t-1}-\y^{t})+ \frac{2(C_y)^2}{\eta_2}(\y^t-\y^{t-1}), \frac{2(C_x)^2}{\eta_1}(\x^{t-1}-\x^{t}),\frac{2(C_y)^2}{\eta_2}(\y^{t-1}-\y^{t})) \in \partial\Phi(\z^t, \z^{t-1})$.
	Thus we have
	\begin{equation}
	\begin{aligned}
	\|\partial\Phi(\z^t, \z^{t-1})\| \leq& (\eta_1+\frac{4(C_x)^2}{\eta_1}) \|\x^t-\x^{t-1}\| + (M+\eta_2+\frac{4(C_y)^2}{\eta_2})\|\y^t-\y^{t-1}\|\\
	&+ C_x\|\x^{t-1}-\x^{t-2}\| + C_y\|\y^{t-1}-\y^{t-2}\|\\
	\leq& (\eta_1+\frac{4(C_x)^2}{\eta_1} + \eta_2+\frac{4(C_y)^2}{\eta_2}) \|\z^t-\z^{t-1}\| +  (C_x + C_y)\|\z^{t-1}-\z^{t-2}\|.
	\end{aligned}
	\end{equation}
	In conclusion, there exists $b > 0$, which is adhering to specific combination forms:
	
	For ``$1-6$'': $b=\max_{t\in \mathbb{N}}\{\zeta_1^{t-1} + M + \eta_2 + \frac{4(C_y)^2}{\eta_2}, C_y\}$;
	
	For ``$2-6$'': $b=\max_{t\in \mathbb{N}}\{\gamma_1^{t-1} + M + \eta_2 + \frac{4(C_y)^2}{\eta_2}, C_y\}$;
	
	For ``$3-4$'': $b=\max_{t\in \mathbb{N}}\{\zeta_2^{t-1} + M + \eta_1+\frac{4(C_x)^2}{\eta_1}, C_x\}$;
	
	For ``$3-5$'': $b=\max_{t\in \mathbb{N}}\{\gamma_2^{t-1} + M + \eta_1+\frac{4(C_x)^2}{\eta_1}, C_x\}$;
	
	For ``$3-6$'': $b=\max\{M+\eta_1+ \eta_2+\frac{4(C_x)^2}{\eta_1} +\frac{4(C_y)^2}{\eta_2}, C_x + C_y\}$,
	
	such that
	\begin{equation}
	\|\partial\Phi(\z^t, \z^{t-1})\| \leq b (\|\z^t-\z^{t-1}\| + \|\z^{t-1}-\z^{t-2}\|).
	\end{equation}
	Thus, we have proved the second assertion in the Proposition \ref{Converg_Ana:main_theorem}.
	
	(\emph{Bondedness of sequence})
	
	Lastly, $\Psi(\mathbf{z})$ is a coercive function, so does $\Phi(\mathbf{z}, \w)$.
	Then, the sufficient descent property of $\Phi(\mathbf{z}, \w)$ surely brings the boundedness of the sequence $\{\mathbf{z}^k\}_{k\in\mathbb{N}}$ by using the coercive property of $\Phi(\mathbf{z}, \w)$.
	In conclusion, we have finished the proof of the Proposition \ref{Converg_Ana:main_theorem} so far.
	$\blacksquare$
\end{proof}

\subsection{Proof for Theorem \ref{Converg_Ana:key_theorem}}
The proof of the main theorem, i.e., Theorem \ref{Converg_Ana:key_theorem} contains two parts.
First, we need to give the proof on establishing the uniformized K{\L} property \cite{bolte2014proximal}.
Then, together with the assertions in Proposition \ref{Converg_Ana:main_theorem}, we can prove the main convergence result.
\begin{proof}
	
	(\emph{Uniformized K{\L} property})
	
	From the sufficient descent property of $\Phi(\mathbf{z}, \w)$, we have
	\begin{equation}
	\sum_{t = 0}^{N-1} \|\z^{t+1}-\z^t\|^2 \geq \frac{1}{a}(\Phi(\z^0, \z^0) - \Phi(\z^N, \z^{N-1})),
	\end{equation}
	for a positive integer $N$.
	Since $\Phi(\mathbf{z}, \w)$ is bounded from below, we have $\lim_{t\rightarrow \infty}\|\mathbf{z}^{t+1}-\mathbf{z}^t\|=0$ by taking the limit as $N\rightarrow \infty$.
	On the other hand, from $\|\mathbf{e}_x^{t+1}\|\leq C_x\|\mathbf{x}^{t}-\mathbf{x}^{t-1}\|$ and $\|\mathbf{e}_y^{t+1}\|\leq C_y\|\mathbf{y}^{t}-\mathbf{y}^{t-1}\|$, we have $\mathbf{e}_x^{t+1} \rightarrow \mathbf{0}$ and $\mathbf{e}_y^{t+1} \rightarrow \mathbf{0}$ as $t\rightarrow\infty$.

	Furthermore, through denoting $\mathbf{P}_x^t$, $\mathbf{P}_y^t$, $\mathbf{P}_p^t$ and $\mathbf{P}_q^t$ for different cases of TECU:
	
	For ``$1-6$'':\\
	$\mathbf{P}_x^t := \nabla_{\x} H(\x^t, \y^t)- \nabla_{\x} H(\x^t, \y^{t-1}) + \zeta_1^{t-1}(\x^{t-1}-\x^{t})$, $\mathbf{P}_y^t :=\e_y^{t} + \eta_2(\y^{t-1}-\y^{t})+ \frac{2(C_y)^2}{\eta_2}(\y^t-\y^{t-1})$,\\
	$\mathbf{P}_p^t :=\mathbf{0}$, $\mathbf{P}_q^t:=\frac{2(C_y)^2}{\eta_2}(\y^{t-1}-\y^{t})$;\\
	
	For ``$2-6$'':\\
	$\mathbf{P}_x^t := \nabla_{\x} H(\x^t, \y^t)- \nabla_{\x} H(\x^t, \y^{t-1}) + \gamma_1^{t-1}(\x^{t-1}-\x^{t})$, $\mathbf{P}_y^t :=\e_y^{t} + \eta_2(\y^{t-1}-\y^{t})+ \frac{2(C_y)^2}{\eta_2}(\y^t-\y^{t-1})$,\\
	$\mathbf{P}_p^t :=\mathbf{0}$, $\mathbf{P}_q^t:=\frac{2(C_y)^2}{\eta_2}(\y^{t-1}-\y^{t})$;\\
	
	For ``$3-4$'':\\
	$\mathbf{P}_x^t := \nabla_{\x} H(\x^t, \y^t)- \nabla_{\x} H(\x^t, \y^{t-1}) + \eta_1(\x^{t-1}-\x^{t}) + \e_x^{t} + \frac{2(C_x)^2}{\eta_1}(\x^t-\x^{t-1})$,
	$\mathbf{P}_y^t :=\zeta_2^{t-1}(\y^{t-1}-\y^{t})$, \\
	$\mathbf{P}_p^t :=\frac{2(C_x)^2}{\eta_1}(\x^{t-1}-\x^{t})$, $\mathbf{P}_q^t:=\mathbf{0}$;\\
	
	For ``$3-5$'':\\
	$\mathbf{P}_x^t := \nabla_{\x} H(\x^t, \y^t)- \nabla_{\x} H(\x^t, \y^{t-1}) + \eta_1(\x^{t-1}-\x^{t}) + \e_x^{t} + \frac{2(C_x)^2}{\eta_1}(\x^t-\x^{t-1})$,
	$\mathbf{P}_y^t :=\gamma_2^{t-1}(\y^{t-1}-\y^{t})$, \\$\mathbf{P}_p^t :=\frac{2(C_x)^2}{\eta_1}(\x^{t-1}-\x^{t})$, $\mathbf{P}_q^t:=\mathbf{0}$;\\
	
	For ``$3-6$'':\\
	$\mathbf{P}_x^t := \nabla_{\x} H(\x^t, \y^t)- \nabla_{\x} H(\x^t, \y^{t-1}) + \eta_1(\x^{t-1}-\x^{t}) + \e_x^{t} + \frac{2(C_x)^2}{\eta_1}(\x^t-\x^{t-1})$,
	$\mathbf{P}_y^t :=\e_y^{t} + \eta_2(\y^{t-1}-\y^{t})+ \frac{2(C_y)^2}{\eta_2}(\y^t-\y^{t-1})$,\\ $\mathbf{P}_p^t :=\frac{2(C_x)^2}{\eta_1}(\x^{t-1}-\x^{t})$, $\mathbf{P}_q^t:=\frac{2(C_y)^2}{\eta_2}(\y^{t-1}-\y^{t})$;\\
	
	there obviously has $\mathbf{P}^t:=(\mathbf{P}_x^t, \mathbf{P}_y^t, \mathbf{P}_p^t, \mathbf{P}_q^t) \in \partial\Phi(\mathbf{z}^k, \z^{k-1})$ and $\mathbf{P}^t\rightarrow \mathbf{0}$ as $t\rightarrow\infty$.
	
	Since $\{\mathbf{z}^t\}_{t\in\mathbb{N}}$ is bounded, then there exists a subsequence $\{\mathbf{z}^{t_l}\}_{l\in\mathbb{N}}$ such that $\mathbf{z}^{t_l}\rightarrow \mathbf{z}^{\ast}$ as $l\rightarrow\infty$.
	
	By letting step $t$ as $t_l-1$, then for the case ``3-6'' we have
	\begin{equation}
	\begin{aligned}
	&f(\x^{t_l}) + H(\x^{t_l}, \y^{t_l-1}) + \frac{\eta_1^{t_l-1}}{2}\|\x^{t_l}-\x^{t_l-1}\|^2 - (\e_x^{t_l})^{\top}\x^{t_l} \\
	\leq&  f(\x^{\ast}) + H(\x^{\ast}, \y^{t_l-1}) + \frac{\eta_1^{t_l-1}}{2}\|\x^{\ast}-\x^{t_l-1}\|^2 - (\e_x^{t_l})^{\top}\x^{\ast},
	\end{aligned}
	\end{equation}
	By taking $l\rightarrow\infty$, we have the property:
	\begin{equation}
	\limsup_{l\rightarrow\infty} f(\mathbf{x}^{t_l}) \leq f(\mathbf{x}^{\ast}).
	\end{equation}
	Similarly, we also have
	\begin{equation}
	\limsup_{l\rightarrow\infty} g(\mathbf{y}^{t_l}) \leq g(\mathbf{y}^{\ast}).
	\end{equation}
	Then with the lower semi-continuous properties of functions $f(\mathbf{x})$ and $g(\mathbf{y})$ , we have
	\begin{equation}
	\lim_{l\rightarrow\infty} f(\mathbf{x}^{t_l})= f(\mathbf{x}^{\ast}) \ \mbox{and}\  \lim_{l\rightarrow\infty} g(\mathbf{y}^{t_l})= g(\mathbf{y}^{\ast}),
	\end{equation}
	which further indicate $\Phi(\mathbf{z}^{t_l}, \mathbf{z}^{t_l-1}) \rightarrow \Phi(\mathbf{z}^{\ast}, \mathbf{z}^{\ast})$ as $l\rightarrow\infty$.
	In a similar way, we can easily get that this property holds for all the cases of TECU.

	Then from the closedness property of limiting sub-differential \cite{bolte2014proximal}, we have $\mathbf{0} \in \Phi(\mathbf{z}^{\ast}, \mathbf{z}^{\ast})$, which indicates that $(\mathbf{z}^{\ast}, \mathbf{z}^{\ast})$ is a critical point of $\Phi(\mathbf{z}, \w)$.
	
	Moreover, since $\Phi(\mathbf{z}, \w)$ is bounded from below and is also sufficient descent, $\Phi(\mathbf{z}^t, \mathbf{z}^{t-1})$ has limit value as $t\rightarrow \infty$.
	Together with $\Phi(\mathbf{z}^{t_l}, \mathbf{z}^{t_l-1}) \rightarrow \Phi(\mathbf{z}^{\ast}, \mathbf{z}^{\ast})$, we have that $\Phi(\mathbf{z}, \w)$ is finite and constant on the set of all limit points of the sequence $\{\mathbf{z}^t\}_{t\in\mathbb{N}}$.
	Thus the uniformized K{\L} lemma in \cite{bolte2014proximal} is established.
	
	(\emph{Cauchy convergence})
	
	Since the uniformized K{\L} property is established, then there exists $t>T_l$ such that
	\begin{equation}
	\phi'(\Phi(\z^t, \z^{t-1})- \Phi(\z^{\ast}, \z^{\ast}))dist(0, \partial \Phi(\z^t, \z^{t-1})) \geq 1.
	\end{equation}
	Denoting $\triangle_{t, t+1} := \phi(\Phi(\z^t, \z^{t-1})- \Phi(\z^{\ast}, \z^{\ast})) - \phi(\Phi(\z^{t+1}, \z^{t})- \Phi(\z^{\ast}, \z^{\ast}))$,  we have
	\begin{equation}
	\begin{aligned}
	\triangle_{t, t+1} \geq& \phi'(\Phi(\z^t, \z^{t-1})- \Phi(\z^{\ast}, \z^{\ast}))(\Phi(\z^t, \z^{t-1}) - \Phi(\z^{t+1}, \z^{t}))\\
	\geq& \frac{\Phi(\z^t, \z^{t-1}) - \Phi(\z^{t+1}, \z^{t})}{dist(0, \partial \Phi(\z^t, \z^{t-1}))} \geq \frac{a\|\z^{t+1}-\z^t\|^2}{b(\|\z^t-\z^{t-1}\| + \|\z^{t-1}-\z^{t-2}\|)}.
	\end{aligned}
	\end{equation}
	This is equal to
	\begin{equation}
	4\|\z^{t+1}-\z^t\|^2 \leq 4 \mu\triangle_{t, t+1}(\|\z^t-\z^{t-1}\| + \|\z^{t-1}-\z^{t-2}\|),
	\end{equation}
	where $\mu = \frac{b}{a}$.
	Then from the triangle inequality, we have
	\begin{equation}
	4\|\z^{t+1}-\z^t\| \leq \|\z^t-\z^{t-1}\| + \|\z^{t-1}-\z^{t-2}\| + 4 \mu\triangle_{t, t+1}.
	\end{equation}
	Summing up the above inequality for $t = T_l+1, \ldots, T_L$ yields
	\begin{equation}
	\begin{aligned}
	4\sum_{t = T_l+1}^{T_L}  \|\z^{t+1}-\z^t\| \leq& \sum_{t = T_l+1}^{T_L} \|\z^t-\z^{t-1}\| + \|\z^{t-1}-\z^{t-2}\| + 4 \mu\triangle_{t, t+1}\\
	\leq& 2\|\z^{T_l+1}-\z^{T_l}\|  + \|\z^{T_l}-\z^{T_l-1}\| + 4 \mu\triangle_{T_l+1, T_L+1} + 2\sum_{t = T_l+1}^{T_L}  \|\z^{t+1}-\z^{t}\|.
	\end{aligned}
	\end{equation}
	Since $\phi \geq 0$, we have that for any $t > T_l$ that
	\begin{equation}
	\sum_{t = T_l+1}^{T_L}  \|\z^{t+1}-\z^t\| \leq \|\z^{T_l+1}-\z^{T_l}\| + \frac{1}{2}\|\z^{T_l}-\z^{T_l-1}\| + 2 \mu\triangle_{T_l+1, T_L+1}.
	\end{equation}
	This easily shows that the sequence $\{\z^t\}_{t\in\mathbb{N}}$ has finite length, that is,
	\begin{equation}
	\sum_{t = 0}^{\infty} \|\z^{t+1}-\z^t\| < \infty,
	\end{equation}
	which implies that $\{\z^t\}_{t\in\mathbb{N}}$ is a Cauchy sequence and hence converges to a critical point $\z^{\ast}$ of $\Psi(\z)$.
\end{proof}

\section{More Experimental Results on Low-light Image Enhancement}
In this section, we provide more experimental results (given in Fig. \ref{fig:supp_NASA}) on showing visual comparisons with state-of-the-art methods, on both NASA \cite{nasa} and Non-uniform \cite{wang2013naturalness} datasets.

\begin{figure*}[!htb]
	\centering
	\begin{tabular}{c@{\extracolsep{0.15em}}c@{\extracolsep{0.15em}}c@{\extracolsep{0.15em}}c@{\extracolsep{0.15em}}c@{\extracolsep{0.15em}}c}
		\includegraphics[width=.16\linewidth]{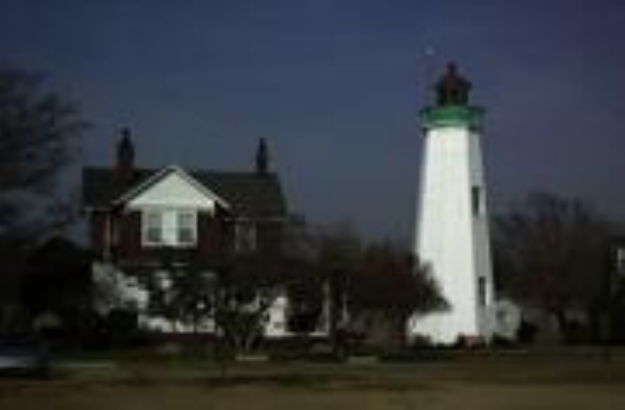}&
		\includegraphics[width=.16\linewidth]{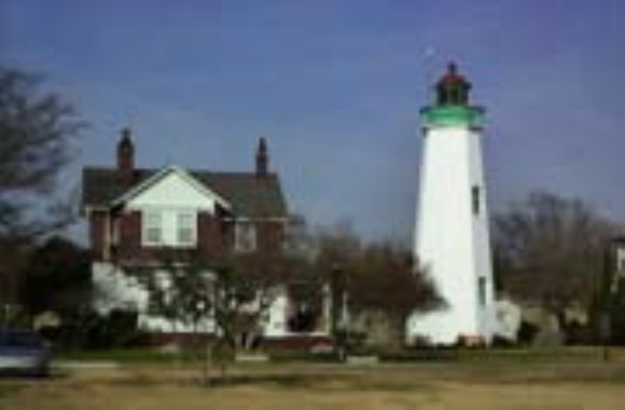}&
		\includegraphics[width=.16\linewidth]{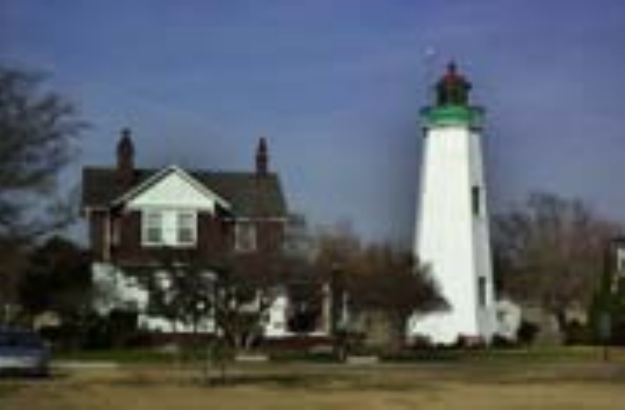}&
		\includegraphics[width=.16\linewidth]{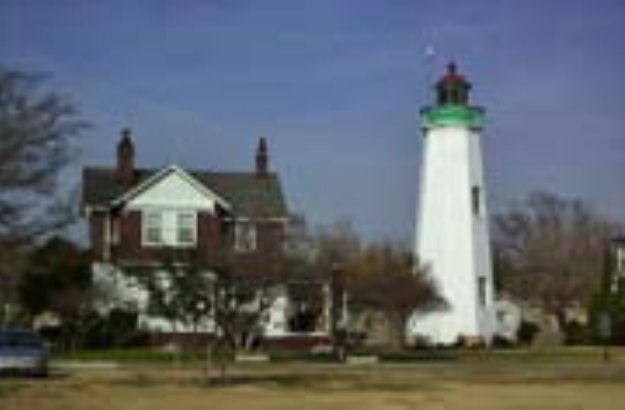}&
		\includegraphics[width=.16\linewidth]{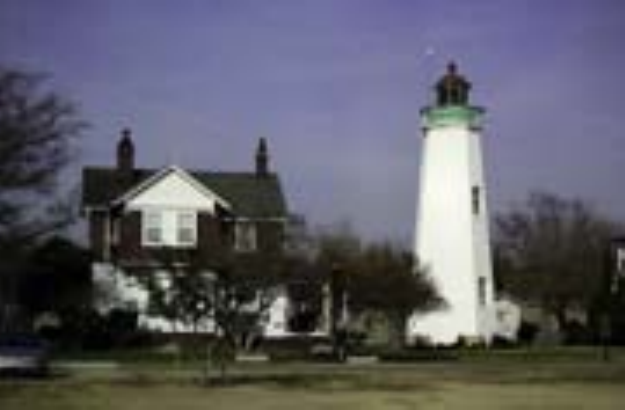}&
		\includegraphics[width=.16\linewidth]{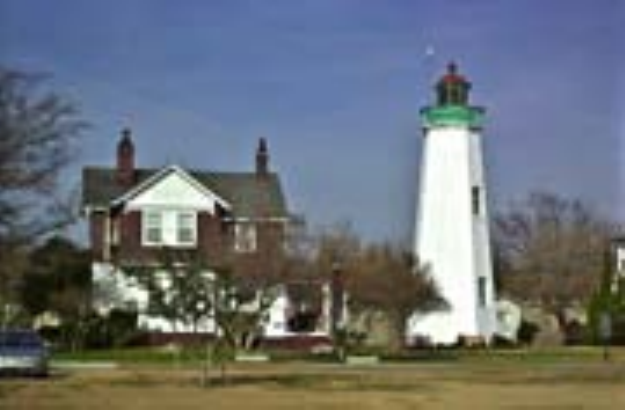}\\
		\small{3.97}&\small{3.10}&\small{3.23}&\small{3.26}&\small{3.31}&\small\textbf{2.56}\\
		\includegraphics[width=.16\linewidth]{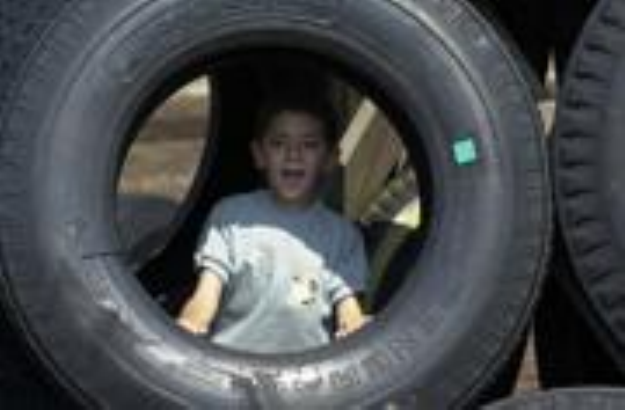}&
		\includegraphics[width=.16\linewidth]{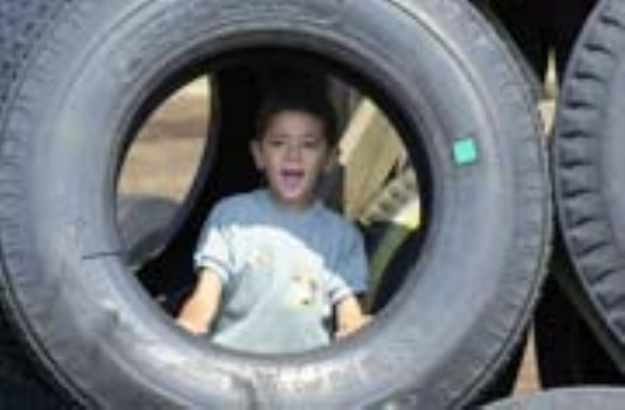}&
		\includegraphics[width=.16\linewidth]{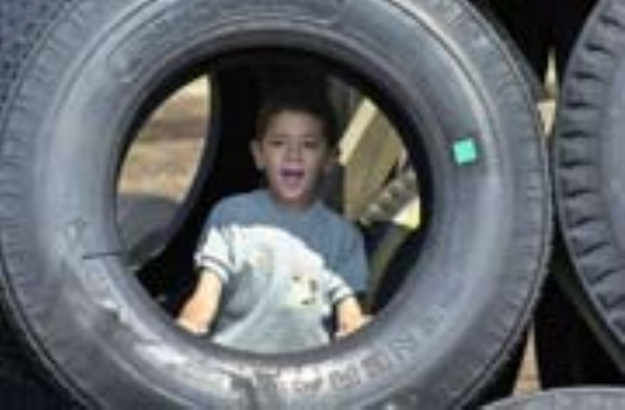}&
		\includegraphics[width=.16\linewidth]{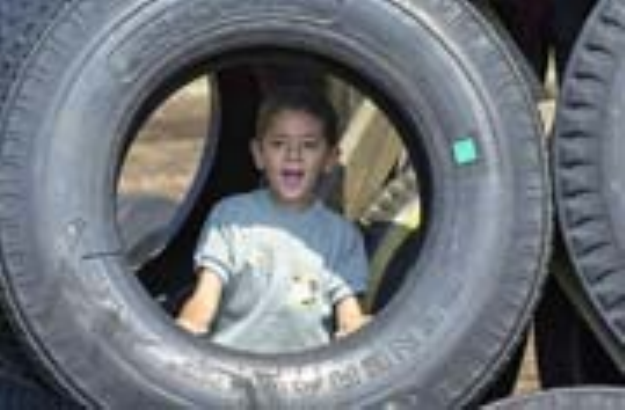}&
		\includegraphics[width=.16\linewidth]{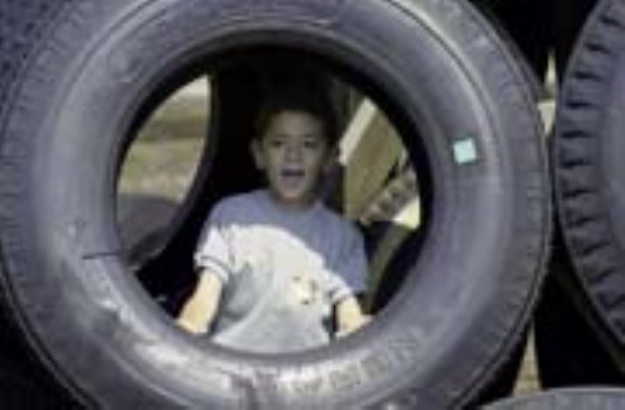}&
		\includegraphics[width=.16\linewidth]{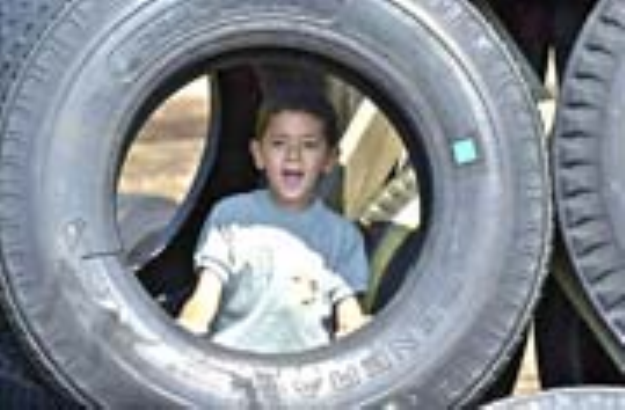}\\
		\small{3.79}&\small{3.64}&\small{3.71}&\small{4.06}&\small{3.70}&\small\textbf{3.49}\\
		\includegraphics[width=.16\linewidth]{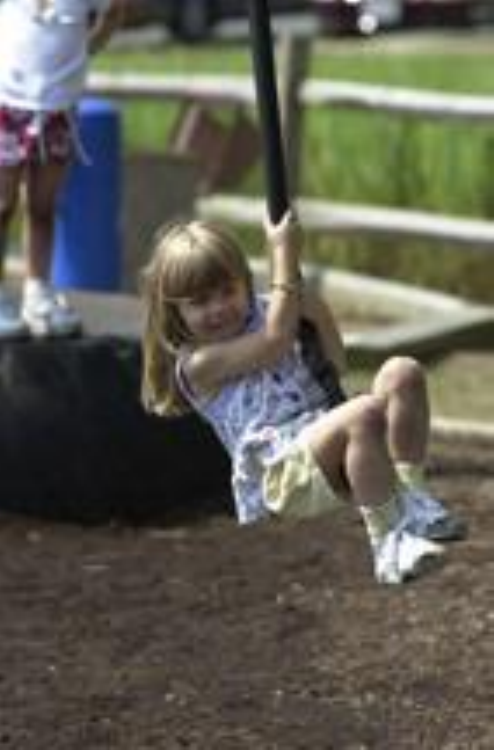}&
		\includegraphics[width=.16\linewidth]{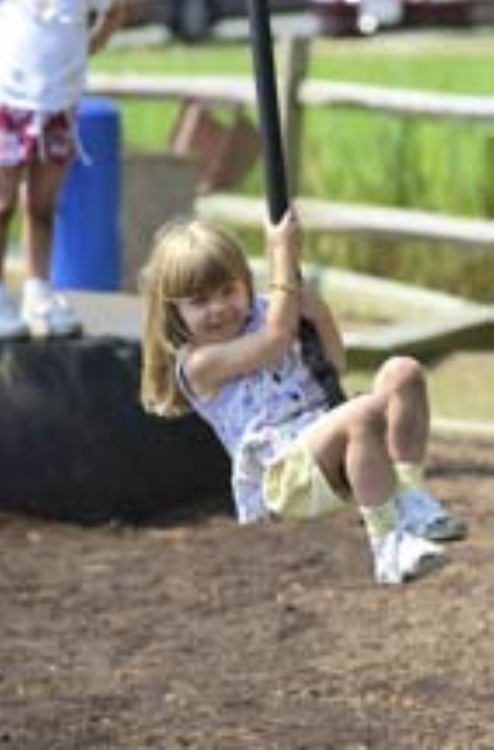}&
		\includegraphics[width=.16\linewidth]{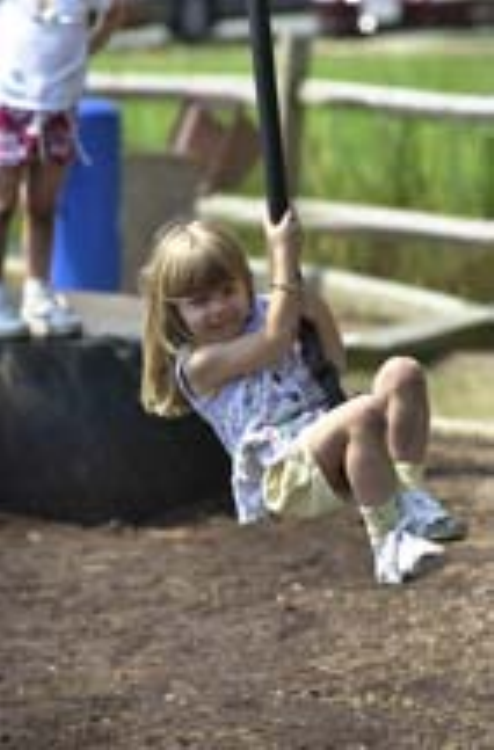}&
		\includegraphics[width=.16\linewidth]{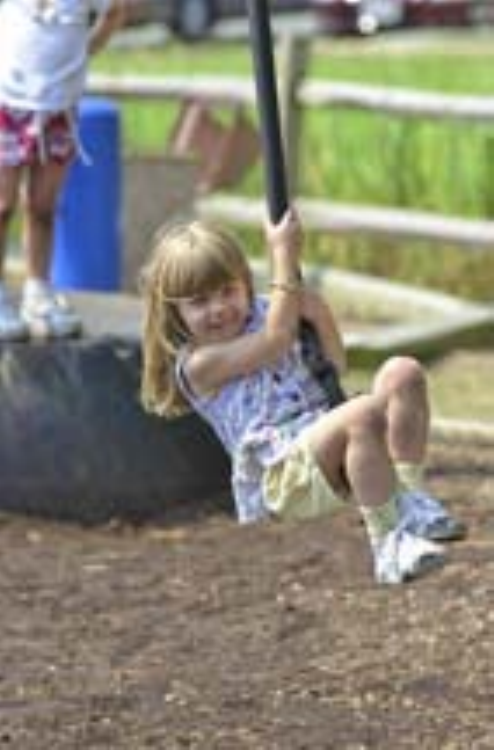}&
		\includegraphics[width=.16\linewidth]{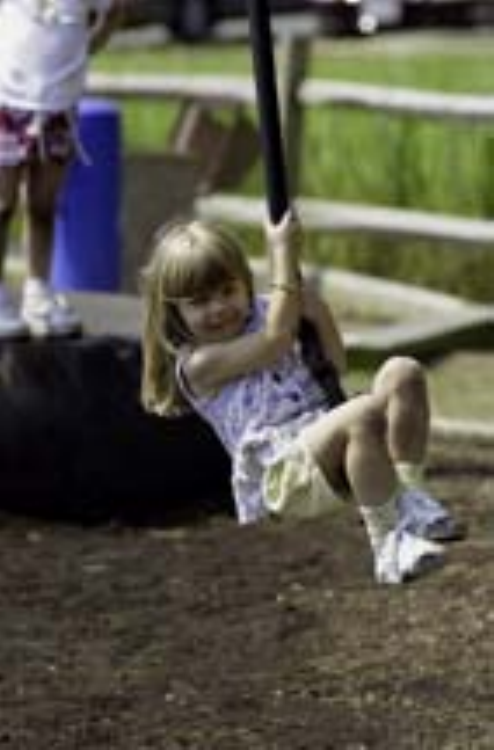}&
		\includegraphics[width=.16\linewidth]{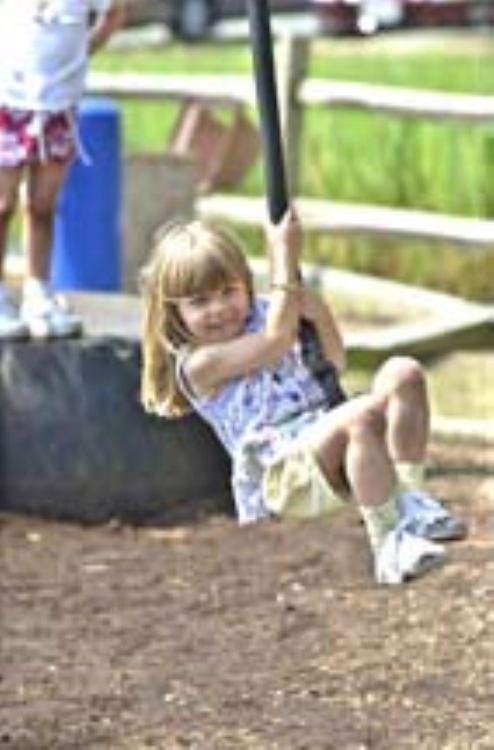}\\
		\small{3.42}&\small{3.29}&\small{3.34}&\small{3.26}&\small{3.25}&\small\textbf{2.89}\\
		\includegraphics[width=.16\linewidth]{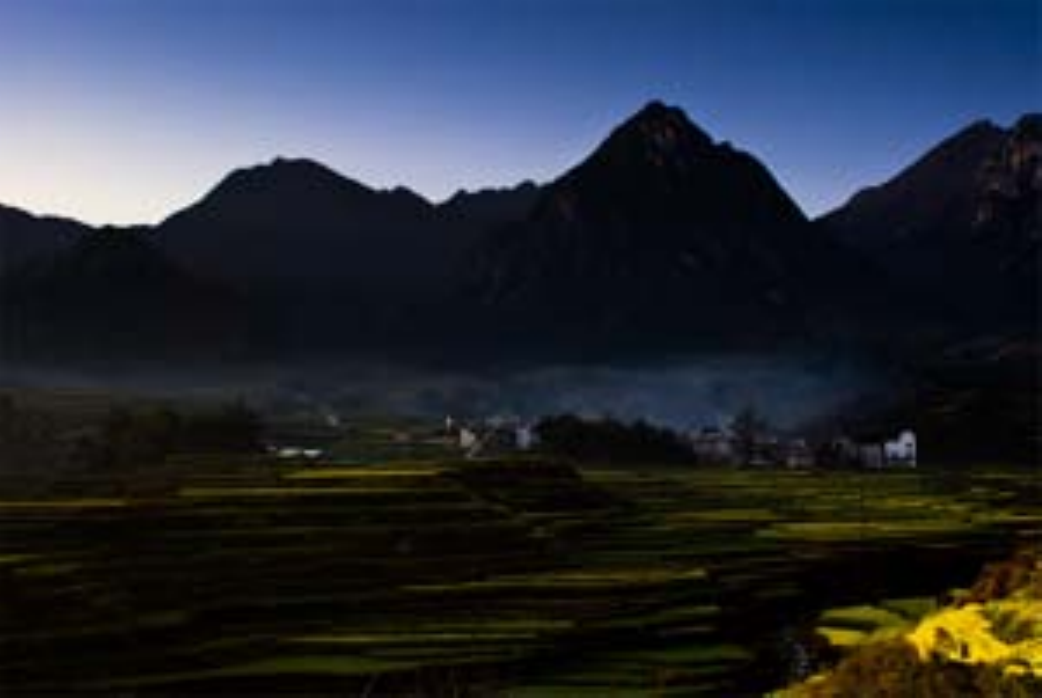}&
		\includegraphics[width=.16\linewidth]{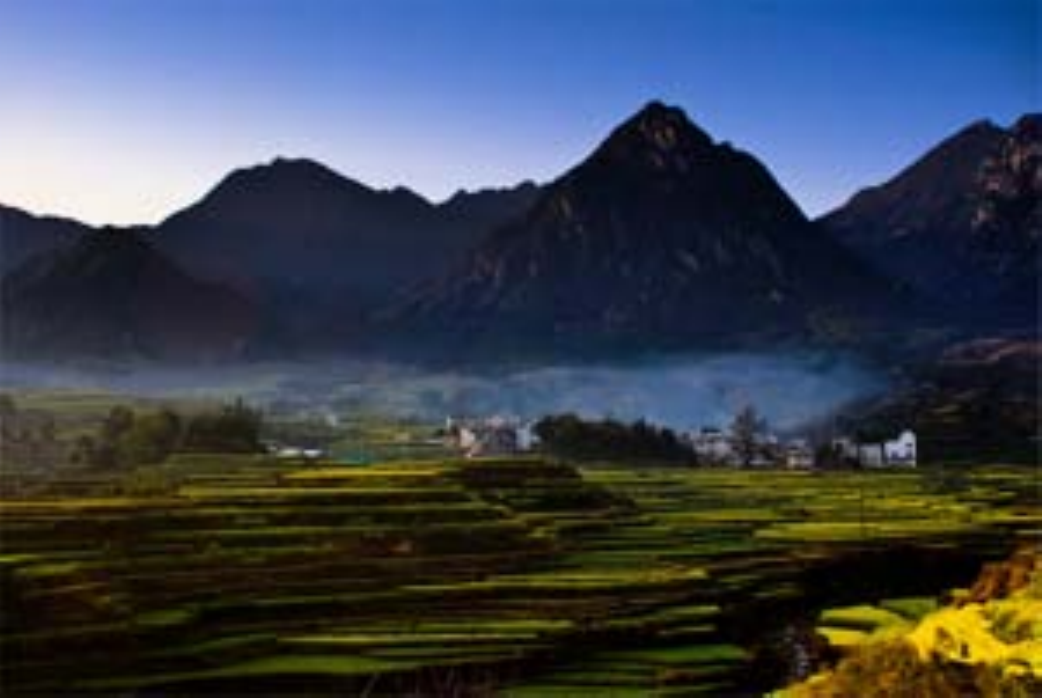}&
		\includegraphics[width=.16\linewidth]{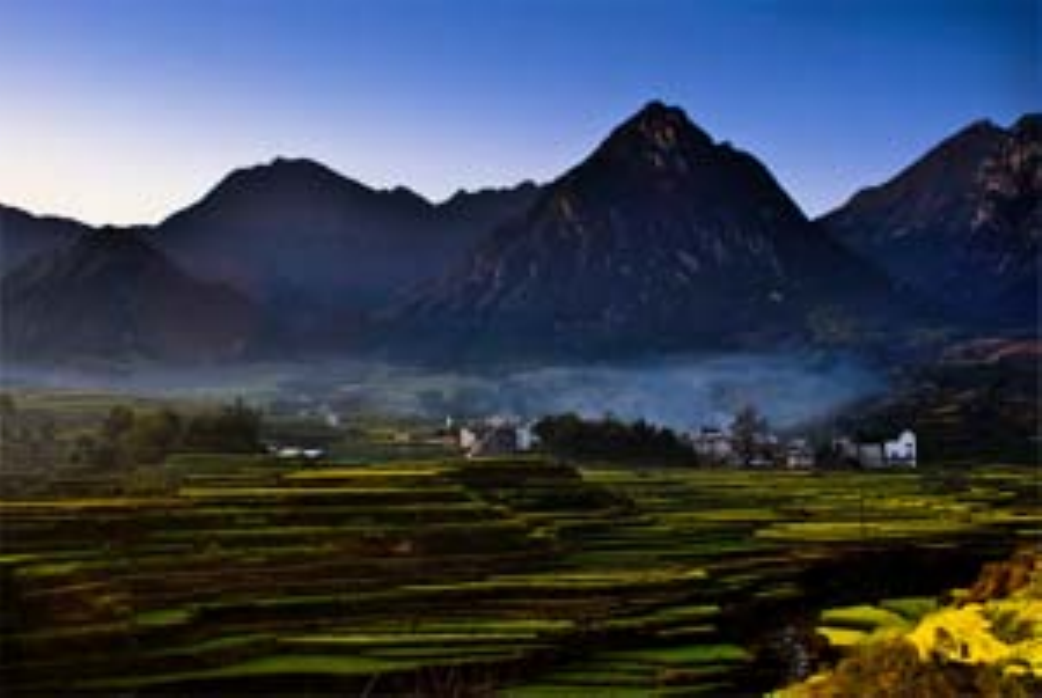}&
		\includegraphics[width=.16\linewidth]{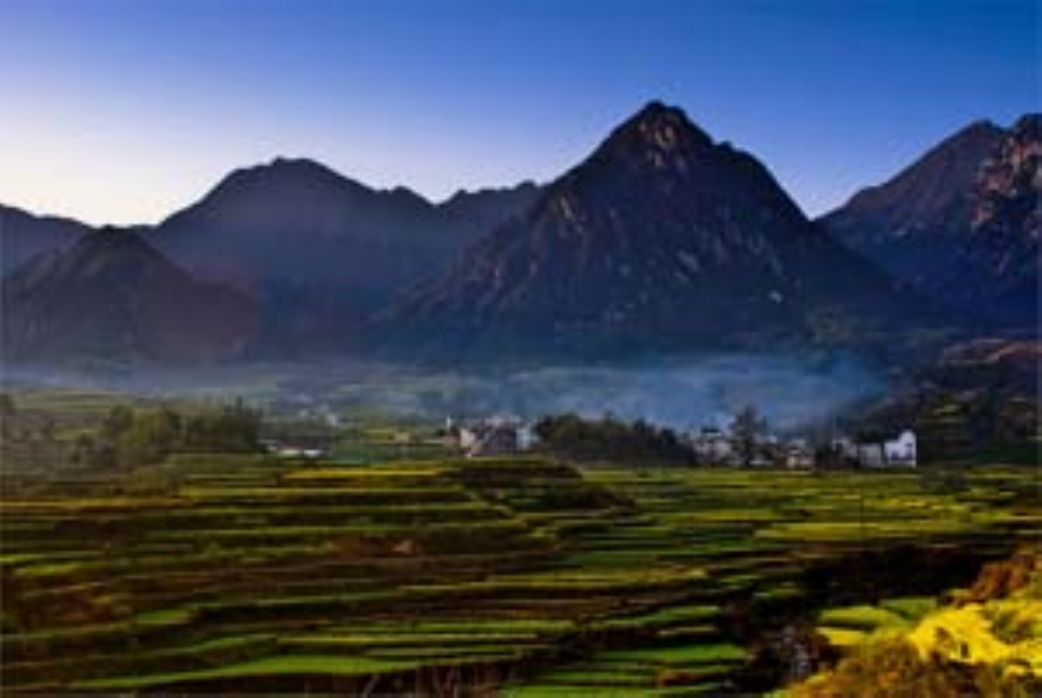}&
		\includegraphics[width=.16\linewidth]{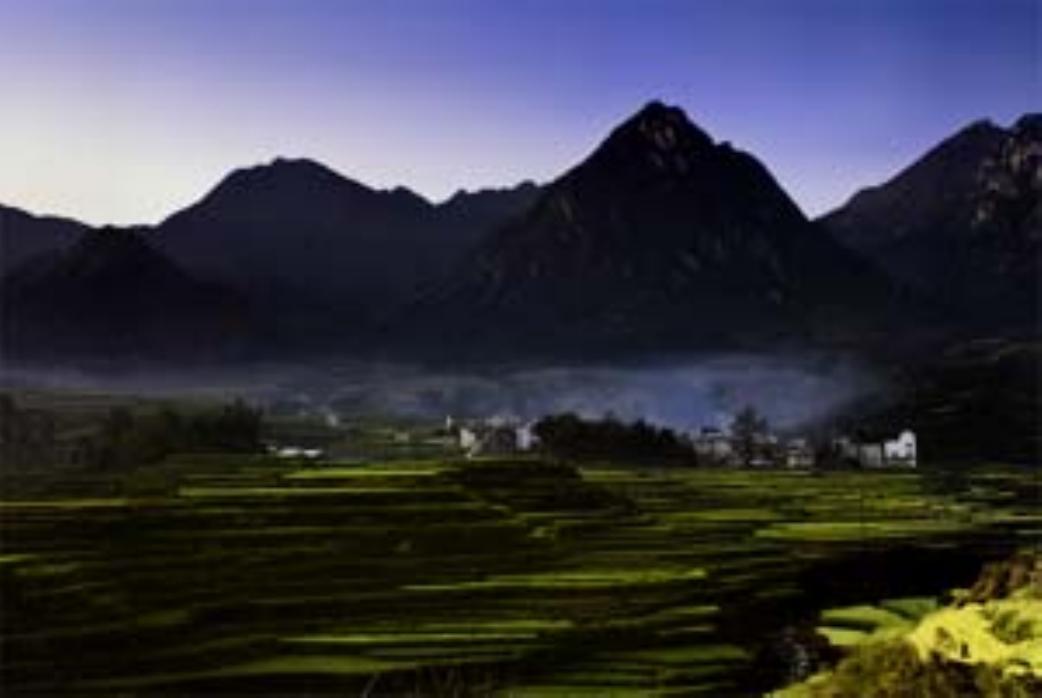}&
		\includegraphics[width=.16\linewidth]{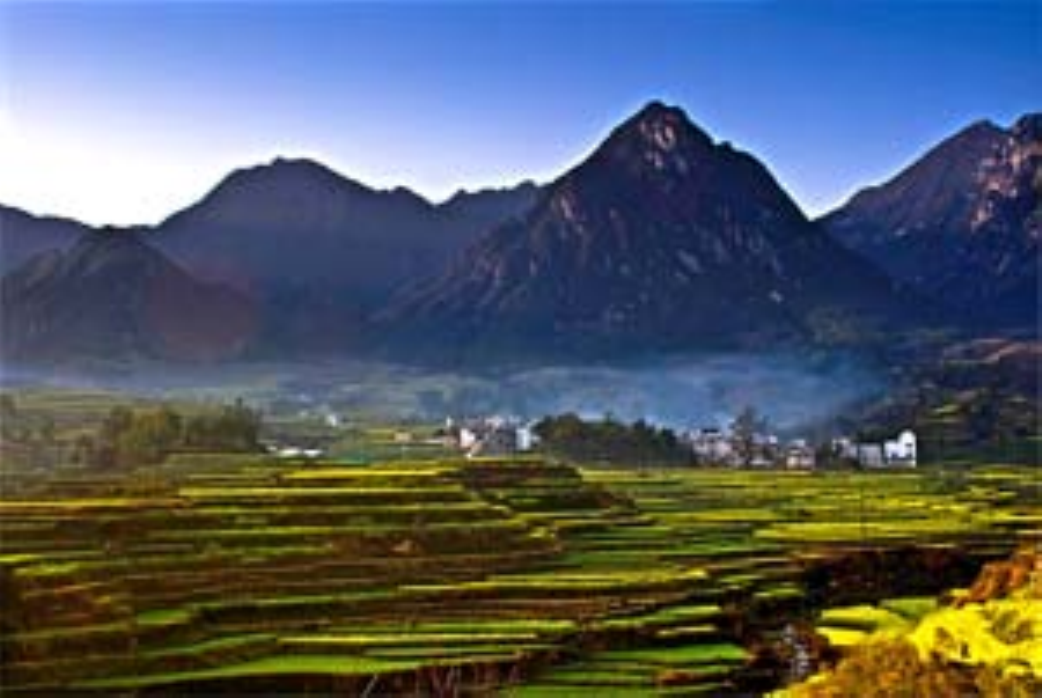}\\
		\small{3.83}&\small{3.16}&\small{3.01}&\small{2.96}&\small{3.28}&\small\textbf{2.92}\\
		\includegraphics[width=.16\linewidth]{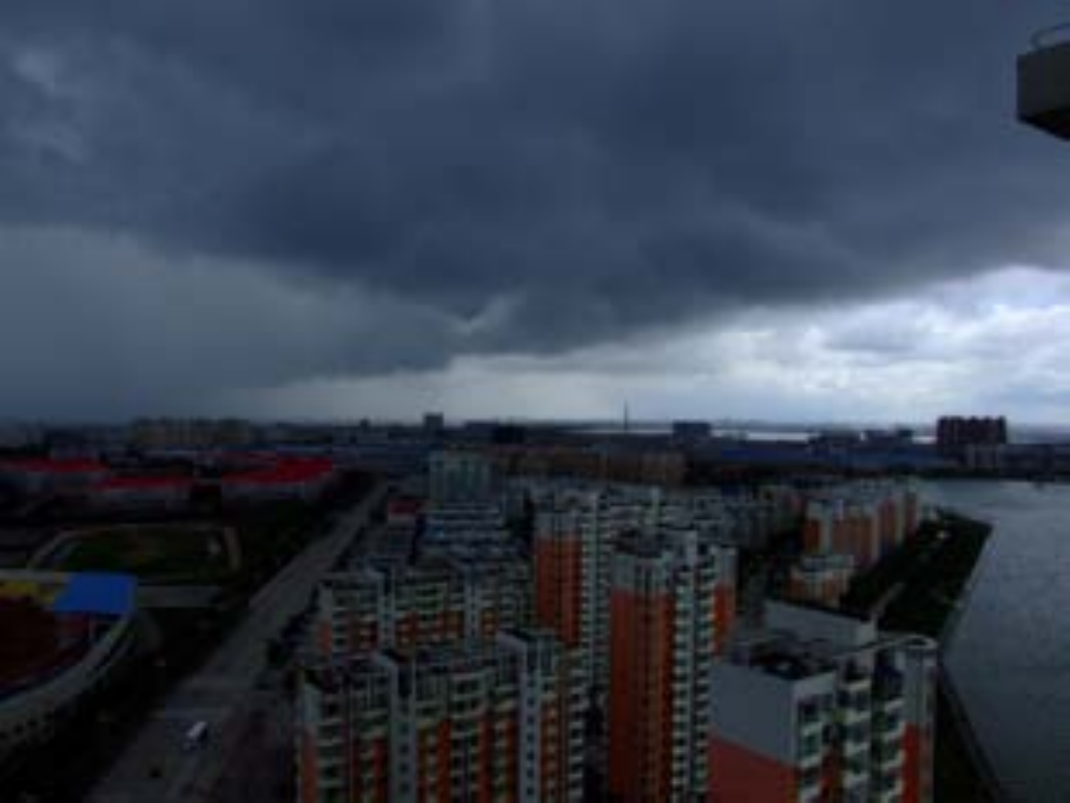}&
		\includegraphics[width=.16\linewidth]{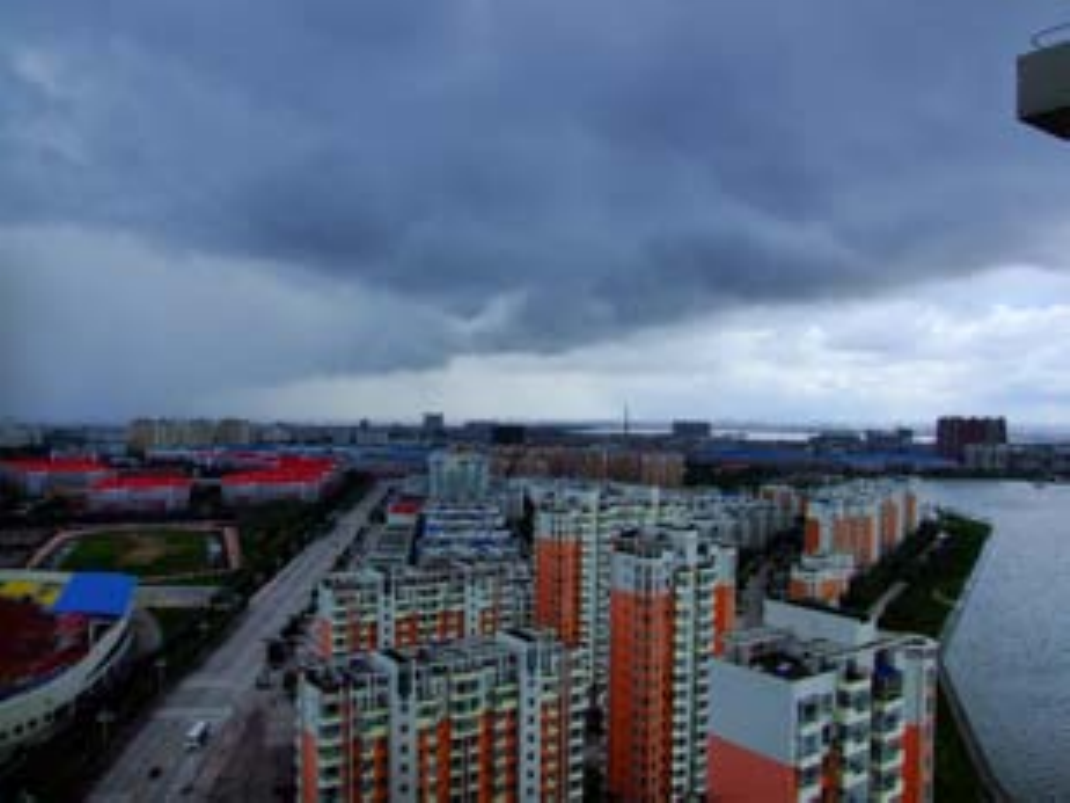}&
		\includegraphics[width=.16\linewidth]{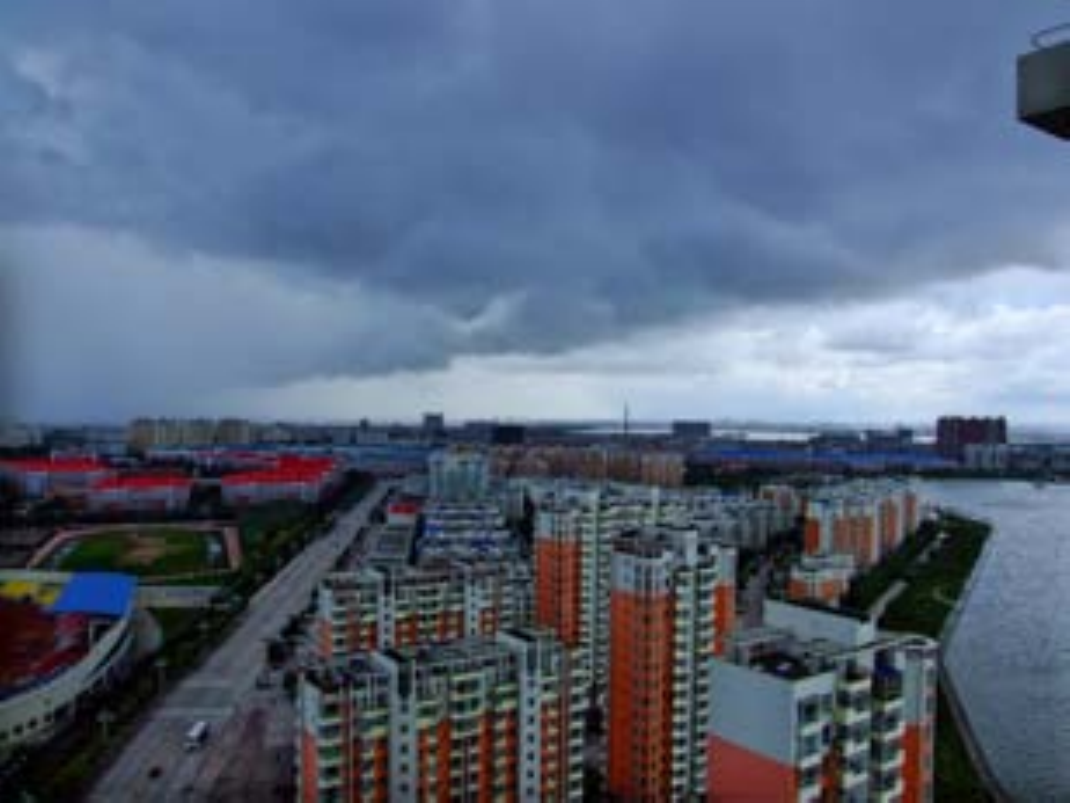}&
		\includegraphics[width=.16\linewidth]{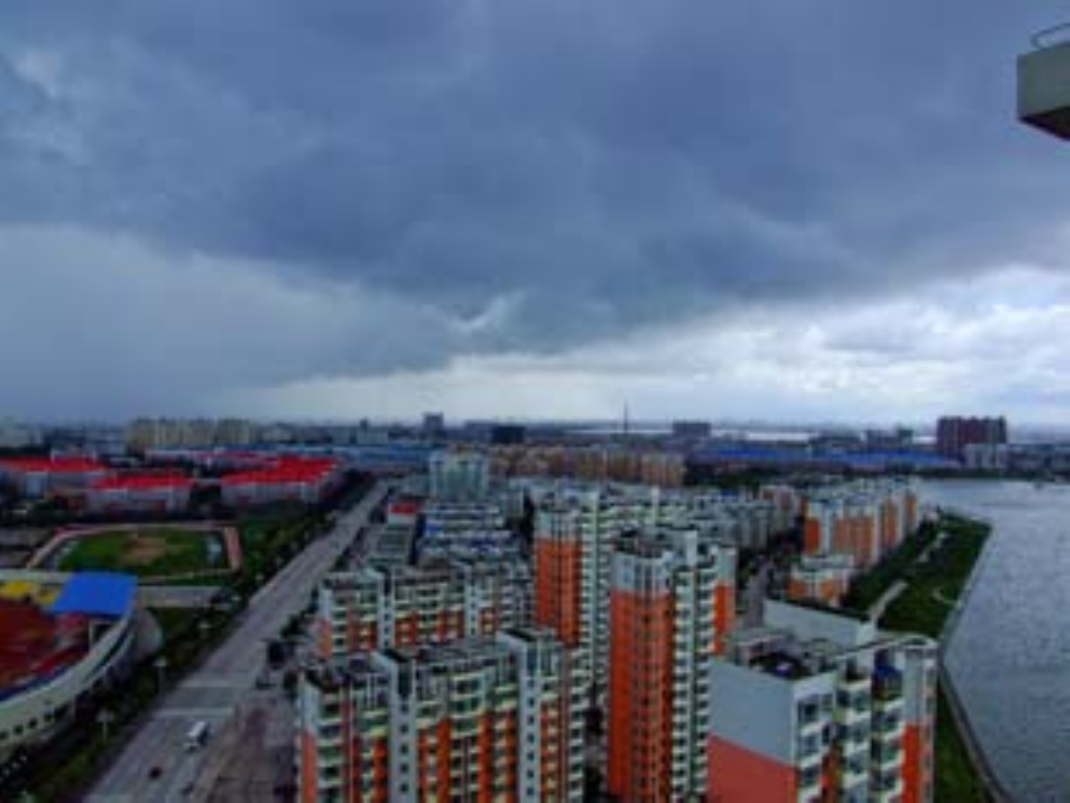}&
		\includegraphics[width=.16\linewidth]{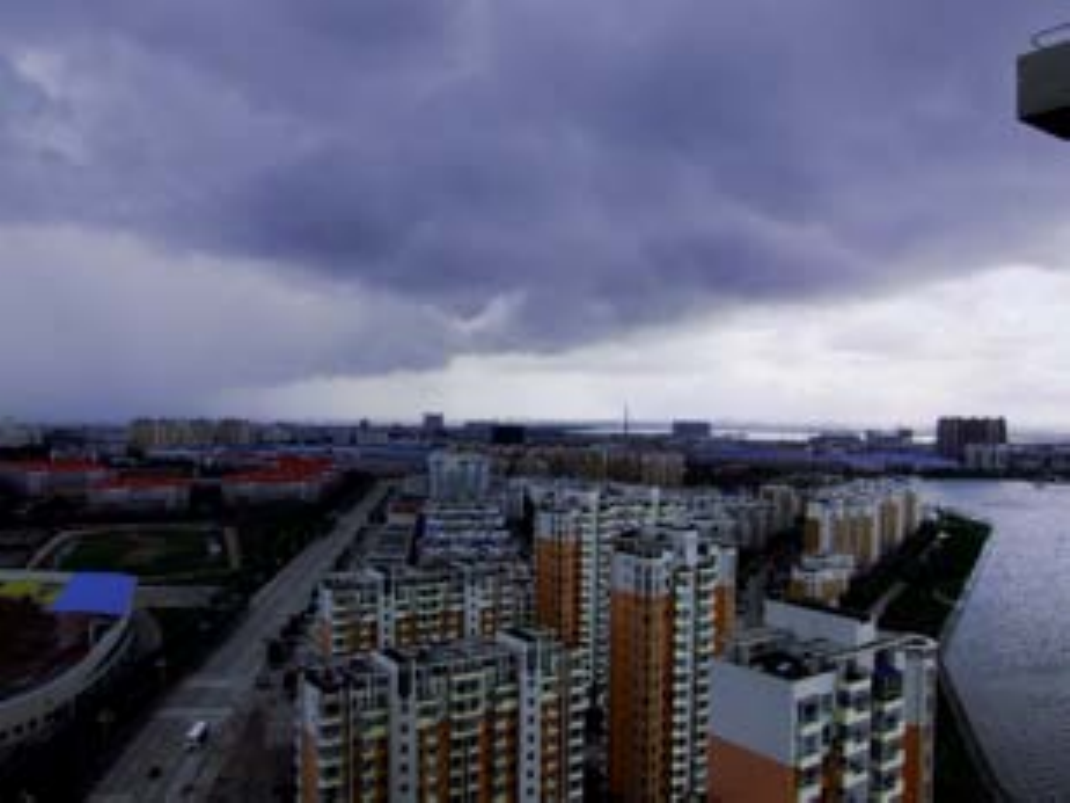}&
		\includegraphics[width=.16\linewidth]{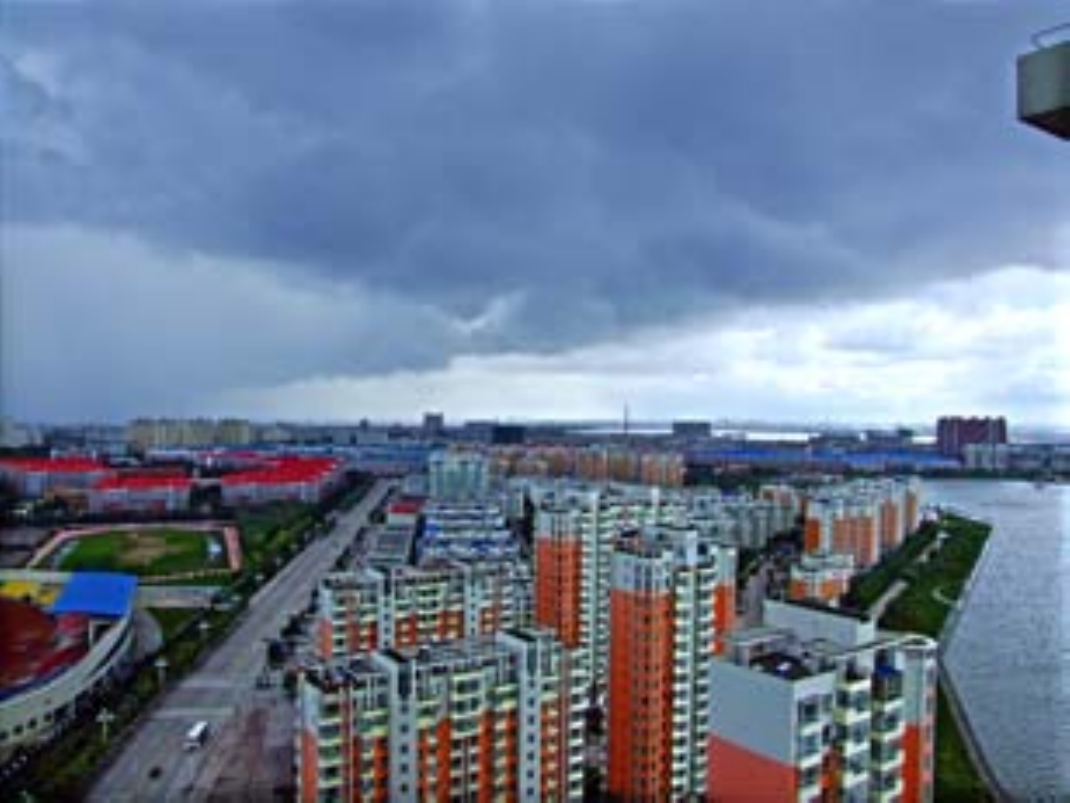}\\
		\small{3.14}&\small{2.85}&\small{2.62}&\small{2.59}&\small{2.86}&\small\textbf{2.37}\\
		\includegraphics[width=.16\linewidth]{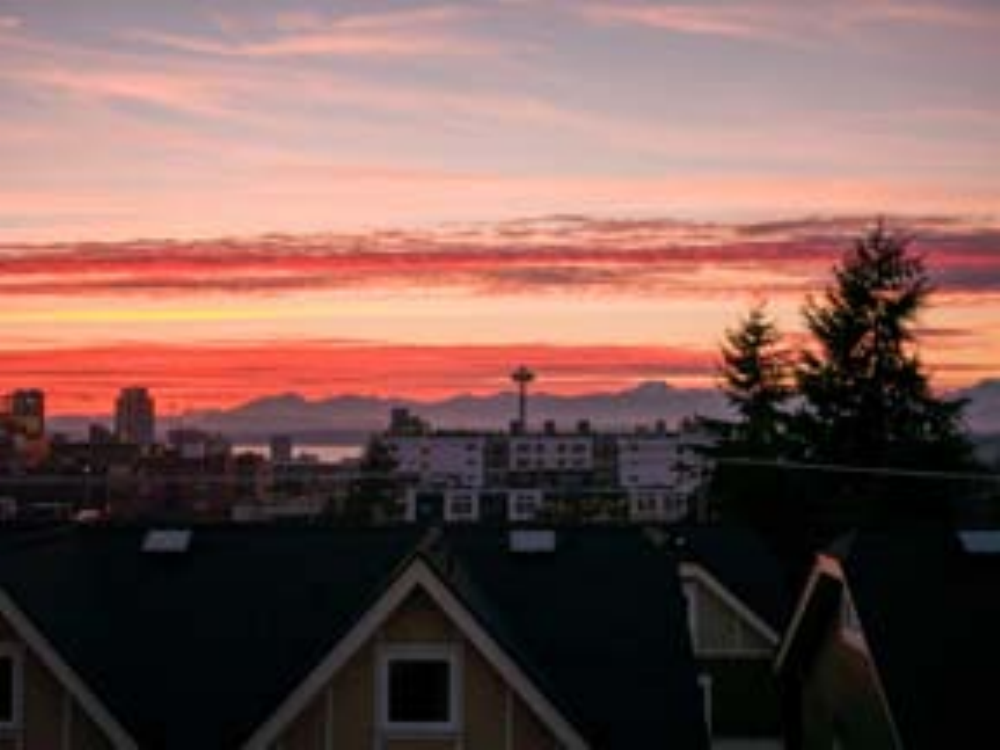}&
		\includegraphics[width=.16\linewidth]{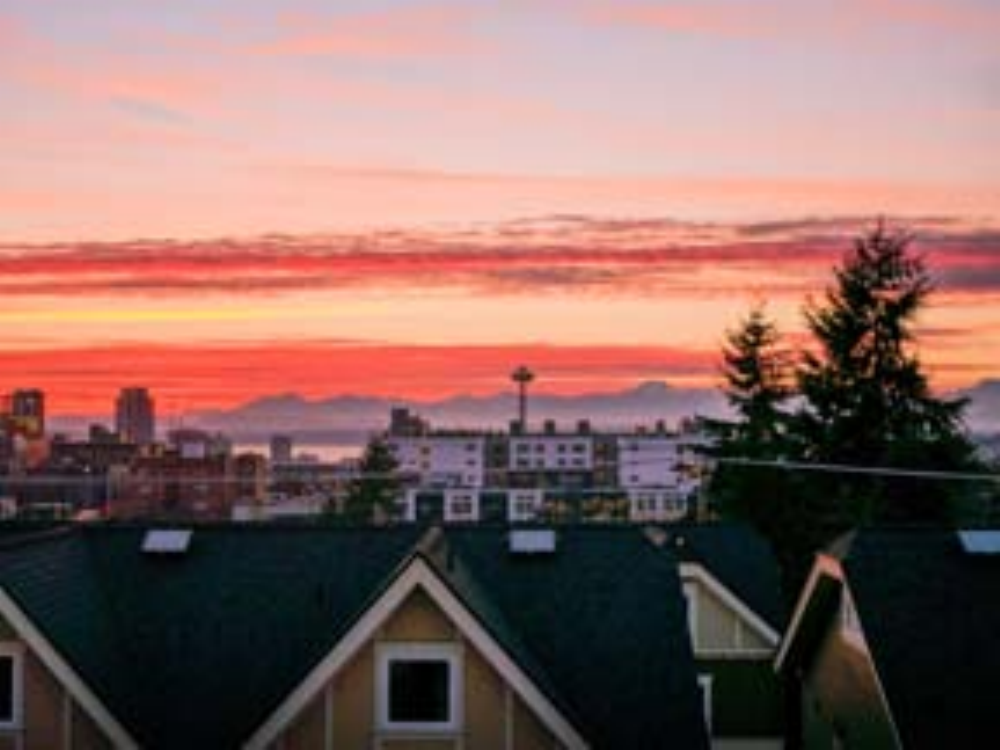}&
		\includegraphics[width=.16\linewidth]{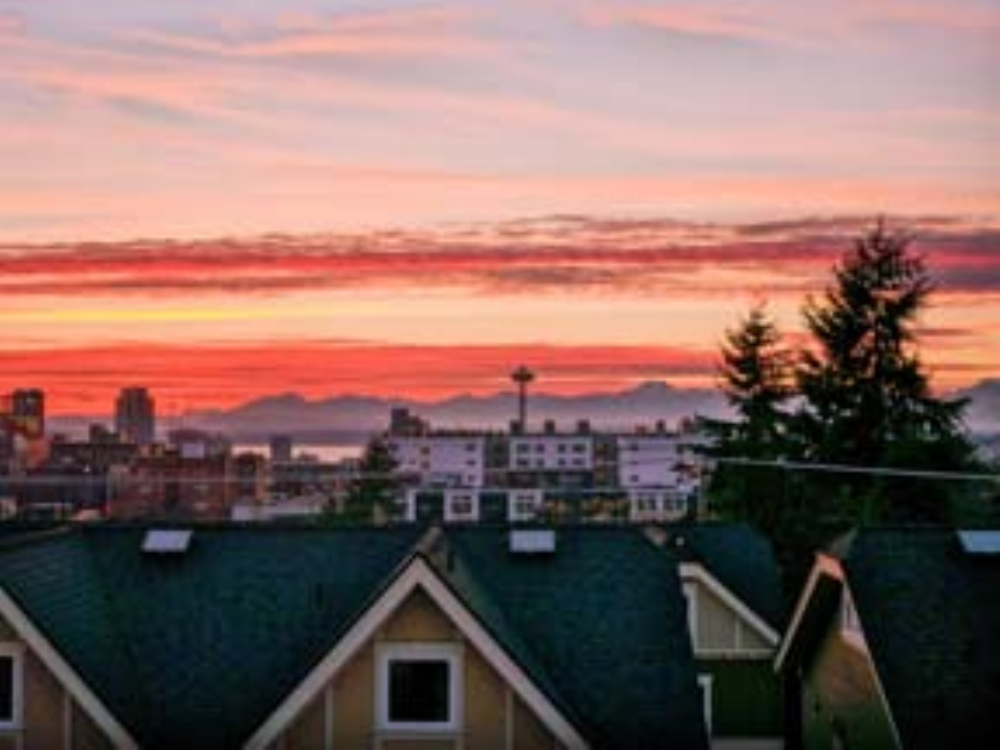}&
		\includegraphics[width=.16\linewidth]{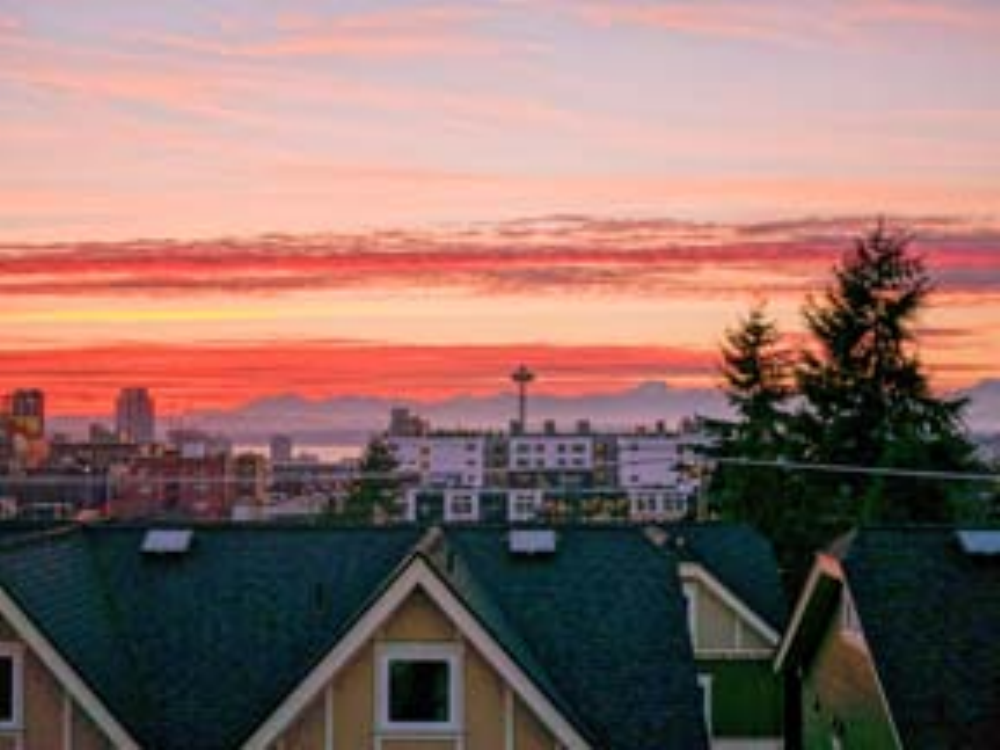}&
		\includegraphics[width=.16\linewidth]{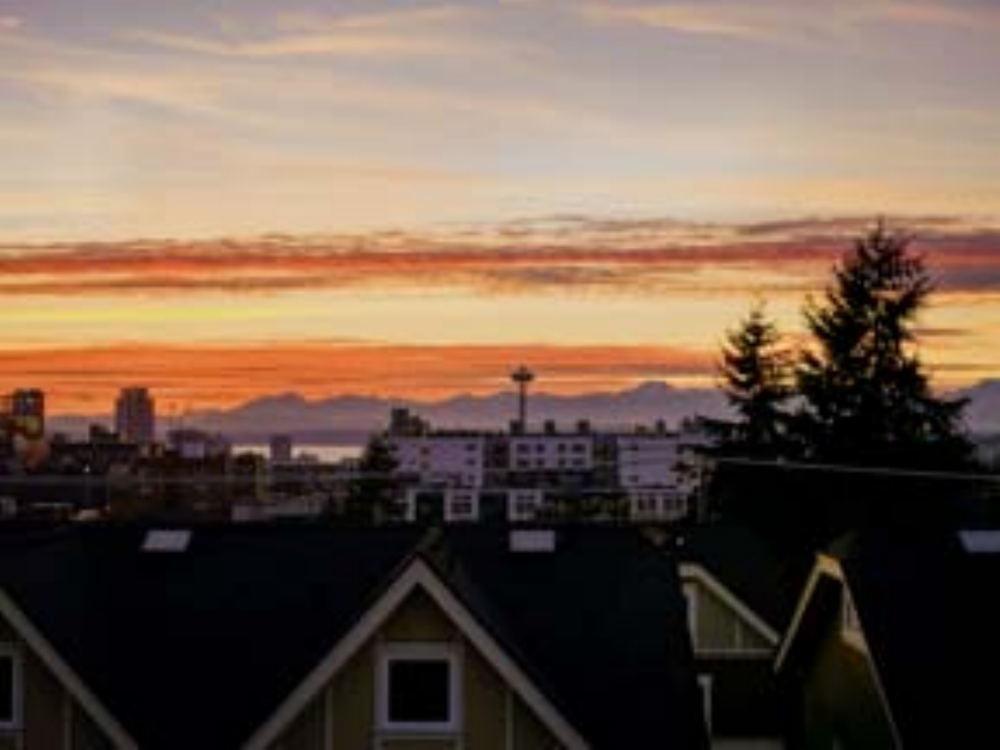}&
		\includegraphics[width=.16\linewidth]{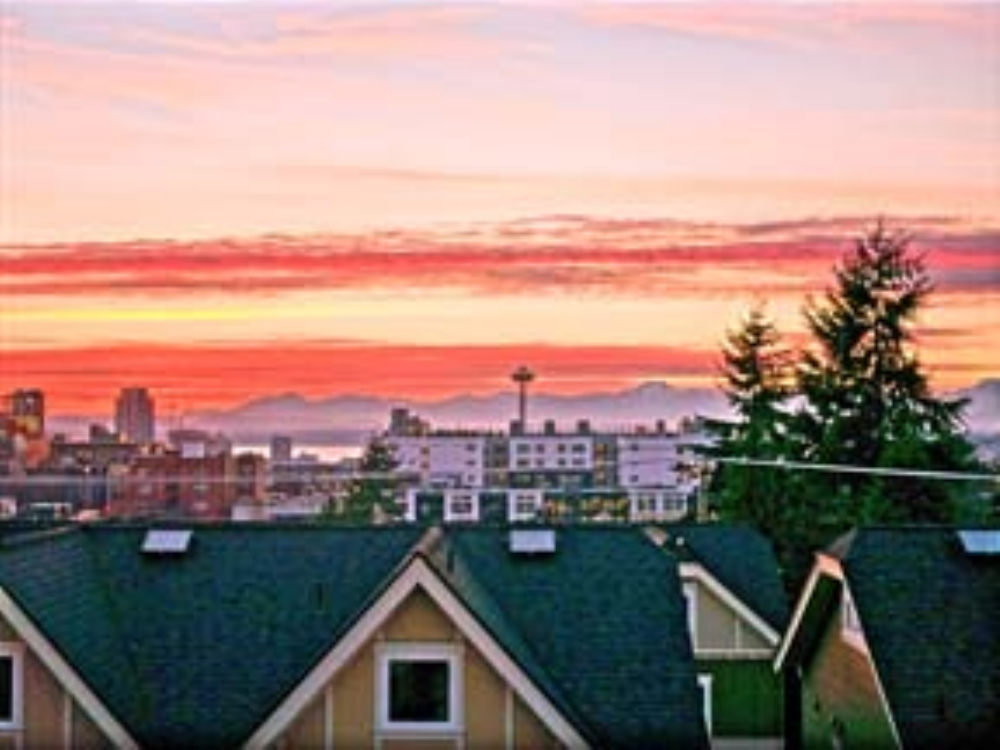}\\
		\small{2.88}&\small{2.64}&\small{2.55}&\small{2.55}&\small{2.96}&\small\textbf{2.49}\\
		\includegraphics[width=.16\linewidth]{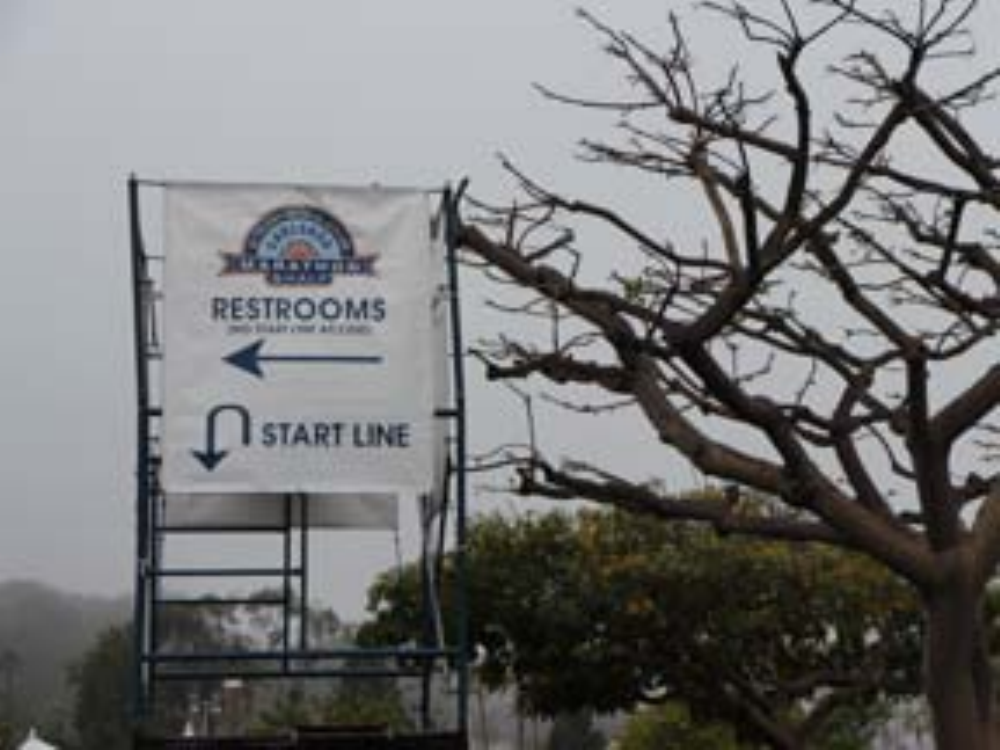}&
		\includegraphics[width=.16\linewidth]{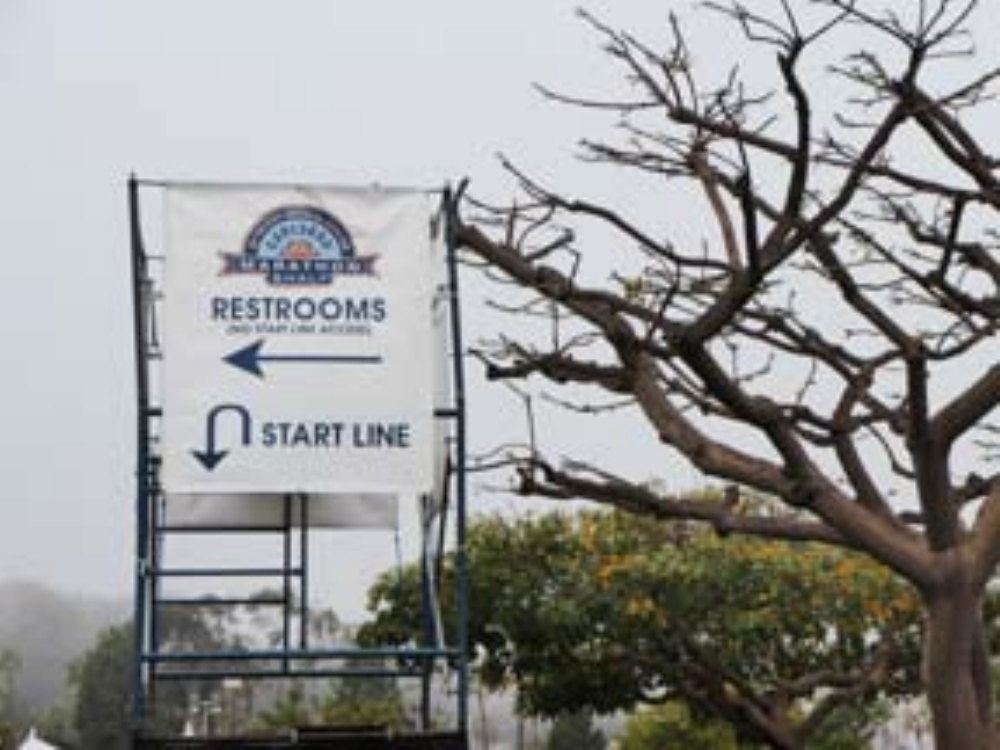}&
		\includegraphics[width=.16\linewidth]{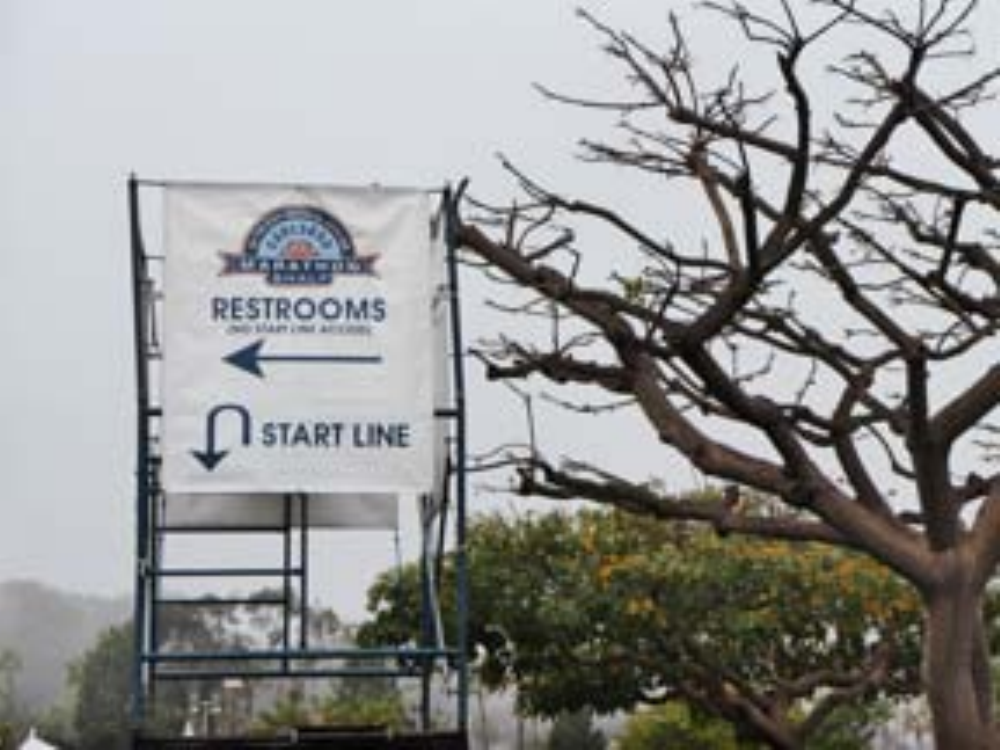}&
		\includegraphics[width=.16\linewidth]{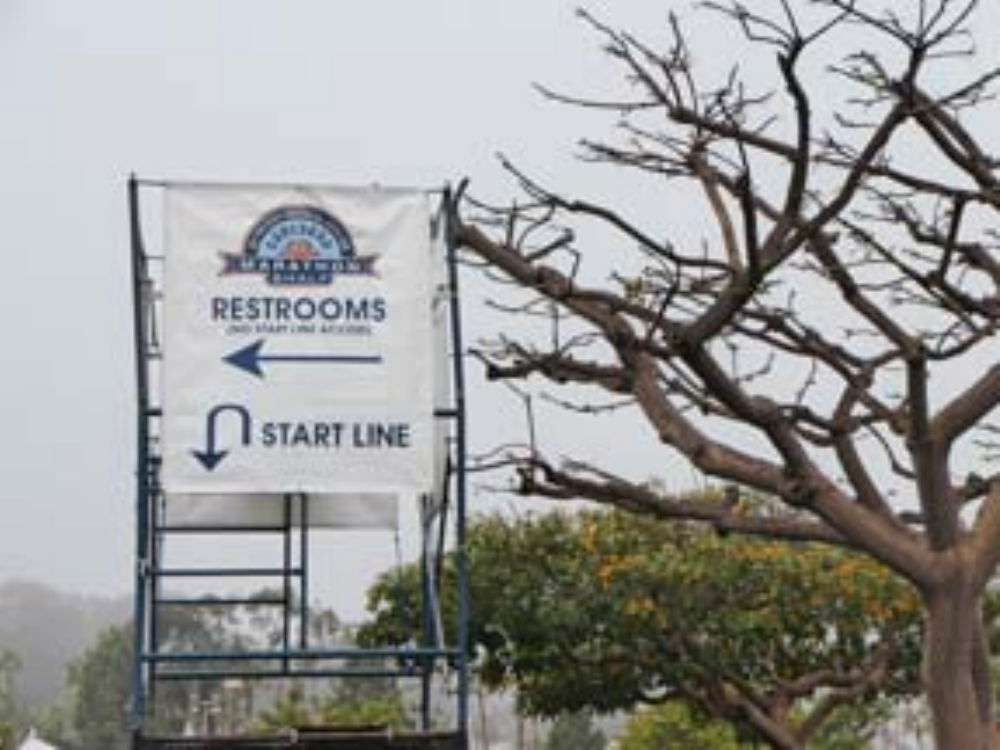}&
		\includegraphics[width=.16\linewidth]{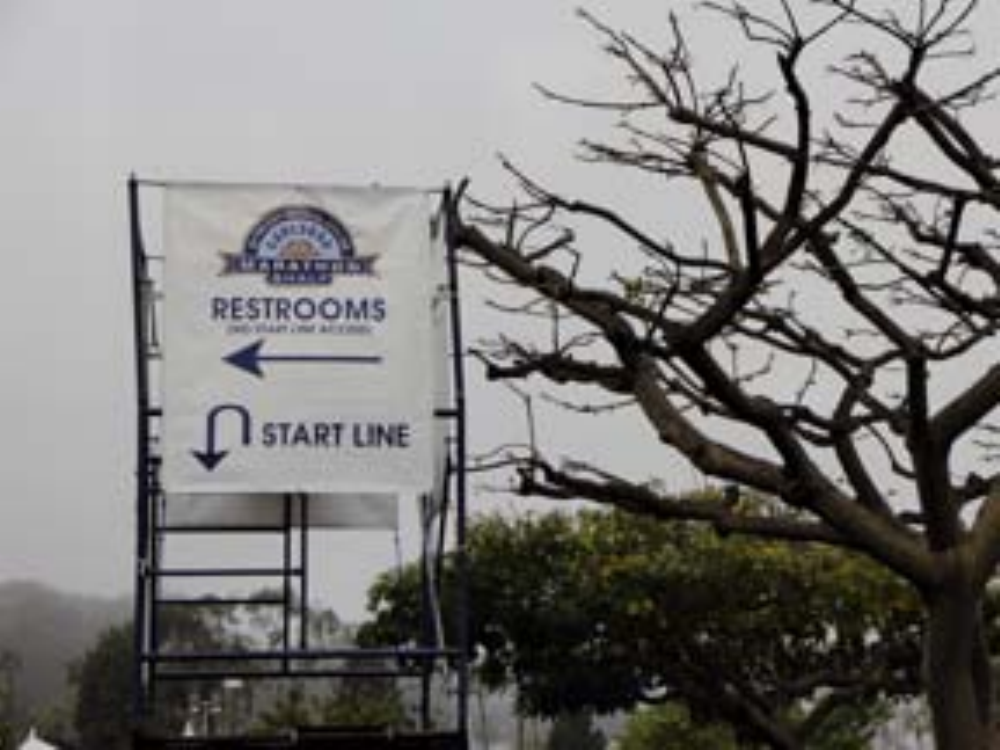}&
		\includegraphics[width=.16\linewidth]{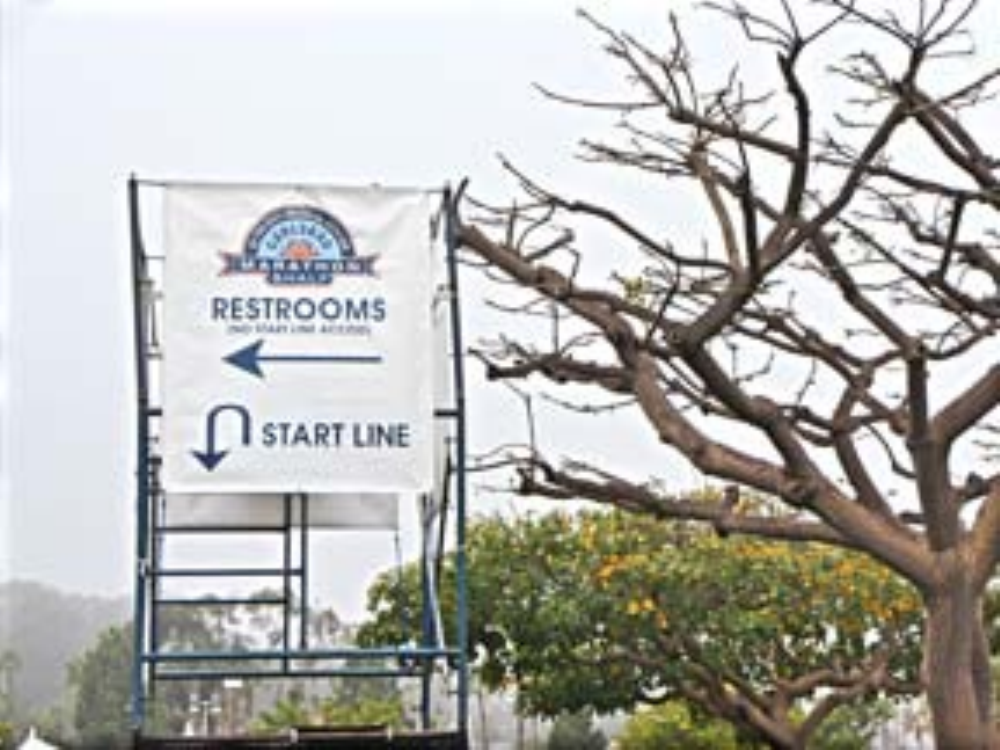}\\
		\small{3.19}&\small{3.45}&\small{3.36}&\small{2.92}&\small{3.22}&\small\textbf{2.76}\\
		\includegraphics[width=.16\linewidth]{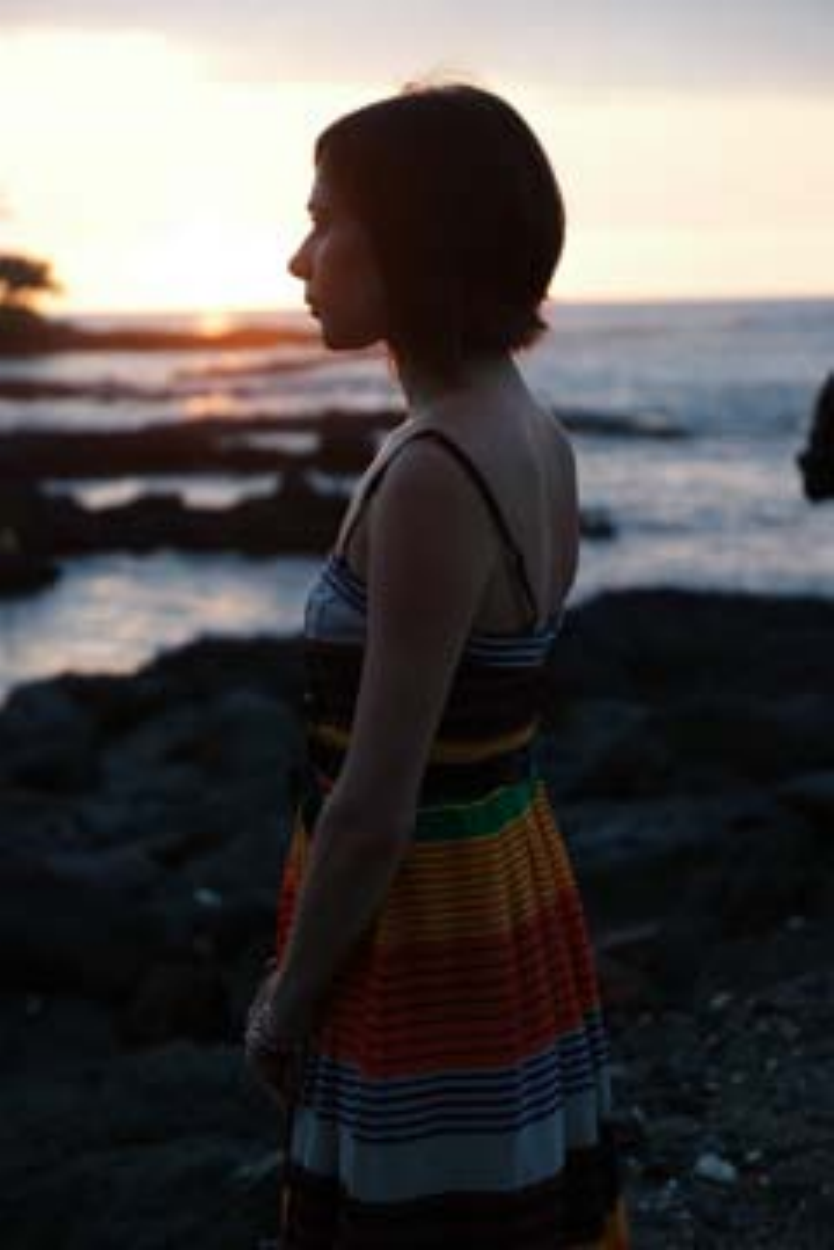}&
		\includegraphics[width=.16\linewidth]{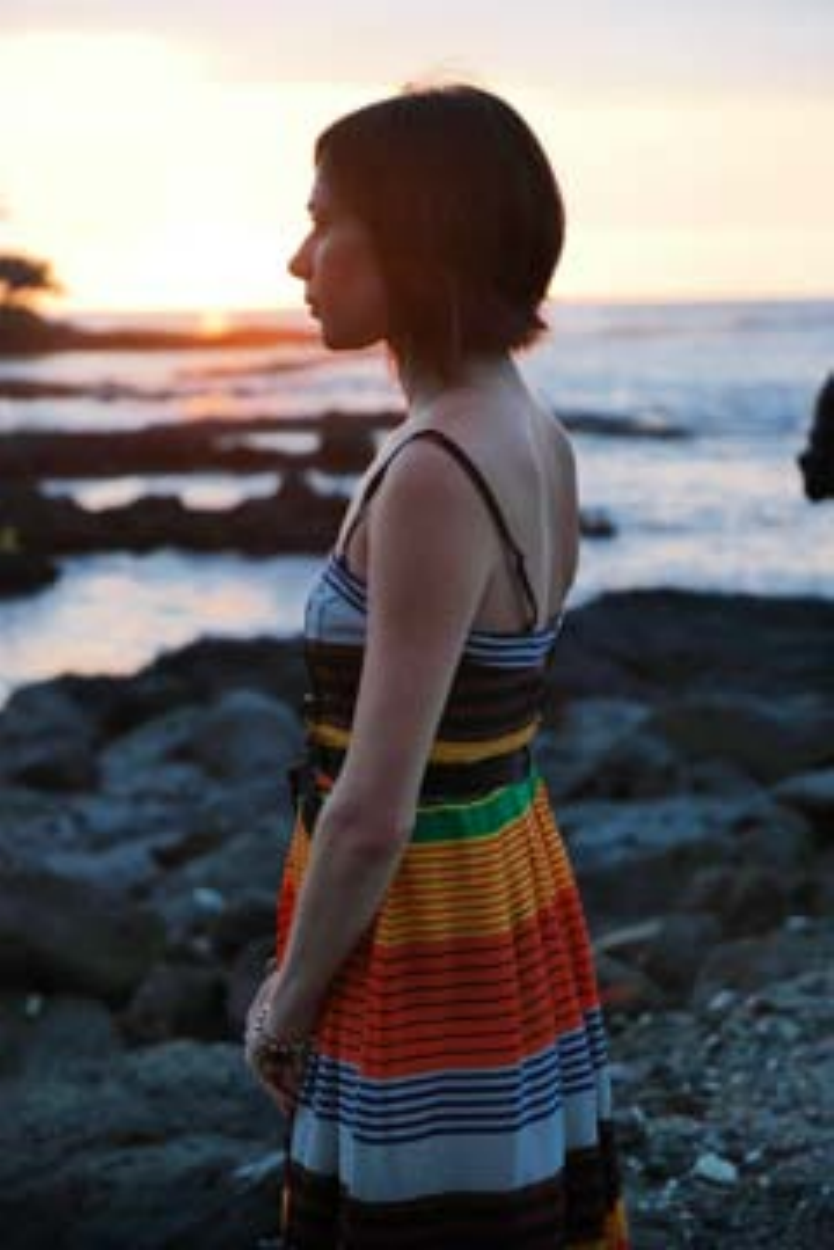}&
		\includegraphics[width=.16\linewidth]{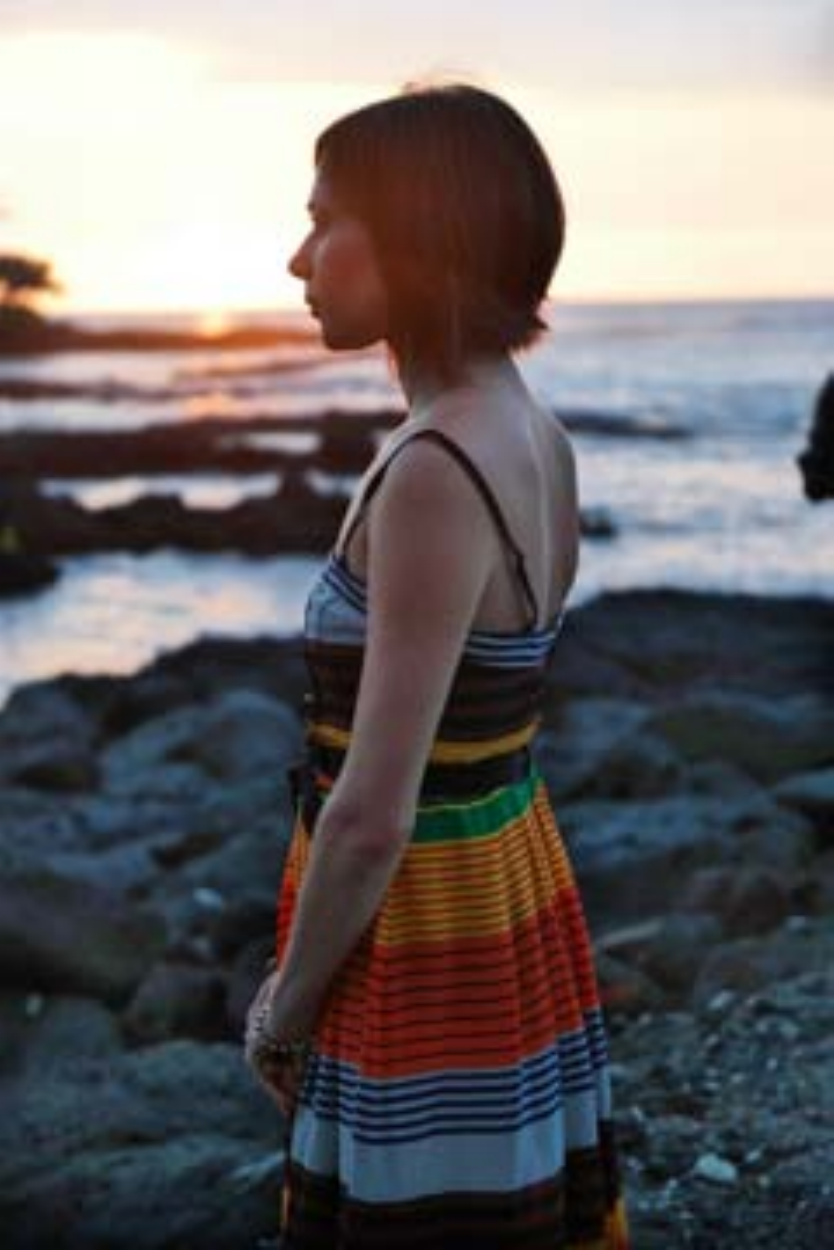}&
		\includegraphics[width=.16\linewidth]{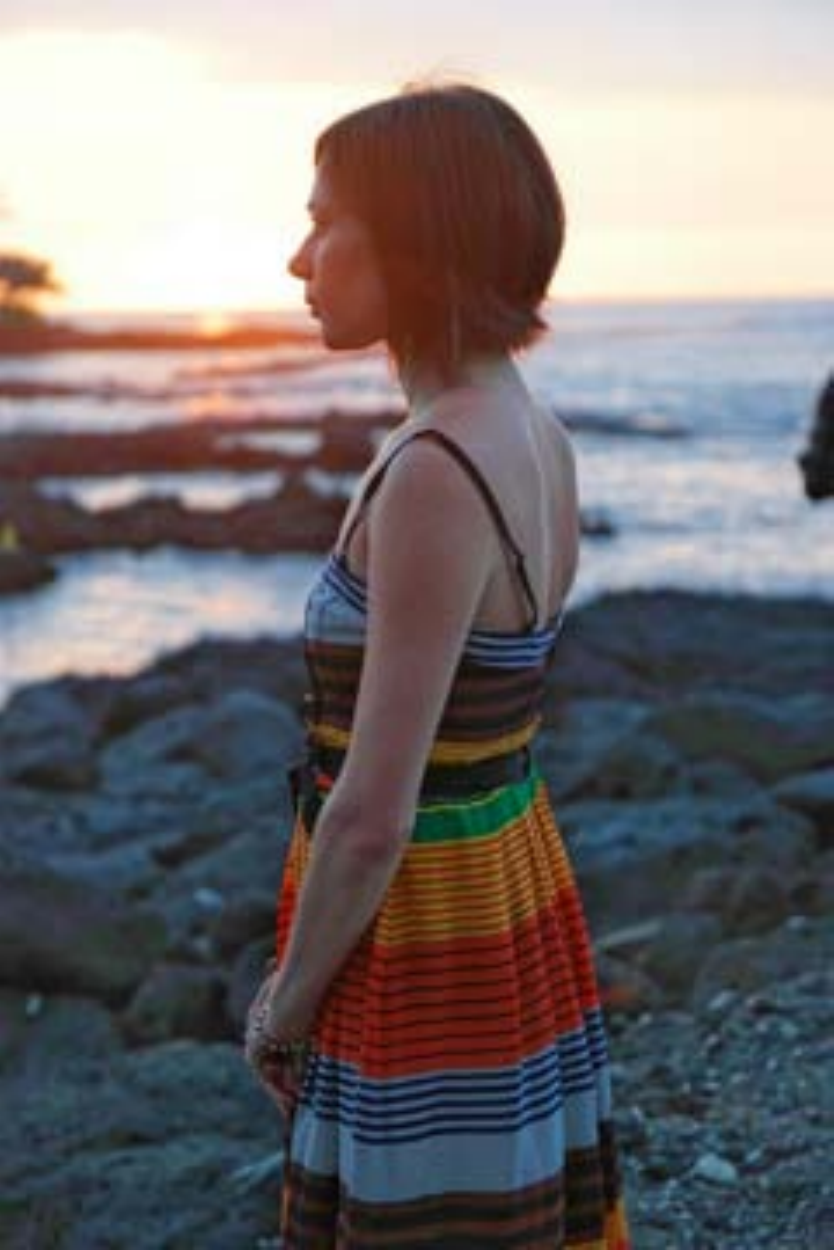}&
		\includegraphics[width=.16\linewidth]{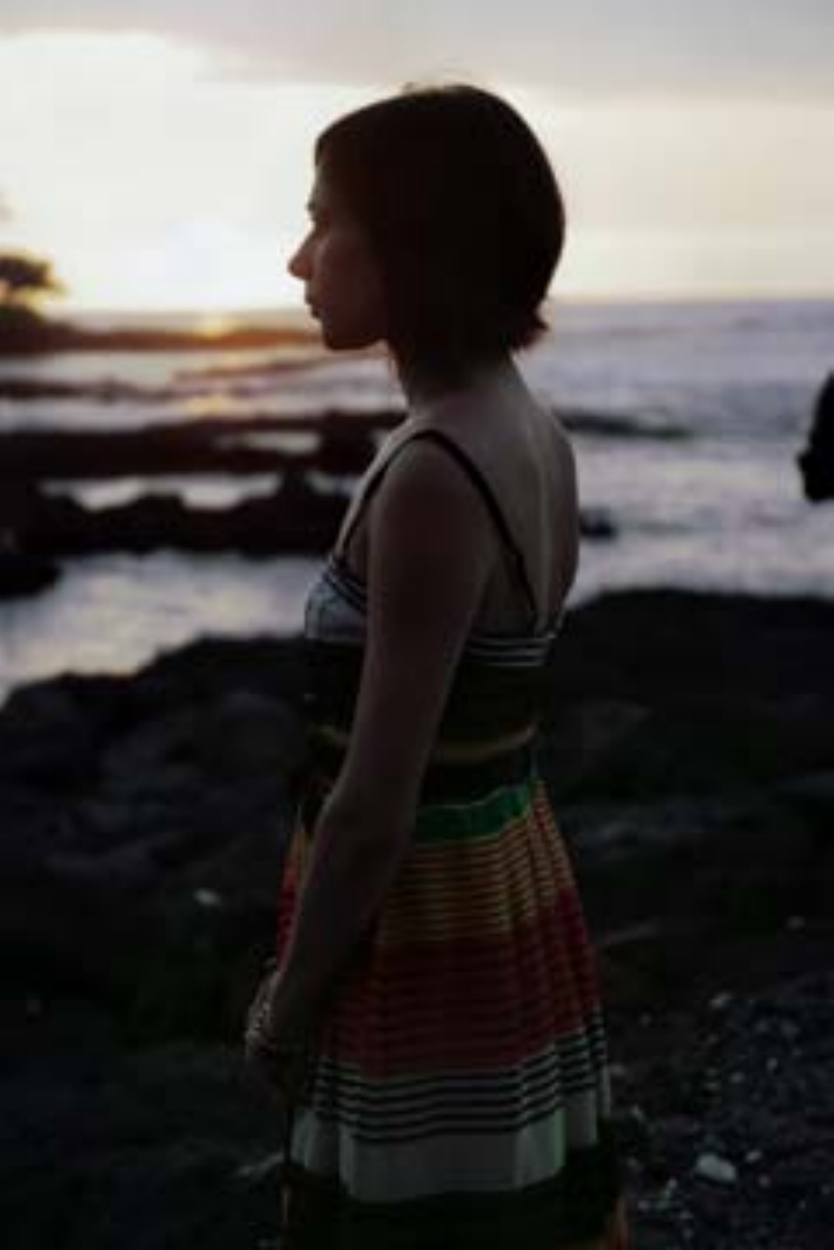}&
		\includegraphics[width=.16\linewidth]{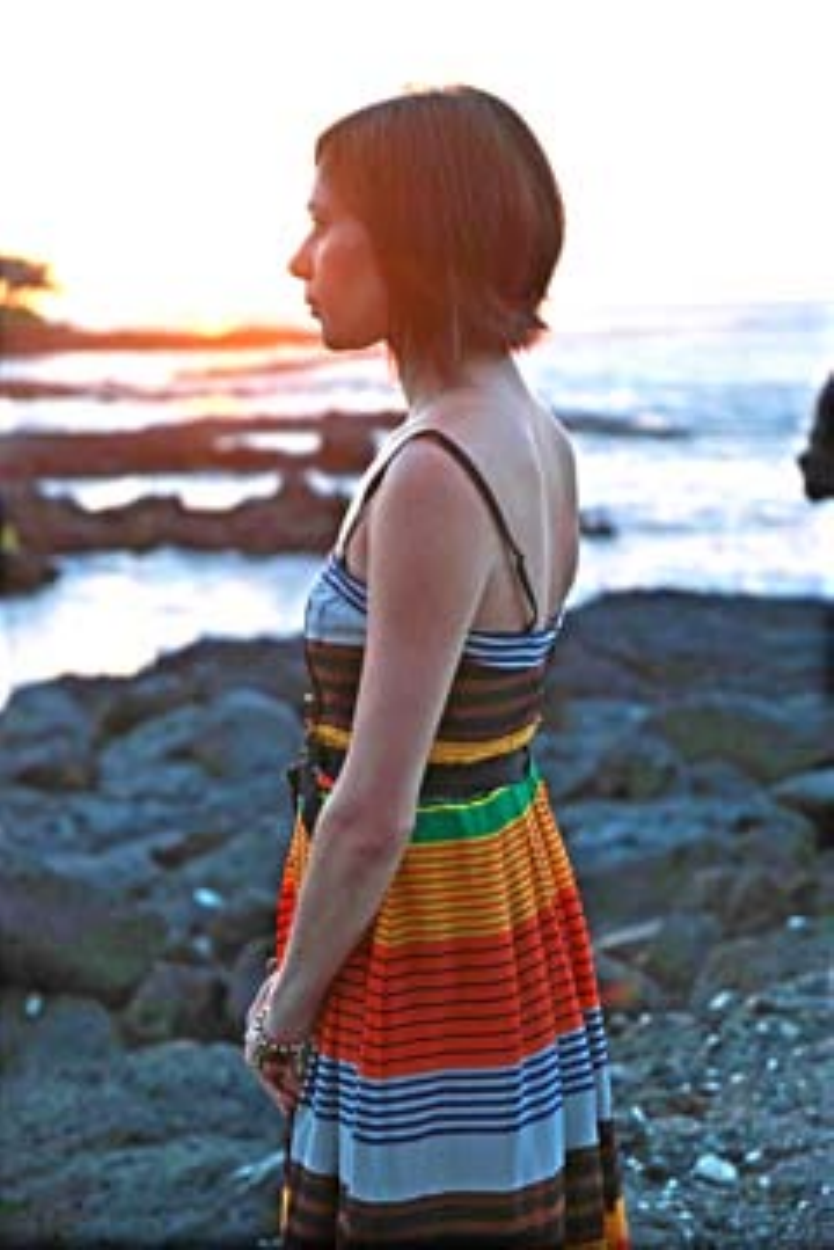}\\
		\small{5.07}&\small{4.43}&\small{4.27}&\small{3.98}&\small{4.66}&\small\textbf{3.91}\\
		\small{Input}&\small{SRIE}&\small{WVM}&\small{JIEP}&\small{HDRNet}&\small{TECU}\\
	\end{tabular}
	\vspace{-10pt}
	\caption{Comparisons on more examples selected from the NASA dataset \cite{nasa} (top three examples) and Non-uniform dataset \cite{wang2013naturalness} (last five examples). The NIQE scores are reported below each image.}
	\label{fig:supp_NASA}
\end{figure*}

\end{onecolumn}

\end{document}